\documentclass{article}

\usepackage[preprint]{neurips_2026}
\usepackage{graphicx}
\usepackage{algorithm}
\usepackage{algorithmic}
\usepackage{amsmath}
\usepackage{amsthm}
\usepackage{multirow}
\usepackage{subcaption}
\usepackage{silence}
\usepackage[utf8]{inputenc} 
\usepackage[T1]{fontenc}    
\usepackage{hyperref}       
\usepackage{url}            
\usepackage{booktabs}       
\usepackage{amsfonts}       
\usepackage{nicefrac}       
\usepackage{microtype}      
\usepackage{xcolor}         

\newif\ifrebuttal
\ifrebuttal
  
\else
  
\fi

\newtheorem{theorem}{Theorem}[section]

\newtheorem{definition}[theorem]{Definition}

\newtheorem{remark}{Remark}[section]

\usepackage{adjustbox}

\title{Sparse Deep Additive Model with Interactions:\\ Enhancing Interpretability and Predictability}

\author{%
  Noah Yi-Ting Hung\\
  Department of Mathematics and Statistics\\
  Georgia State University\\
  Atlanta, GA 30303 \\
  \texttt{yhung7@gsu.edu} \\
  \And
  Li-Hsiang Lin\\
  Department of Mathematics and Statistics\\
  Georgia State University\\
  Atlanta, GA 30303 \\
  \texttt{lhlin@gsu.edu} \\
  \And
  Vince D. Calhoun\\
  Tri-institutional Center for Translational Research in Neuroimaging and Data Science\\ Georgia State University, Georgia Institute of Technology,  Emory University\\
  Atlanta, GA 30303\\
  \texttt{vcalhoun@gsu.edu}
}

\begin{document}

\maketitle

\begin{abstract}
  Recent advances in deep learning highlight the need for personalized models that can learn from small samples, handle high-dimensional features, and remain interpretable. To address this, we propose the Sparse Deep Additive Model with Interactions (SDAMI), a framework that combines sparsity-driven feature selection with deep subnetworks for flexible function approximation. Central to SDAMI is the Effect Footprint principle, which posits that higher-order interactions leave detectable marginal traces on constituent variables, enabling their discovery without exhaustive search. SDAMI executes this principle through a three-stage strategy: (1) screening for footprint variables, (2) disentangling main effects from interactions via group lasso, and (3) modeling components with dedicated deep subnetworks. Theoretical analysis confirms that footprints vanish only under measure-zero symmetry conditions that are rare in practice, ensuring consistent interaction recovery. Extensive simulations demonstrate that SDAMI successfully identifies pure interactions that heredity-based baselines fundamentally miss, recovering complex effect structures with near-zero false positive rates. Together, these results position SDAMI as a principled framework for interpretable high-dimensional regression.
\end{abstract}

\section{Introduction}
Deep learning has achieved strong performance in data-rich settings~\citep{he2020deep}, but many scientific problems arise in small-$n$, large-$k$ regimes where observations are limited, predictors are numerous, and interpretability is essential~\citep{cesario2024survey, collins2024tripod+,jain2002personalized,stefanicka2023personalised,zhou2015learning}. In these settings, highly flexible black--box models can overfit, while aggressive screening may discard weak but scientifically meaningful signal. This tension is especially pronounced in application such as neuroscience, where we analyze single-cell activity with roughly $n=500$ observations and over $k=11{,}000$ candidate features and the researchers seek both accurate prediction and interpretable scientific discovery.

Scientific studies often require more than accurate prediction: they also require identifying which variables matter and how they affect the response~\citep{wang2021hybrid,molnar2020interpretable,hastie2009elements}. Structured regression models are appealing in such settings because they separate effects into interpretable components that support hypothesis and diagnostics. However, sparse additive models primarily capture univariate structure and can miss interaction-driven signals, particularly when interactions are present without strong associated main effects. For example, in V1 fMRI studies, classical sparse additive models can flexibly model marginal effects but may miss biologically meaningful higher-order associations among Gabor features~\citep{kay2008identifying,vu2008nonparametric}. 

To address this challenge, we propose the Sparse Deep Additive Model with Interactions (SDAMI), a structured deep additive framework for discovering and estimating main and interaction effects without imposing heredity constraints. SDAMI is built on a new principle, the \emph{Effect Footprint}, which states that higher-order interactions can leave detectable marginal traces on constituent variables even when strong main effects are absent. This differs fundamentally from heredity-based approaches, which restrict interactions to cases where at least one or both of the constituent main effects are present~\citep{bien2013lasso, lim2015learning, choi2010variable}. The distinction is important in applications where interaction-driven signal may exist without strong univariate effects.

SDAMI operationalizes this idea in three stages. First, it screens variables with either main effects or interaction footprints. Second, it uses structured regularization to disentangle main effects from interactions~\citep{simon2013sparse, yuan2009structured, zhao2009composite}. Third, it assigns selected effects to dedicated subnetworks, yielding flexible nonlinear estimation with effect-level interpretability. This design balances transparency and expressive power, and simulations show that it recovers effect structure while avoiding underfitting of main effects and overfitting of interactions.

{\bf Related work and differences.} Existing approaches to interpretable high-dimensional regression fall into three broad categories. First, conventional deep neural networks can represent complex interactions, but their entangled architectures obscure variable-level contributions and post-hoc explanations can become unstable in small-$n$ or low-signal settings~\citep{he2020deep, molnar2020interpretable}. Second, additive neural models and sparse additive methods improve interpretability, but heredity constraints limit their ability to detect pure or higher-order interactions~\citep{agarwal2021neural, vaughan2018explainable, yang2021gami,fan2011nonparametric,ravikumar2009sparse,fan2011sparse}. Third, structured sparsity methods and their deep extensions provide principled feature selection, but they typically retain heredity-type restrictions and do not explicitly disentangle main-effect capacity from interaction capacity~\citep{yuan2006model, scardapane2017group,wen2016learning, xu2023sparse, chang2021node, enouen2022sparse, kim2022higher}. SDAMI unifies and extends these directions. It uses effect footprints to screen influential variables, separates main and interaction roles before final network fitting, and introduces interaction subnetworks only when supported by the data. As a result, SDAMI permits models driven primarily by interactions without imposing heredity constraints, while retaining structured sparsity and effect-level interpretability~\citep{patel2020machine, shah2016modelling}. Beyond these lines of work, the broader statistics literature contains several families of interpretable nonlinear models, including multivariate or fractional-polynomial constructions, symbolic or logic-based regression, and Bayesian generalized nonlinear models~\citep{royston1994regression,sauerbrei1999building,schmidt2009distilling,fahrmeir2001bayesian}. These approaches offer different trade-offs in functional flexibility, search strategy, uncertainty quantification, and scalability. Our goal is not to subsume all such approaches, but to address a specific regime: small-$n$, large-$k$ regression with potentially pure or higher-order interactions, where exhaustive interaction search is infeasible and effect-level visualization remains desirable. In that sense, SDAMI should be viewed as a structured screening-and-estimation framework tailored to high-dimensional interaction discovery rather than a universal replacement for all interpretable nonlinear modeling approaches.

SDAMI introduces a principled framework, tailored to small-$n$, large-$k$ regression, that discovers and models higher-order interactions while maintaining effect-level interpretability. The framework rests on two core ideas: (1) leveraging the \emph{Effect Footprint} to screen variables before exploring the full interaction space, and (2) structuring deep 
subnetworks according to the type of regression effects (main vs. interaction) recovered in Stages 1--2. Our main contributions are:

\begin{itemize}
    \item We propose SDAMI, a three--stage framework for sparse additive-plus-interaction modeling in small-$n$, large-$k$ settings. At its core is the \emph{effect footprint} principle, derived from the Hoeffding-Sobol decomposition, which detects interaction-bearing variables through their marginal traces enabling discovery of pure and higher-order interactions without imposing heredity constraints, and reducing the second-order search space from $\binom{k}{2}$ pairs to $\binom{s}{2}$ where $s$ is the size of the screened variable set and $s\ll k$.

    \item We establish four theoretical results: Theorem~\ref{thm:sdami-vanish} characterizes when effect footprints are detectable, showing that they vanish only under degenerate symmetry conditions that are rare in practice; Theorem~\ref{thm:sdami-consistency} establishes effect-level selection consistency without imposing heredity constraints; Theorem~\ref{thm:sdami-prob-conv} proves prediction convergence in probability for the final SDAMI estimator; and Theorem~\ref{thm:sdami-finite-sample} provides a finite-sample prediction bound conditional on the selected active set.
    \item We empirically demonstrate that SDAMI improves both predictability and interpretability over state-of-the-art interpretable models, achieving high true positive rate (TPR) with near-zero false positive rate (FPR) and producing informative component-function visualizations on synthetic and real-data benchmarks.
\end{itemize}

\section{Problem setup and response-guided structured deep framework}

We observe regression data $\{(\mathbf{X}_i,Y_i)\}_{i=1}^n$, where $\mathbf{X}_i=(X_{i1},\ldots,X_{ik})^\top \in \mathbb{R}^k$ denotes the predictors and $Y_i \in \mathbb{R}$ is the response. The true regression function is assumed to follow a sparse additive-plus-interaction structure of the form
\begin{equation}\label{eq:true}
    Y_i = \sum_{j\in \mathcal{M}} f_j(X_{ij}) + f(\mathbf{X}_{i,\mathcal{I}}) + \epsilon_i,
\end{equation}
where $\mathcal{M}\subseteq \{1,\ldots,k\}$ denotes the set of active main effects, $\mathcal{I}\subseteq \{1,\ldots,k\}$ denotes the set of variables entering the interaction component, and $\epsilon_i$ is a random error with $\mathbb{E}[\epsilon_i]=0$ and $\mathrm{Var}(\epsilon_i)=\sigma^2$. The interaction component, $f(\mathbf{X}_{i,\mathcal{I}})$, is not restricted to pairwise effects and may represent a higher--order multivariate effect over $\mathcal{I}$; it captures interactions of arbitrary order $d\leq|\mathcal{I}|$ among the selected variables. We assume $|\mathcal{M}|=p \ll k$, so that only a small fraction of predictors directly contribute as main effects. We define $|\mathcal{I}\setminus \mathcal{M}|=q$, capturing variables that contribute exclusively through interactions but not as main effects. The sets $\mathcal{M}$ and $\mathcal{I}$ are not necessarily nested. In general, $\mathcal{I}$ may contain variables that contribute only through interactions but not as main effects, i.e., $q\neq 0$, including a scenario corresponds to interaction-only effects.

To estimate the model (\ref{eq:true}), we use separate subnetworks for univariate main effects $f_j$ and for the interaction component $f_{\mathcal{I}}(\cdot)$. Specifically, each selected $j^\text{th}$ main effect and interaction component are modeled with their own parameters $\theta_j$ and $\theta_{\mathcal{I}}$, respectively. Denote by $W^{(1)}_{\mathcal{M},j}$ the weight vector in the first hidden layer connecting input $X_j$ to its main-effect subnetwork, and by $W^{(1)}_{\mathcal{I},j}$ the weight vector connecting $X_j$ to the interaction subnetwork. The estimation problem is then formulated as
\begin{align}
    &\min_{\theta}\ \frac{1}{n}\sum_{i=1}^n 
\Bigg( Y_i 
- \sum_{j\in\mathcal{M}} \mathrm{NN}^{(j)}(X_{ij};\theta_j)
- \mathrm{NN}^{(\mathcal{I})}(\mathbf{X}_{i,\mathcal{I}};\theta_{\mathcal{I}})\Bigg)^2\label{eq:sdami-obj}\\
&\text{subject to}\quad 
\|W^{(1)}_{\mathcal{M},j}\|_\infty \;\leq\;  \kappa_{\mathcal{M}} \|f_j\|,
\quad j=1,\ldots,k,
\quad\|W^{(1)}_{\mathcal{I},j}\|_\infty \;\leq\; \kappa_{\mathcal{I}}  \|f_{\mathcal{I}}\|,
\quad j\in \mathcal{I}\label{eq:sdami-con}.
\end{align}
Here, $\mathrm{NN}^{(j)}(X_{ij};\theta_j)$ denotes a \emph{neural network (NN) submodule} dedicated to the $j$-th main effect, parameterized by weights $\theta_j$, while  $\mathrm{NN}^{(\mathcal{I})}(\mathbf{X}_{i,\mathcal{I}};\theta_{\mathcal{I}})$ 
denotes a subnetwork for the interaction set $\mathcal{I}$, parameterized by 
$\theta_{\mathcal{I}}$. Each NN is a standard feedforward network with hidden layers 
and nonlinear activations, serving as a flexible nonlinear approximator. We use ReLU subnetworks in the current theory mainly for analytical convenience. The footprint-based screening and grouped decomposition steps do not depend on the activation choice, and the practical implementation can also employ other activations such as GELU, SiLU, or Tanh. The reference functions $f_j$ and $f_{\mathcal I}$ represent the true main-effect and interaction-effect components of the regression function $f^\star$. The constraints in \eqref{eq:sdami-con} regulate the first-layer weights $W^{(1)}$ relative to 
$\|f_j\|$ and $\|f_{\mathcal I}\|$, ensuring that each subnetwork remains aligned 
with the magnitude of its corresponding effect and thereby preserving hierarchical 
structure and interpretability. If $\|f_j\|=0$, the outgoing weights 
$W^{(1)}_{\mathcal{M},j}$ vanish, excluding $X_j$ from its subnetwork. Similarly, if $\|f_{\mathcal{I}}\|=0$, connections into the interaction subnetwork are eliminated. Thus sparsity and interpretability are achieved not through explicit penalties, but through norm-based constraints that prune irrelevant effects, while the loss in \eqref{eq:sdami-obj} enforces predictive accuracy.

Direct optimization of the constrained problem (\ref{eq:sdami-obj}) without additional structure becomes infeasible in high dimensions, since it is difficult to distinguish relevant main effects from irrelevant variables or latent contributors to interactions. Another challenge in discovering the interaction component is that even if one restricts attention to pairwise interactions, the number of candidate interaction pairs grows quadratically with the feature count$-$namely, $\binom{k}{2}$ possible pairs. To overcome these challenges, we introduce the \emph{effect footprint} principle providing a mechanism for linking variable screening directly to the objective function and guiding the activation of subnetworks in a statistically coherent manner.

\section{Fitting sparse deep additive models with interactions (SDAMI)}

\begin{figure*}[t] 
    \centering
    \includegraphics[width=0.8\linewidth]{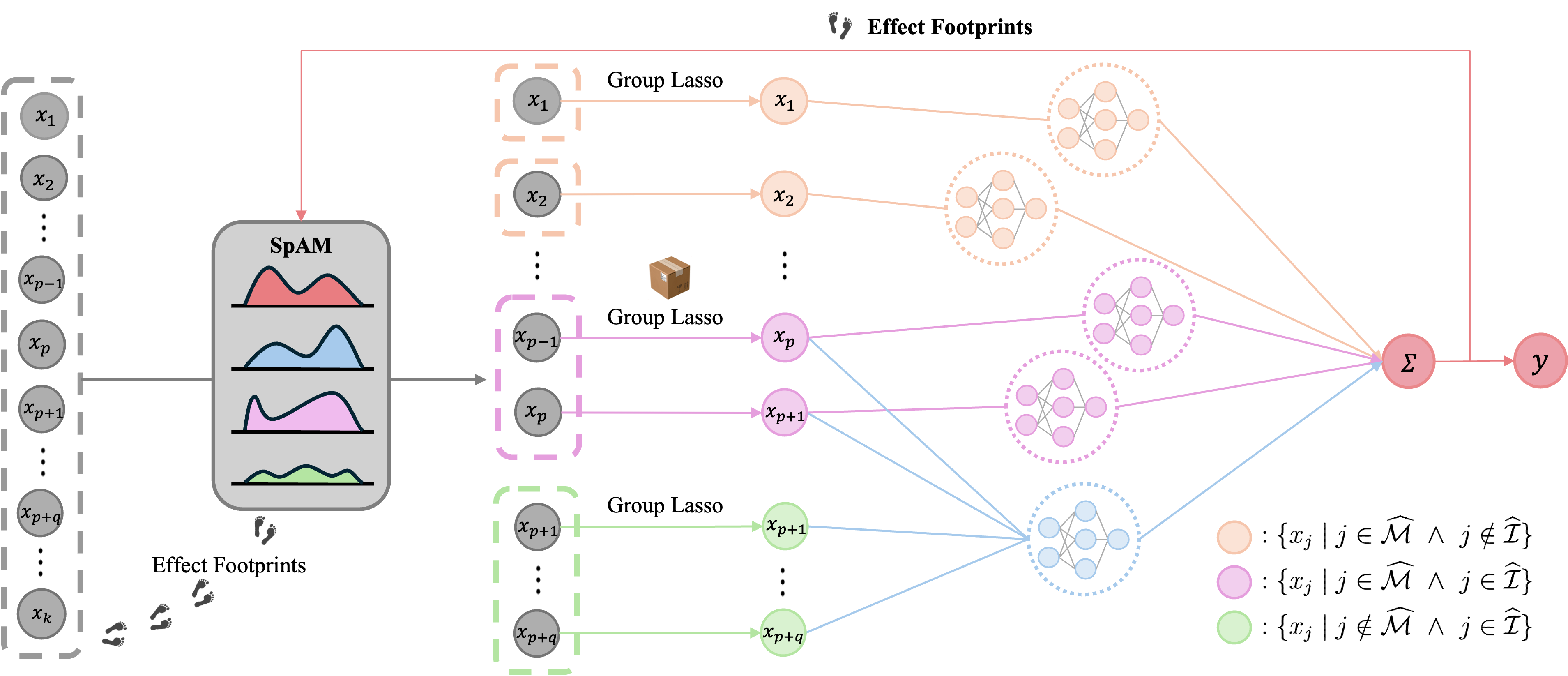}
    \caption{The SDAMI architecture. Stage 1 screens main and footprint variables; Stage 2 decomposes them into $\mathcal{M}$ and $\mathcal{I}$; Stage 3 activates dedicated main--effect subnetworks and an interaction subnetwork to enforce effect-level interpretable structure.}\label{fig: sdam_arc}
\end{figure*}

Since directly solving \eqref{eq:sdami-obj}--\eqref{eq:sdami-con} is difficult in high dimensions, SDAMI proceeds in three stages: screening, decomposition, and structured estimation. The first two stages identify a structural skeleton consisting of candidate main effect and interaction--relevant variables. The final stage then fits a structured neural predictor conditional on its recovered skeleton. In practice, we consider two variants: a version in which all variables in  $\widehat{\mathcal{I}}$ are passed to a shared interaction subnetwork $\mathrm{NN}^{(\mathcal{I})}(\mathbf{X}_{i,\mathcal{I}};\theta_{\mathcal{I}})$, and a pairwise variant, SDAMI$-p$, that operates on selected pairwise interactions.

\begin{definition}[Effect Footprint]\label{def:ef}
    If $X_j\in \mathcal{I}$, the \emph{effect footprint} is defined as:
    \[
        m_j(x) = \mathbb{E}\big[f(\mathbf{X}_{\mathcal{I}})\mid X_j = x\big];
    \]
\end{definition}

\paragraph{Stage 1: Effect footprint screening produces $\hat S$.}
We use Definition~\ref{def:ef} to identify an estimated active set $\widehat{\mathcal{S}}$. In particular, we fit the augmented additive model
\begin{equation}\label{eq:additive}
    Y_i = \sum_{j=1}^{p+q} f_j(X_{ij}) + \varepsilon_i,
\end{equation}
where indices $1,\ldots,p$ correspond to true main effects and $p+1,\ldots,p+q$ correspond to footprint variables. The model is fitted via SpAM with B--spline bases and the $\widehat{\mathcal{S}}=\{\, j : \hat f_j \neq 0 \,\}$ include both true main effects and interaction--only variables with detectable marginal footprints. This screening step avoids exhaustive search over the full $\binom{k}{2}$ interaction space and retains variables with either true main effects or non--negligible footprints.

\textit{Theoretical guarantee.}
Theorem 4.1 ensures that $m_j(x)$ vanishes only when the first-order Hoeffding--Sobol projection of $f(X_{\mathcal{I}})$ onto $X_j$ is zero, which corresponds to a rare perfect-symmetry condition. Thus, any $j \in (\mathcal{I}\setminus\mathcal{M})$ produces a nonzero footprint and is retained in $\hat S$ with high probability.

\paragraph{Stage 2: Group lasso decomposition produces $\hat{\mathcal{M}}, \hat{\mathcal{I}}$.}
To operationalize this decomposition, we construct grouped basis expansions on the screened set $\hat S$. For each variable $j \in \hat S$, we form a univariate basis block $\Phi_j \in \mathbb{R}^{n \times b}$ to represent candidate main effects. For each pair $(j,k)$ in the screened set, we form a pairwise interaction block $\Phi_{jk} = \Phi_j \otimes \Phi_k \in \mathbb{R}^{n \times b^2}$ using the row-wise tensor product. The group index set $\mathcal{G}$ contains $|\hat S|$ main-effect groups and $\binom{|\hat S|}{2}$ interaction groups, which is far smaller than the original $\binom{k}{2}$ candidate space. 

The stacked design is
$
X = [\, \Phi_1,\ldots,\Phi_{|\hat S|},\Phi_{12},\ldots \,]
$
with block coefficient vector
$\theta = (\beta_1,\ldots,\beta_{|\hat S|},\gamma_{12},\ldots).$ The block group-lasso solves
\[
\hat\theta \in \arg\min_{\theta}
\left\{
\frac{1}{2n}\|Y-X\theta\|_2^2
+\lambda_2\sum_{g\in\mathcal{G}} w_g \|\theta_g\|_2
\right\}.
\]
Here $\lambda_2$ is selected by 5-fold cross-validation. The output is
\[
\hat{\mathcal{M}}=\{\,j:\|\hat\beta_j\|_2\neq 0\,\},
\qquad
\hat{\mathcal{I}}=\{\, (j,k):\|\hat\gamma_{jk}\|_2\neq 0\,\}.
\]

\textit{Theoretical guarantee.}
Theorem 4.2 shows that under Assumptions A1--A7, the primal--dual witness construction yields
\[
\mathbb{P}\!\left(\{j:\hat f_j\neq 0\}=\mathcal{M}
\ \text{and}\ 
(\hat f_{\mathcal{I}}\neq 0 \Leftrightarrow f_{\mathcal{I}}\neq 0)\right)\to 1.
\]
This is effect-level oracle recovery without heredity constraints. Remark 4.3 emphasizes that this consistency depends only on convex Stages 1--2, independent of Stage 3.

\paragraph{Stage 3: Constrained deep regression.}
Given $\hat{\mathcal{M}}$ and $\hat{\mathcal{I}}$, we build
(i) a dedicated univariate subnetwork
$
\mathrm{NN}^{(j)}(X_{ij};\theta_j)
$
for each main effect $j\in\hat{\mathcal{M}}$, and
(ii) a shared multivariate subnetwork
$\mathrm{NN}^{(\mathcal{I})}(X_{i,\hat{\mathcal{I}}};\theta_{\mathcal{I}})$
for interactions. Sparsity is enforced via norm constraints
\[
\|W^{(1)}_{\mathcal{M},j}\|_\infty \le \kappa_{\mathcal{M}} \|f_j\|,
\qquad
\|W^{(1)}_{\mathcal{I},j}\|_\infty \le \kappa_{\mathcal{I}} \|f_{\mathcal{I}}\|.
\]
In implementation, the unknown norms $\|f_j\|$ and $\|f_\mathcal{I}\|$ in \eqref{eq:sdami-con} are replaced by $\|\hat{\beta}_j\|_2$ and $\|\hat{\gamma}_{jk}\|_2$. After each gradient step, first-layer weights are projected onto these bounds; if $|\hat f_j|=0$, the outgoing weights vanish, thereby pruning the corresponding subnetwork. 

For identifiability, main effects are defined under the usual centering constraints from additive modeling, while interaction structure is identified through the screening decomposition steps rather than prediction error alone. Figure~\ref{fig: sdam_arc} illustrates the SDAMI architecture and how structured constraints impose sparsity on the network. Additional implementation and tuning details are deferred to Appendix~\ref{alg: sdam}. Neural network architectures are selected via 5-fold cross-validation and the hyperparameter setting is provided in Appendix~\ref{appendix:hyperparameter}.

\section{Theoretical analysis: the role of effect footprint, selection consistency, model convergence}\label{sec:thm}
Under the model space definition in Appendix~\ref{App: proofthm41}, we provide the theoretical guarantees for SDAMI, focusing on three key strengths: (1) the robustness of \emph{effect footprint} as a screening proxy, (2) the \emph{effect-level selection consistency} of the model even in the presence of pure interactions, and (3) the predictive validity by proving that the fitted predictor converges in probability to the true model (\ref{eq:true}). Detailed assumptions are provided in Appendix~\ref{App: proofthm42}. 
\begin{theorem}[Ubiquity and robustness of effect footprints]\label{thm:sdami-vanish}
Let $\mathbf X_\mathcal I=(X_j,\mathbf Z)$ be the variables in an interaction $f(\mathbf X_\mathcal I)$ with $\mathbb E[f(\mathbf X_\mathcal I)]=0$. The marginal footprint is defined as 
\[
m_j(x)=\mathbb E[f(\mathbf X_\mathcal I)\mid X_j=x].
\]
Then $m_j(x)$ is constant iff the first–order projection of $f$ onto functions of $X_j$ vanishes in the Hoeffding–Sobol decomposition \citep{sobol1990sensitivity,sobol2001global}. In this case, $f$ contains only higher–order components involving $X_j$.
\end{theorem}

This characterization isolates the exceptional cases in which footprints fail: a variable leaves no detectable footprint precisely when its influence appears solely through higher–order interactions that vanish after averaging over the remaining inputs. Such a variable may still be essential via interactions, but univariate screening cannot detect it. Two canonical settings illustrate this: (i) independence with centering (e.g., bilinear forms of independent, mean-centered inputs), and (ii) perfect symmetry with antisymmetric interactions (e.g., the XOR rule for binary data or odd functions under symmetric continuous inputs). These conditions are stringent; in practice predictors are correlated, distributions seldom perfectly symmetric, and noise disrupts exact cancellations. Consequently, footprints typically exist, providing a robust signal for screening. A detailed proof is given in Appendix~\ref{App: proofthm41} of the supplementary material.

\begin{theorem}[Effect-level selection consistency of SDAMI]
\label{thm:sdami-consistency}
Under the assumptions detailed in Appendix~\ref{App: proofthm42}, as $n\to\infty$,
\[
\mathbb{P}\!\left(
  \big\{j:\ \widehat f_j\neq 0\big\}=\mathcal M\ \text{ and }\
  (\widehat f_{\mathcal I}\neq 0\Leftrightarrow f_{\mathcal I}\neq 0)
\right)\ \longrightarrow\ 1
\]
\end{theorem}
Thus SDAMI does not merely exploit footprints heuristically; it achieves a rigorous form of oracle recovery. Crucially, the consistency holds without enforcing the Effect Heredity constraint. As $n$ grows, SDAMI selects exactly the true set of main effects and correctly detects the interaction with probability tending to one, ensuring that the discovered structure reflects the underlying generative mechanism. The proof (Appendix~\ref{App: proofthm42} of the supplementary material) employs a block-wise primal–dual witness argument for the group-lasso formulation, leveraging footprint-induced group signals and oracle inequalities for group sparsity \citep{Lounici2011GroupSparsity,negahban2009unified}.

\begin{remark}
    [Independence from neural optimization] The consistency result in Theorem~\ref{thm:sdami-consistency} depends solely on the convex surrogate loss used in Stages 1 and 2. Consequently, the identification of the structural skeleton is statistically guaranteed, independent of the global convergence properties of the non-convex neural network training in Stage 3.
\end{remark}
\begin{theorem}[Prediction convergence in probability for SDAMI]
\label{thm:sdami-prob-conv}
Let $\widehat A_n$ be the SDAMI-selected index set and let $\widehat f_n$ be the SDAMI estimator. Suppose \textnormal{(B1)–(B6)} hold. Then, for every fixed $\varepsilon>0$,
\[
\mathbb P\!\left(\,|\widehat f_n(\mathbf X)-f^\star(\mathbf X)|\ \ge\ \varepsilon\,\right)\ \longrightarrow\ 0
\quad\text{as } n\to\infty,
\]
\end{theorem}
\begin{remark}
    The convergence Theorem \ref{thm:sdami-prob-conv} relies on Assumption (B4), requiring the optimizer to achieve near-optimal empirical risk. This explicitly acknowledges the dependency on the optimization algorithm successfully finding a good local minimum within the over-parameterized networks. This is a standard assumption in deep learning theory and is empirically validated by our experiments.
\end{remark}
\vspace{-0.4cm}
The key idea is to combine sieve approximation with uniform generalization. Selection consistency concentrates learning on the correct coordinates; empirical risk minimization up to a vanishing tolerance, together with a uniform law of large numbers for squared loss (via Rademacher and covering bounds for norm–constrained networks), transfers empirical to population $L_2$–risk \citep{BartlettMendelson2002,MohriRostamizadehTalwalkar2018,vandeGeer2000}. In parallel, ReLU approximation theory ensures the sieve approximates the oracle regression under a suitable growth schedule \citep{Barron1993,Yarotsky2017,SchmidtHieber2020,Suzuki2019}. A uniform $L_2$ envelope (implied by norm constraints and square-integrability) guarantees uniform integrability, so vanishing population risk implies vanishing misfit probability via a Markov-type bound. Full details appear in Appendix~\ref{App: proofthm43} of the supplementary material.

The preceding results establish asymptotic effect-level recovery and prediction consistency. We next state a finite-sample counterpart, which
makes explicit how the prediction error depends on the probability of selection failure, the complexity of the constrained SDAMI class, and the optimization tolerance in Stage 3.  A detailed proof is given in Appendix~\ref{App: proofthm44} of the supplementary material.

\begin{theorem}[Finite-sample prediction guarantee under the selected active effect set]
\label{thm:sdami-finite-sample}
Suppose Assumptions \textnormal{(A1)--(A7)} and \textnormal{(B1)--(B6)}
hold. Let \(P\) denote the population expectation with respect to an
independent test point \((\mathbf X,Y)\), let \(P_n\) denote the empirical
expectation over the training dataset
\(\mathcal D_n=\{(\mathbf X_i,Y_i)\}_{i=1}^n\), and let
\(f^\star(\mathbf X)\) be the true regression function specified in
\eqref{eq:true}. Let \(\widehat A_n\) be the active effect set selected by
Stages 1--2 of SDAMI, and let $\widehat f_n\in \mathcal F_n^{\mathrm{SDAMI}}(\widehat A_n)$
be the Stage 3 estimator fitted over the selected SDAMI class. Then, for every \(\varepsilon>0\),
\[
\mathbb P_{\mathbf X}
\left(
|\widehat f_n(\mathbf X)-f^\star(\mathbf X)|\ge \varepsilon
\,\middle|\,\mathcal D_n
\right)
\le
\frac{
\inf_{f\in\mathcal F_n^{\mathrm{SDAMI}}(\widehat A_n)}
P\!\left[
\{f(\mathbf X)-f^\star(\mathbf X)\}^2
\right]
+
2R_n(\widehat A_n)
+
\delta_n
}{
\varepsilon^2
},
\]
where \(\mathbb P_{\mathbf X}(\cdot\mid \mathcal D_n)\) denotes probability
with respect to an independent test covariate \(\mathbf X\), conditional on
the training data. The selected-class generalization error is defined as
\[
R_n(\widehat A_n)
=
\sup_{f\in\mathcal F_n^{\mathrm{SDAMI}}(\widehat A_n)}
\left|
P\!\left[\{Y-f(\mathbf X)\}^2\right]
-
P_n\!\left[\{Y-f(\mathbf X)\}^2\right]
\right|,
\]
and \(\delta_n\ge0\) is the Stage 3 optimization tolerance satisfying
\[
P_n\!\left[\{Y-\widehat f_n(\mathbf X)\}^2\right]
\le
\inf_{f\in\mathcal F_n^{\mathrm{SDAMI}}(\widehat A_n)}
P_n\!\left[\{Y-f(\mathbf X)\}^2\right]
+\delta_n.
\]
\end{theorem}

Theorem~\ref{thm:sdami-finite-sample} provides a finite-sample
counterpart to Theorem~\ref{thm:sdami-prob-conv}. It shows that, on the
event that SDAMI recovers the correct effect structure, the prediction
error is controlled by the oracle approximation error, the finite-sample
generalization error, and the numerical optimization.

\section{Numerical experiments}\label{sec: ns}

We evaluate SDAMI on six synthetic settings designed to isolate different combinations of main effects and interactions, including weak--main, pure--interaction, and overlap scenarios. For each setting, we consider sample sizes \(n \in \{150,300,450\}\) with feature dimension fixed at \(k=150\), where only a small subset of variables is active. Responses are generated as model~\ref{eq:true}, with \(X_i \sim \mathrm{Uniform}(-2.5,2.5)\) independently and \(\epsilon_i \sim N(0,\sigma^2)\) with $\sigma^2=1$ in our setting. The true functions are drawn from representative linear and nonlinear forms, and the detailed functional forms are provided in Table~\ref{tab: setting}. We compare SDAMI with interpretable baselines including NAM, GAMI-Net, NODE-GAM, NODE-GA2M, fSpAM, and LASSO, as well as a standard DNN~\citep{agarwal2021neural,yang2021gami,chang2021node,lemhadri2021lassonet,ravikumar2009sparse}. Architecture selection for SDAMI is performed by cross-validation, with implementation details summarized in Appendix~\ref{appendix:hyperparameter} and Appendix~\ref{appendix:computation}.

\begin{table}[t]
    \centering
    \caption{The summary table for numerical simulation models; Case 1: Only main effects; Case 2: Main effects with a weak signal; Case 3--5: Main effects with different degree of overlap; Case 6: Only interaction effects}
    \label{tab: setting}
    \begin{tabular}{cl||ll}
    Case & Functional Form & Function & Definition \\ \hline
    1    & $y=f_1(x_1)+f_2(x_2)+f_3(x_3)+f_4(x_4)$ & $f_1(x_i)$ & $-2\sin(2x_i)$\\
    2    &  $y=f_1(x_1)+f_2(x_2)+f_3(x_3)+0.01f_4(x_4)$ & $f_2(x_i)$ & $\frac{x_i^2}{2}+1$\\
    3    & $y=f_1(x_1)+f_2(x_2)+f_3(x_3)+f_5(x_4,x_5)$ & $f_3(x_i)$ & $x_i-\frac{1}{2}$\\
    4    &  $y=f_1(x_1)+f_2(x_2)+f_3(x_3)+f_5(x_3,x_4)$ & $f_4(x_i)$ & $e^{-x_i}+e^{-1}-1$\\
    5    &  $y=f_1(x_1)+f_2(x_2)+f_3(x_3)+f_5(x_2,x_3)$ & $f_5(x_i,x_j)$ & $e^{\sin(x_i)+\cos(x_j)-1}$\\
    6    &  $y=f_5(x_1,x_2)+f_5(x_3,x_4)$ & $-$& $-$ 
    \end{tabular}
\end{table}

\begin{figure}[t!]
    \centering
    \includegraphics[width=1.0\linewidth]{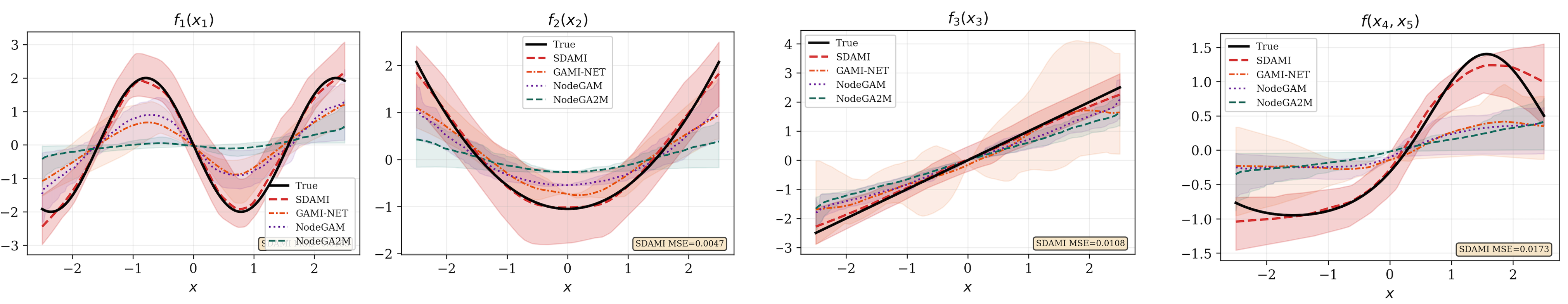}
    \caption{(Case 3) The left three panels show recovered main effects and the right panel shows the recovered interaction surface, together with Wald-type 95\% bands for the estimated components.}
    \label{fig:case3visual}
    \vspace{-5mm}
\end{figure}

Across all six settings, SDAMI is consistently among the strongest methods across the six settings. In Table~\ref{tab: n150}, it captures nonlinear main and interaction effects without sacrificing interpretability. The advantage is most pronounced in interaction-dominant settings (Case 3--6), where heredity-constrained models (NAM, GAMI-Net) are less competitive, whereas SDAMI significantly reduces RMSE (e.g, 0.46 vs. 0.97 for NAM in Case 6), confirming that SDAMI continues to recover the underlying structure. Figure~\ref{fig:case3visual} illustrates that SDAMI can recover both main-effect shapes and pure interaction surfaces in a representative synthetic case. Stability across 100 replications affirms robustness, while improvements from $n=150$ to $n=300$ confirm scalability.

We further evaluate effect recovery using true positive rate (TPR) and false positive rate (FPR). Table~\ref{tab: tpr150} shows that SDAMI maintains high TPR while keeping FPR near zero, especially in interaction settings where LASSONET and SODA lose sensitivity or introduce more false discoveries \citep{lemhadri2021lassonet,yang2015}. In more complex settings with overlapping and non-overlapping interactions (Cases 3–6), SDAMI maintains substantially higher TPRs than LASSONET and SODA, which experience steep sensitivity drops. Crucially, regarding interpretability, SDAMI achieves a FPR of near zero. This conservatism is important for trustworthy interpretation: unlike LASSONET or SODA which may identify spurious interactions, SDAMI is conservative. These results indicate that effect-footprint screening reduces the effective interaction search space while preserving reliable support recovery.

Additional results for \(n=300\) and \(n=450\), component-wise estimation error, and robustness under more realistic design conditions are reported in the appendix. In particular, to address concerns that the main benchmark tables provide only a partial view of performance, Appendix~\ref{app:pareto}--\ref{app:component} further evaluate SDAMI in terms of learning behavior, stage-wise ablation, and the accuracy and stability of recovered component functions. Additionally, to further assess the practical robustness of SDAMI beyond the baseline synthetic design, we additionally consider stress-test settings with correlated covariates (Appendix~\ref{app:corr}), heteroscedastic noise (Appendix~\ref{app:heteroscedastic}), and binary GLM responses (Appendix~\ref{app:binomial}), which probe more realistic departures from the idealized assumptions and address finite-sample concerns about sensitivity to design misspecification. Across these additional experiments, SDAMI remains competitive and typically retains strong support recovery and predictive performance, indicating that its empirical advantages are not confined to the independent Gaussian regression setting. Detailed results are deferred to Appendix H and the supplementary material.

\begin{table*}[t]
    \centering 
    \small
    \caption{RMSE over 100 simulations $\pm$ standard deviation ($n=150$).}
    \label{tab: n150}
    \begin{adjustbox}{max width=1.0\textwidth}
    \begin{tabular}{lcccccccccc}
    \toprule
    & SDAMI & DNN & fSpAM & LASSO & NAM & GAMI-NET & NODE-GA$^2$M & NODE-GAM \\
    \midrule
    Case 1 &\textbf{0.68}$_{\pm 0.59}$ & 14.37$_{\pm 1.03}$ & 5.84$_{\pm 0.52}$ & 4.77$_{\pm 0.94}$& 14.36$_{\pm 1.35}$ & 5.83$_{\pm 3.09}$ & 1.31$_{\pm 1.53}$ & 2.42$_{\pm 2.52}$   \\
    Case 2 &\textbf{0.57}$_{\pm 0.75}$& 5.39$_{\pm 0.42}$ & 3.22$_{\pm 0.25}$ & 3.02$_{\pm 0.37}$ & 5.19$_{\pm 0.42}$ & 2.26$_{\pm 1.05}$ & 0.72$_{\pm 0.81}$ & 1.58$_{\pm 1.64}$   \\
    Case 3 &\textbf{0.58}$_{\pm 0.88}$& 5.78$_{\pm 0.43}$ & 3.55$_{\pm 0.25}$ & 3.61$_{\pm 0.39}$ & 5.68$_{\pm 0.49}$ & 2.56$_{\pm 1.02}$ & 1.00$_{\pm 1.10}$ & 1.83$_{\pm 1.95}$ \\
    Case 4 &\textbf{0.52}$_{\pm 0.91}$& 7.11$_{\pm 0.51}$ & 3.53$_{\pm 0.27}$ & 3.34$_{\pm 0.40}$ & 6.95$_{\pm 0.56}$ & 3.05$_{\pm 1.29}$ & 0.96$_{\pm 1.07}$ & 1.82$_{\pm 1.93}$  \\
    Case 5 &\textbf{0.44}$_{\pm 0.79}$& 5.90$_{\pm 0.43}$ & 3.65$_{\pm 0.27}$ & 3.59$_{\pm 0.41}$ & 5.77$_{\pm 0.48}$ & 2.33$_{\pm 1.18}$ & 0.85$_{\pm 0.99}$ & 1.75$_{\pm 1.83}$  \\
    Case 6 &\textbf{0.46}$_{\pm 0.24}$& 1.05$_{\pm 0.11}$ & 0.63$_{\pm 0.05}$ & 0.61$_{\pm 0.10}$ & 0.97$_{\pm 0.09}$ & 5.85$_{\pm 3.09}$ & 0.48$_{\pm 0.29}$ & 0.52$_{\pm 0.33}$ \\
    \hline
    \bottomrule
    \end{tabular}
    \end{adjustbox}
    
\end{table*}

Permutation-based significance assessment for selected components is reported in Appendix~\ref{appendix:test}, with Benjamini--Hochberg correction across the tested components. Let $G = \{x_1, \ldots, x_{|{G}|}\}$ denote a fixed evaluation grid of $|{G}| = 200$ equally-spaced points on the feature support $[-2.5, 2.5]$. For each component $j$ in the active set $\hat{\mathcal{A}}_n$ selected by SDAMI's Stage 2, we test $H_0:f_j=0$ versus $H_1:f_j\neq 0$. The test statistic is the squared $L_2$ norm of the centered estimated function on the evaluation grid: $T_j = (1/|G|) \sum_{x\in G} (\hat f_j(x) - \bar f_j)^2$ where $\bar f_j= (1/|G|) \sum_{x\in G} \hat f_j(x)$. The null distribution is generated by permuting the component's columns of the training and validation feature matrices independently across rows, refitting Stage 3 on the permuted data, and computing $T_j^{(b)}$ from the new estimate. With $B = 100$ permutations, the empirical p-value is $p_j = (1 + |\{b : T_j^{(b)} \geq T_j\}|) / (B + 1)$. Multiple-testing correction across components and scenarios is applied via Benjamini-Hochberg at FDR $= 0.05$~\citep{benjamini1995controlling,phipson2016permutation}.

\begin{table*}[t]
    \centering
    \small
    \caption{Mean and standard deviation of TPR (Higher is better) and FPR (Lower is better) over 100 simulations from SDAMI, LASSONET, SODA when $n=150$ where $(-)$ indicates value $<1e^{-5}$.}
    \label{tab: tpr150}
    \begin{adjustbox}{max width=1.0\textwidth}
    \begin{tabular}{lcccccc}
    \toprule
    \multicolumn{1}{c}{Method} & \multicolumn{2}{c}{SDAMI} &\multicolumn{2}{c}{LASSONET} &\multicolumn{2}{c}{SODA} \\
    \cmidrule(lr){2-3} \cmidrule(lr){4-5} \cmidrule(lr){6-7} 
    & TPR $\uparrow$& FPR $\downarrow$& TPR $\uparrow$& FPR $\downarrow$& TPR $\uparrow$& FPR $\downarrow$\\
    \midrule
    Case 1 & \textbf{1.000} $(-)$ & $1.1\times10^{-5}$ $(-)$ & 0.490 (0.049)& 0.004 (0.003) & 0.018 (0.0641)& $6\times 10^{-4}$ ($2\times 10^{-4}$)\\
    Case 2 & \textbf{1.000} $(-)$ & $1.1\times10^{-5}$ $(-)$ & 0.255 (0.035)& 0.010 (0.014) & 0.048 (0.105) & $1\times 10^{-3}$ ($2\times 10^{-4}$)\\
    Case 3 & \textbf{0.750} $(-)$ & $10^{-4}$ ($10^{-5}$) & 0.140 (0.301) & 0.172 (0.281) & 0.025 (0.075) & $6\times 10^{-4}$ ($3\times 10^{-4}$)\\
    Case 4 & \textbf{0.760} $(-)$ & $10^{-4}$ ($10^{-5}$) & 0.130 (0.305) & 0.162 (0.270) & 0.040 (0.105) & $7\times 10^{-4}$ ($3\times 10^{-4}$)\\
    Case 5 & \textbf{0.755} (0.025)& $10^{-4}$ ($10^{-5}$) & 0.125 (0.295) & 0.163 (0.281) & 0.055 (0.110) & $9\times 10^{-4}$ ($2\times 10^{-4}$)\\
    Case 6 & \textbf{0.600} $(-)$& $10^{-4}$ ($10^{-5}$) & 0.110 (0.270) & 0.143 (0.233) & $-(-)$ & $6\times 10^{-4}$ ($2\times 10^{-4}$)\\
    \hline
    \bottomrule
    \end{tabular}
    \end{adjustbox}
    
\end{table*}

\section{Revisit real datasets for better understanding practical use of SDAMI}

\begin{table*}[h]
    \centering
    \small
    \caption{RMSE over 6 medium-sized real world datasets where $(-)$ indicates infeasibility. The Scale column indicates  scaling factor applied to the response variable for comparison during evaluation.}
    \label{tab: real}
    \begin{adjustbox}{max width=1.0\textwidth}
    \begin{tabular}{lccccccccccc|c}
    \toprule
    & SDAMI & DNN & fSpAM & LASSO & LASSONET& NAM & GAMI-NET & NODE-GAM & NODE-GA$^2$M & \multicolumn{1}{|c}{Scale}\\
    \midrule
    Chip & \textbf{0.244} & 0.927 & 0.753 & 0.276 & 0.904 & 0.967 & 0.495 & 0.455 & 0.546 &\multicolumn{1}{|c}{$\times 1$}\\
    Diabetes & \textbf{0.524} & 0.584 & 0.588 & 0.595 & 0.566 & 0.556 & 0.542 & 0.622 & 0.674 &\multicolumn{1}{|c}{$\times 0.01$}\\
    V1 Cell & \textbf{0.622} & 0.702 & 0.793 & 0.789 & 0.792 & $-$ & 0.734 & 0.772 & 0.739 &\multicolumn{1}{|c}{$\times 1$}\\
    \midrule
    Wine & \textbf{0.672} & 0.702 & 0.771 & 0.745 & 0.703 & 0.712 & 0.701 & 0.721 & 0.698 &\multicolumn{1}{|c}{$\times 1$}\\
    BikeShare & \textbf{0.440} & 0.459 & 1.484 & 1.434 & 0.468 & 1.001 & 0.592 & 1.001 & 0.554 &\multicolumn{1}{|c}{$\times 0.01$}\\
    CA Housing & 0.529 & 0.531 & 0.821 & 0.731 & 0.536 & 0.579 & 0.528 & 0.571 & \textbf{0.503} &\multicolumn{1}{|c}{$\times 1$}\\
    \hline
    \bottomrule
    \end{tabular}
    \end{adjustbox}
    
\end{table*}

\begin{figure}[t!]
    \centering
    \includegraphics[width=14cm]{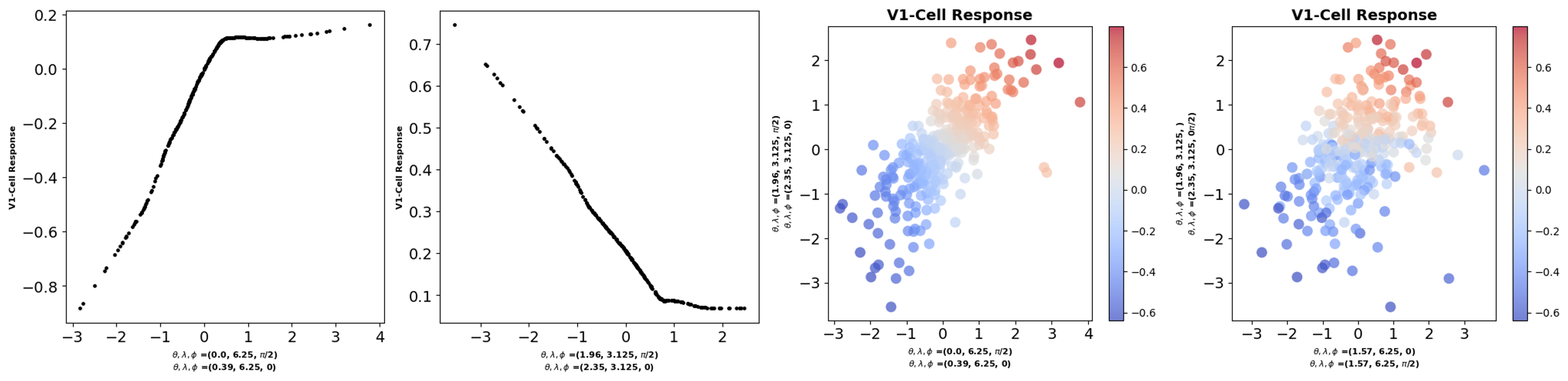}
    \caption{(V1 Cell): Post-hoc analysis of the identified interaction subnetwork) Left two panels: predicted marginal main effects. Right two panels: the estimated response surface for interactions.}
    \label{fig: v1cell}
    \vspace{-5mm}
\end{figure}

The V1 fMRI dataset \cite{kay2008identifying} records voxel responses from human primary visual cortex and provides a prototypical small-$n$, large-$k$ setting, with 300 natural images and 1,800 Gabor-filter features derived from complex-cell processing. Prior studies suggest that interactions among such features are biologically relevant, but modeling them while retaining interpretability remains challenging \cite{kay2008identifying, vu2008nonparametric}. Preprocessing details are given in Appendix~\ref{appendix-v1}. Figure~\ref{fig:img2cp} sketches the predictor generation, and Figure~\ref{fig:cp2rs} shows the SDAMI linkage.

Across the real-data benchmarks in Table~\ref{tab: real}, SDAMI achieves the best or near-best RMSE in most datasets, with particularly strong performance on Chip, V1Cell, and BikeShare. Although NODE-GA2M is competitive on California Housing and Wine, SDAMI remains robust across domains while preserving effect-level interpretability through component visualization.

For the V1 data, Figure~\ref{fig: v1cell} shows the recovered components that suggest heterogeneous marginal trends and non-additive feature combinations. These patterns are descriptive rather than causal, but they indicate that SDAMI can recover structured nonlinear effects that may be useful for downstream scientific interpretation. The primary visual cortex (V1) processes visual information through a hierarchical organization where simple cells respond to oriented edges at specific spatial positions and frequencies, characterized by Gabor filter parameters including orientation angle $\theta$, spatial wavelength $\lambda$, and phase $\phi$. The two leftmost panels show completely different trends. The right two panels show the decrease from initial response to near-zero, suggesting this complex cell is driven by orthogonal orientations that suppress its baseline activity. The right two panels visualize pairwise complex-cell interactions, where both axes represent the response magnitudes of two distinct complex cell. These results indicate that SDAMI can recover interpretable higher-order structure in high-dimensional neural-response data while remaining competitive in prediction.

Together, these results show that SDAMI delivers superior prediction and interpretability with plausible biological alignment in small-\(n\), large-\(k\) regimes, establishing a principled framework for response modeling in neuroscience and other high-dimensional domains. Additional dataset-specific preprocessing, implementation details, and extended real-data analyses are provided in Appendix~\ref{appendix:real-data}.

\section{Conclusion}\label{sec:conclusion}

We presented SDAMI, a structured framework for small-$n$, large-$k$ regression that combines footprint--based screening, grouped structural decomposition, and neural effect estimation. SDAMI is designed to recover interpretable main and interaction structure while remaining competitive in prediction, particularly in interaction dominant regimes. Under the stated assumptions, our theoretical analysis established effect-level structural recovery for the screening and decomposition stages and prediction convergence in probability for the final estimator. Empirically, the method performs strongly on both synthetic and real-data benchmarks, where it provides interpretable component estimates together with competitive predictive accuracy.

{\bf Limitations and future directions.} 
Future work may broaden both evaluation and theory to richer higher-order
interaction structures. Important directions include addressing computational
scalability via SIS-based screening \citep{fan2008sure, fan2011nonparametric},
establishing finite-sample convergence rates for both main and interaction
effects in sparse high-dimensional additive models \citep{gregory2021statistical},
reducing candidate interactions through safe screening \citep{nakagawa2016safe},
and exploring penalized spline activations for computational efficiency
\citep{hung2025deep}. Another limitation is that the footprint principle may weaken under exact
symmetry, particularly for categorical or binary predictors. A possible
extension of the current theory is to incorporate suitable discrete-factor
encodings into the second-stage group-lasso decomposition, such as
sparse-group-lasso-type formulations \citep{simon2013sparse}. Developing the
corresponding theory for mixed predictor settings is an interesting direction
for future work.

\newpage

\bibliography{nips2026}
\bibliographystyle{plainnat}

\newpage
\appendix

{\bf\Large \textsc{Supplementary Material for Sparse Deep Additive Model with Interactions: Enhancing Interpretability and Predictability}}

\appendix
\section{SDAMI Algorithm}\label{alg: sdam}
This section describes the detail of the SDAMI algorithm and how the model fitting works. For reference, the projected update for constrained training. After each optimizer step, we check every main-effect and interaction parameter block and rescale any block whose norm exceeds the prescribed bound. This implements the hard norm constraints in the SDAMI objective directly, rather than replacing them with soft regularization terms.

\begin{algorithm}[H]
\caption{SDAMI Fitting}
\label{alg:sdami}
\begin{algorithmic}[1]
\REQUIRE Data $\{(X_i,Y_i)\}_{i=1}^n$, tuning parameters $\lambda_1,\lambda_2$
\STATE \textbf{Step 1: Effect Footprint Screening (SpAM).}
\begin{itemize}
    \item Fit the sparse additive model
    \[
    Y_i = \sum_{j=1}^{p+q} f_j(X_{ij}) + \epsilon_i
    \]
    using SpAM with penalty $\lambda_1$.
    \item Obtain estimated active set $\widehat{\mathcal{S}} 
    \subseteq \{1,\ldots,p+q\}$ containing both true main effects 
    and footprint variables.
\end{itemize}
\STATE \textbf{Step 2: Decomposition of Active Set (Group Lasso).}
\begin{itemize}
    \item Apply group lasso with orthogonal basis expansion on $\widehat{\mathcal{S}}$.
    \item Decomposition of into $\widehat{\mathcal{M}}$ and $\widehat{\mathcal{I}}$.
    \item Select penalty $\lambda_2$ via cross-validation 
    (with $\lambda_1$ selected by Mallow’s $C_p$).
\end{itemize}
\STATE \textbf{Step 3: SDAMI Model Fitting.}
\begin{itemize}
    \item Fit the constrained deep regression model  using $\widehat{\mathcal{M}}$ and $\widehat{\mathcal{I}}$.
    \item Implement subnetworks in PyTorch, with sparsity 
    imposed via norm-based constraints.
\end{itemize}
\ENSURE Estimated main-effect subnetworks 
$\{\mathrm{NN}^{(j)}\}_{j\in \widehat{\mathcal{M}}}$ and 
interaction subnetworks 
$\{\mathrm{NN}^{(\mathcal{I})}\}_{\mathcal{I}\in \widehat{\mathcal{I}}}$.
\end{algorithmic}
\end{algorithm}

\paragraph{Regularization Parameter Selection.} The regularization parameters $\lambda_1,\lambda_2$ are selected by minimizing the estimated risk and by cross-validation, respectively. The effective degree of freedom is defined as $\text{df}(\lambda)=\underset{j}{\sum}\nu_jI(\|\hat{f}_j\|\neq 0),$ where $\nu_i=\text{trace}(S_j)$ and $S_j$ denotes the smoothing matrix for the $j$-th dimension. The estimate is given by 
\begin{equation*}
    C_p=\frac{1}{n}\sum^n_{i=1}\bigg(Y_i-\sum^p_{j=1}\hat{f}_j(X_j)\bigg)^2+\frac{2\hat{\sigma}^2}{n}\text{df}(\lambda).
\end{equation*}

\section{Proof of Theorem~4.1}\label{App: proofthm41}

\paragraph{Model space.}
Let $\mathcal M \subseteq \{1,\dots,p\}$ be the index set for additive
(univariate) components, and let 
$\mathcal I \subseteq \{1,\dots,q\}$ be the index set for the 
multivariate interaction component.
For $j \in \mathcal M$, let 
$\mathrm{NN}^{(j)}(x_j;\theta_j)$ denote the univariate neural network,
and let 
$\mathrm{NN}^{(\mathcal I)}(x_{\mathcal I};\theta_{\mathcal I})$
denote the multivariate neural network on coordinates $\mathcal I$.

We define the hypothesis classes induced by the first-layer constraints
\begin{align*}
    &\|W^{(1)}_{\mathcal M,j}(\theta_j)\|_{\infty}
\;\le\; \kappa_{\mathcal M}\, \|f_j\|,
\quad j\in\mathcal M,\\
    & \|W^{(1)}_{\mathcal I}(\theta_{\mathcal I})\|_{\infty}
\;\le\; \kappa_{\mathcal I}\, \|f_{\mathcal I}\|.
\end{align*}

The admissible univariate and multivariate function classes are
\[
\mathcal H_j
=
\left\{
f_j(\cdot)
=
\mathrm{NN}^{(j)}(\cdot;\theta_j)
:
\|W^{(1)}_{\mathcal M,j}(\theta_j)\|_{\infty}
\le 
\kappa_{\mathcal M}\,\|f_j\|
\right\},
\qquad j\in\mathcal M,
\]
\[
\mathcal G_{\mathcal I}
=
\left\{
g(\cdot)
=
\mathrm{NN}^{(\mathcal I)}(\cdot;\theta_{\mathcal I})
:
\|W^{(1)}_{\mathcal I}(\theta_{\mathcal I})\|_{\infty}
\le 
\kappa_{\mathcal I}\,\|g\|
\right\}.
\]

\begin{definition}[Model space of the structured neural network]
The functional model space associated with the neural network estimator
in \eqref{eq:sdami-obj}--\eqref{eq:sdami-con} is
\[
\mathcal F_{\mathrm{NN}}(\mathcal M,\mathcal I)
=
\left\{
f(x)
=
\sum_{j\in\mathcal M} f_j(x_j)
\;+\;
g(x_{\mathcal I})
\;:\;
f_j \in \mathcal H_j,\;
g \in \mathcal G_{\mathcal I}
\right\}.
\]
\end{definition}

If sparsity over the sets $\mathcal M$ and $\mathcal I$ is desired, the
overall sparse model space is
\[
\mathcal F_{\mathrm{NN}}(s_1,s_2)
=
\bigcup_{\substack{
\mathcal M \subseteq \{1,\dots,p\},\; |\mathcal M| \le s_1 \\
\mathcal I \subseteq \{1,\dots,q\},\; |\mathcal I| \le s_2}}
\mathcal F_{\mathrm{NN}}(\mathcal M,\mathcal I).
\]
This section provides the detailed proof of Theorem~4.1, which establishes the equivalence between vanishing effect footprints and the disappearance of the first–order projection in the Hoeffding–Sobol decomposition. The result clarifies when a variable contributes only through higher–order interactions and thus leaves no detectable marginal footprint.

We begin with the Hoeffding–Sobol decomposition. Let $f(\mathbf X_{\mathcal I})$ be a centered function, i.e., $\mathbb E[f(\mathbf X_{\mathcal I})]=0$. Then $f$ admits the unique expansion
\[
f(\mathbf X_{\mathcal I}) 
= f_{\{j\}}(X_j) 
+ \sum_{\substack{S\subseteq \mathcal I,\, j\in S,\,|S|\ge 2}} f_S(\mathbf X_S)
+ \sum_{\substack{S\subseteq \mathcal I,\, j\notin S,\,|S|\ge 1}} f_S(\mathbf X_S),
\]
where the components $f_S$ are mutually orthogonal in $L^2$, each has mean zero, and $f_{\{j\}}(X_j)$ represents the unique first–order contribution of $X_j$. The remaining terms correspond either to higher–order interactions involving $X_j$ or to effects of variables not involving $X_j$.

Conditional expectation with respect to $X_j$ is the orthogonal projection of $f$ onto the subspace of $L^2$ functions of $X_j$, as ensured by the Doob–Dynkin lemma and the Hilbert projection theorem. Hence the footprint $m_j(X_j)=\mathbb E[f(\mathbf X_{\mathcal I})\mid X_j]$ coincides with this projection. By uniqueness of the Hoeffding–Sobol components, this projection is exactly $f_{\{j\}}(X_j)$.  
The two directions now follow. If $f_{\{j\}}$ vanishes identically, then conditioning the decomposition on $X_j$ eliminates all other terms: for $S$ not containing $j$, centeredness of $f_S$ implies $\mathbb E[f_S(\mathbf X_S)\mid X_j]=0$, while for $S$ containing $j$ with $|S|\ge 2$, orthogonality ensures $\mathbb E[f_S(\mathbf X_S)\mid X_j]=0$. Thus $m_j(X_j)=0$, which is constant, so $X_j$ leaves no footprint. Conversely, if $m_j(X_j)$ is constant almost surely, then $\mathbb E[f(\mathbf X_{\mathcal I})\mid X_j]$ is identically zero because $f$ is centered. Since this conditional expectation is the projection of $f$ onto the space of functions of $X_j$, it follows that $f_{\{j\}}(X_j)\equiv 0$.  

Therefore, the footprint $m_j(x)$ is constant if and only if the first–order projection $f_{\{j\}}(X_j)$ vanishes. In this case, the variable $X_j$ contributes only through higher–order interactions, and its marginal influence disappears in expectation, thereby proving Theorem~4.1.

\section{Conditions and Proof of Theorem 4.2}\label{App: proofthm42}

This section establishes the effect-level selection consistency of SDAMI. We begin by introducing the technical assumptions that govern the noise, design structure, signal strength, and basis expansion. These conditions provide the foundation for analyzing the group-lasso estimator used in SDAMI and for verifying the primal–dual witness construction that guarantees selection consistency.

\paragraph{Assumptions (Conditions for effect-level selection).}
\leavevmode
\begin{enumerate}
\item[(A1)] (\emph{Noise}) The errors $\epsilon_i$ in the true function (1) of the main paper are sub-Gaussian with mean zero and variance proxy $\sigma^2$.

\item[(A2)] (\emph{Within-group orthonormality}) For each main effect $j$, 
\[
\frac{1}{n}\Phi_j^\top \Phi_j = I,
\]
and for the interaction block $\Phi_{\mathcal I}$,
\[
\frac{1}{n}\Phi_{\mathcal I}^\top \Phi_{\mathcal I} = I, 
\qquad 
\frac{1}{n}\Phi_{\mathcal I}^\top \Phi_j = 0 \ \ (j\in \mathcal I).
\]

\item[(A3)] (\emph{Block coherence}) For $g\neq g'$,
\[
\Big\|\tfrac{1}{n}X_g^\top X_{g'}\Big\|_{\mathrm{op}} \le \mu < 1,
\]
where $X_g$ denotes the block of design columns for group $g$.

\item[(A4)] (\emph{Restricted eigenvalue}) The Gram matrix on the active set 
\[
\Sigma_{A^\star A^\star}=\tfrac{1}{n}X_{A^\star}^\top X_{A^\star}, 
\qquad A^\star=\mathcal M\cup\{\mathcal I\},
\]
satisfies $\lambda_{\min}(\Sigma_{A^\star A^\star}) \ge \kappa_{\min} > 0$ and the method for constructing the Gram matrix is defined in assumption (A7).

\item[(A5)] (\emph{Irrepresentability}) There exists $\eta>0$ such that
\[
\|\Sigma_{A^{\star c}A^\star}\Sigma_{A^\star A^\star}^{-1}\|_{2,\infty}\le 1-\eta.
\]

\item[(A6)] (\emph{Signal strength}) With group weights $w_g\in[1,C_w]$ and tuning parameter 
$\lambda_n \asymp \sigma \sqrt{\tfrac{\log G}{n}}$ (where $G$ is the number of candidate groups),
\[
\min_{j\in \mathcal M}\|f_j\| \ge c_0 \lambda_n,
\qquad 
\|f_{\mathcal I}\|\ge c_0 \lambda_n \ \ \text{if the interaction is present},
\]
for some $c_0 > 2/\eta$.

\item[(A7)] (\emph{Finite orthonormal basis representation}) Each function $f_j$ and the interaction 
$f_{\mathcal I}$ is represented in an orthonormal basis expansion of finite dimension (at most quadratic order), with corresponding design blocks $\Phi_j$ and $\Phi_{\mathcal I}$.
\end{enumerate}

Having specified the assumptions, we now turn to the proof. The role of (A7) is to provide a finite orthonormal basis representation of all effects, which allows us to formulate the regression problem as a finite-dimensional block group-lasso. Assumptions (A1)–(A6) then control the noise, dependence, eigenstructure, and signal strength needed to verify that the primal–dual witness construction recovers the correct support with probability tending to one. 

By (A7), each main effect $f_j$ and the interaction $f_{\mathcal I}$ admits a finite-dimensional orthonormal basis representation, say
\[
f_j(x_j) = \Phi_j(x_j)^\top \beta_j, 
\qquad 
f_{\mathcal I}(\mathbf X_{\mathcal I}) = \Phi_{\mathcal I}(\mathbf X_{\mathcal I})^\top \gamma,
\]
where $\Phi_j\in\mathbb R^{n\times m_j}$ and $\Phi_{\mathcal I}\in\mathbb R^{n\times m_{\mathcal I}}$ collect the basis evaluations across $n$ samples. Stacking these blocks gives the design matrix
\[
X = [X_1,\dots,X_k,X_{\mathcal I}], \qquad X_j:=\Phi_j,\quad X_{\mathcal I}:=\Phi_{\mathcal I},
\]
with block coefficient vector $\theta=(\beta_1,\dots,\beta_k,\gamma)$. The true active set is $A^\star=\mathcal M\cup\{\mathcal I: f_{\mathcal I}\neq 0\}$ and the inactive set is $I^\star=\mathcal G\setminus A^\star$, where $\mathcal G$ denotes all candidate groups.

The SDAMI estimator solves the block group-lasso problem
\[
\widehat\theta\in \arg\min_{\theta}\ \frac{1}{2n}\|y-X\theta\|_2^2 + \lambda_n\sum_{g\in\mathcal G} w_g\|\theta_g\|_2,
\]
with tuning parameter $\lambda_n \asymp \sigma \sqrt{\tfrac{\log G}{n}}$ and group weights $w_g\in[1,C_w]$. The associated KKT conditions are
\[
\frac{1}{n}X_g^\top(y-X\widehat\theta) = \lambda_n w_g \widehat z_g,\qquad \|\widehat z_g\|_2\le 1,\qquad 
\widehat z_g = \frac{\widehat\theta_g}{\|\widehat\theta_g\|_2} \ \ \text{if }\widehat\theta_g\neq 0.
\]

Assumption (A1) ensures that the error vector $\varepsilon$ is sub-Gaussian. By a union bound over all blocks and coordinates, with probability $1-o(1)$ the event
\[
\max_{g\in\mathcal G}\frac1n \|X_g^\top \varepsilon\|_2 \le \tfrac12 \lambda_n w_g
\]
holds, providing high-probability control of noise terms in the KKT system. Assumptions (A2) and (A3) impose within-block orthonormality and block coherence, ensuring that $\Sigma = X^\top X/n$ has bounded eigenvalues and limited inter-block correlations. Assumption (A4) states a restricted eigenvalue condition, which guarantees that for any deviation vector $\Delta_{A^\star}$ supported on the active set,
\[
\frac{1}{n}\|X_{A^\star}\Delta_{A^\star}\|_2^2 \ge \kappa_{\min}\|\Delta_{A^\star}\|_2^2.
\]
Assumption (A5) provides the irrepresentability condition, ensuring that inactive blocks cannot mimic active ones in the dual constraints. Finally, assumption (A6) requires minimal signal strength $\|f_g\|\ge c_0\lambda_n$ on all active blocks, so that true coefficients dominate the estimation error.

Under these conditions, the restricted problem on $A^\star$ yields an estimator $\widehat\theta_{A^\star}$ with error bound
\[
\|\widehat\theta_{A^\star}-\theta^\star_{A^\star}\|_2 \le \frac{3\lambda_n}{\kappa_{\min}}\Big(\sum_{g\in A^\star}w_g^2\Big)^{1/2}.
\]
Because $c_0>2/\eta$, this error is asymptotically smaller than the true signal size, ensuring $\widehat\theta_g\neq 0$ for all $g\in A^\star$. Thus, no active block is missed. For inactive groups, the dual feasibility condition requires $\frac1n\|X_g^\top(y-X_{A^\star}\widehat\theta_{A^\star})\|_2 < \lambda_n w_g$. The residual expands as $\widehat r=\varepsilon-X_{A^\star}(\widehat\theta_{A^\star}-\theta^\star_{A^\star})$. The first term is controlled by (A1), while the second is bounded by (A3) and (A5) together with the error rate above. Consequently, inactive groups satisfy strict dual feasibility, forcing $\widehat\theta_g=0$ for all $g\in I^\star$. This establishes absence of false positives.

For the interaction, if $f_{\mathcal I}=0$, then $\mathcal I\in I^\star$ and the dual condition implies $\widehat f_{\mathcal I}=0$. If $f_{\mathcal I}\neq 0$, then $\mathcal I\in A^\star$ and the signal strength bound ensures $\widehat f_{\mathcal I}\neq 0$. Combining all pieces, with probability tending to one we have
\[
\{j:\widehat f_j\neq 0\}=\mathcal M, 
\qquad
\widehat f_{\mathcal I}\neq 0 \ \Leftrightarrow \ f_{\mathcal I}\neq 0,
\]
which proves the effect-level selection consistency of SDAMI as stated in Theorem~\ref{thm:sdami-consistency}.

\section{Conditions and proof of Theorem 4.3}\label{App: proofthm43}

To ground the proof, we first specify the SDAMI function class and estimator used throughout.

\paragraph{Model class of SDAMI.}
Let $A \subseteq \{1,\dots,p\}$ index a subset of active main effects and interactions. For each main effect $j \in A_{\mathrm{main}}$ and interaction $\mathcal I \in A_{\mathrm{int}}$, let $\mathcal N_{L,W,B}$ denote the class of feedforward ReLU subnetworks of depth $L$ and maximal width $W$ whose parameters satisfy a norm constraint (e.g., path norm, spectral norm, or $\ell_2$ decay) bounded by $B$. For a growth schedule $(L_n,W_n,B_n)$, define the SDAMI sieve over $A$ by
\[
\mathcal F_n^{\mathrm{SDAMI}}(A)
=\Bigg\{\, f(x)=\sum_{j\in A_{\mathrm{main}}} g_j(x_j) \;+\;  h_{\mathcal I}(x_{\mathcal I})
\;:\; g_j\in \mathcal N_{L_n,W_n,B_n},\ \ h_{\mathcal I}\in \mathcal N_{L_n,W_n,B_n} \Bigg\}.
\]
Thus SDAMI is an additive model with interactions, where each component is realized by a subnetwork from $\mathcal N_{L_n,W_n,B_n}$ restricted to its own argument(s).

\paragraph{Assumptions.}
\begin{enumerate}
\item[(B1)] \emph{Sampling, noise, and approximation.} 
The data $(\mathbf X_i,Y_i)_{i=1}^n$ are i.i.d. from model (1) in the main content with $\epsilon_{i}$ satisfying  $E[\epsilon_i] = 0$ and $\mathrm{Var}(\epsilon_i) = \sigma^2 < \infty$. The covariates $\mathbf X$ have either bounded support or sub-Gaussian tails, and the true regression function $f^\star \in L_2(P_{\mathbf X})$ lies in the $L_2(P_{\mathbf X})$-closure of the sieve
\[
\bigcup_{n=1}^\infty \mathcal F_n^{\mathrm{SDAMI}}(A),
\]
so that for any $\varepsilon > 0$ there exists $n$ and $f \in \mathcal F_n^{\mathrm{SDAMI}}(A)$ with 
$\|f - f^\star\|_{L_2(P_{\mathbf X})} \le \varepsilon$.

\item[(B2)] \emph{Effect-level selection consistency (SDAMI).}
Let $A^\star$ be the true set of active main effects and interactions. Then
$\mathbb P(\widehat A_n = A^\star)\to1$.

\item[(B3)] \emph{Approximation (DNN sieve over true inputs).}
For the restricted DNN class $\mathcal F_n^{\mathrm{DNN}}(A^\star)$ with schedule $(L_n,W_n,B_n)$,
the sieve approximation error vanishes:
\[
\alpha_n \;:=\; \inf_{f\in \mathcal F_n^{\mathrm{DNN}}(A^\star)} P\!\big[(f-f^\star_{A^\star})^2\big]\ \longrightarrow\ 0.
\]

\item[(B4)] \emph{Empirical risk minimization up to tolerance.}
The trained $\widehat f_n\in \mathcal F_n^{\mathrm{SDAMI}}(\widehat A_n)$ satisfies
\[
P_n\!\big[(\widehat f_n - Y)^2\big] \ \le\ 
\inf_{f\in\mathcal F_n^{\mathrm{SDAMI}}(\widehat A_n)} P_n\!\big[(f-Y)^2\big]\ +\ \delta_n,
\qquad \delta_n\downarrow 0.
\]

\item[(B5)] \emph{Capacity control and uniform generalization.}
The norm constraint $B_n$ (and/or width $W_n$) ensures a vanishing complexity for squared loss:
\[
\mathfrak R_n(\mathcal L_n)\ =\ o(1),
\qquad 
\mathcal L_n:=\{(f-g)^2: f\in \mathcal F_n^{\mathrm{SDAMI}}(A),\, g\in \mathcal F_n^{\mathrm{SDAMI}}(A),\, A\subseteq\{1,\dots,p\}\},
\]
so that
\[
\sup_{h\in \mathcal L_n}\big|(P-P_n)h\big|\ =\ o_p(1).
\]

\item[(B6)] (\emph{Measurability and uniform $L_2$ envelope})
Each $f\in \mathcal F_n^{\mathrm{SDAMI}}(A)$ is measurable, and there exists a constant $M<\infty$ (independent of $n$, $A$, and $f$) such that
\[
\sup_{A\subseteq[p]}\ \sup_{f\in \mathcal F_n^{\mathrm{SDAMI}}(A)} P f^2 \ \le\ M .
\]
In particular, for the data–dependent active set $\widehat A_n$, the trained $\widehat f_n\in \mathcal F_n^{\mathrm{SDAMI}}(\widehat A_n)$ is measurable and satisfies $P\widehat f_n^2\le M$ almost surely. Hence $\{P\ell(\widehat f_n)\}_n$ is uniformly integrable. 

\end{enumerate}

With the SDAMI sieve $\mathcal F_n^{\mathrm{SDAMI}}(\widehat A_n)$ specified and assumptions (B1)–(B6) in place, we now prove Theorem~\ref{thm:sdami-prob-conv} by analyzing the empirical minimizer within this class and translating vanishing risk into prediction convergence.

Let $P$ denote expectation with respect to $P_{\mathbf X}$ and $P_n$ the empirical average over the training inputs. Write the squared excess prediction loss as $\ell(f):=(f-f^\star)^2$. By the selection consistency of SDAMI (B2), $\mathbb P(\widehat A_n=A^\star)\to1$, so it suffices to analyze $\widehat f_n\in\mathcal F_n^{\mathrm{SDAMI}}(A^\star)$ and the conclusions will then hold unconditionally. Using the empirical-to-population decomposition,
\[
P\ell(\widehat f_n)\;=\;P_n\ell(\widehat f_n)\;+\;(P-P_n)\ell(\widehat f_n).
\]
To control $P_n\ell(\widehat f_n)$, expand the empirical squared loss around $Y=f^\star+\epsilon$:
\[
P_n\big[(\widehat f_n-Y)^2\big]
= P_n\ell(\widehat f_n)\;+\;P_n[\epsilon^2]\;+\;2\,P_n\!\big[(f^\star-\widehat f_n)\epsilon\big].
\]
By the empirical optimality up to tolerance (B4), for any $f\in\mathcal F_n^{\mathrm{SDAMI}}(A^\star)$,
\[
P_n\ell(\widehat f_n)
\;\le\;
P_n\ell(f)\;+\;2\,\Big|P_n\!\big[(f^\star-\widehat f_n)\epsilon\big]\Big|
\;+\;2\,\Big|P_n\!\big[(f^\star-f)\epsilon\big]\Big|
\;+\;\delta_n.
\]
The noise is centered with bounded conditional variance (B1) and the SDAMI sieve is capacity–controlled (B5), hence the stochastic inner products above are $o_p(1)$ uniformly over $f\in\mathcal F_n^{\mathrm{SDAMI}}(A^\star)$ by standard symmetrization/contraction bounds for squared loss. Taking the infimum over $f\in\mathcal F_n^{\mathrm{SDAMI}}(A^\star)$ yields
\[
P_n\ell(\widehat f_n)
\;\le\;
\inf_{f\in\mathcal F_n^{\mathrm{SDAMI}}(A^\star)} P_n\ell(f)
\;+\;o_p(1)\;+\;\delta_n.
\]
Adding and subtracting population risks and invoking the uniform generalization bound for squared loss from (B5),
\[
P\ell(\widehat f_n)
\;\le\;
\inf_{f\in\mathcal F_n^{\mathrm{SDAMI}}(A^\star)} P\ell(f)
\;+\;o_p(1)\;+\;\delta_n.
\]
By the approximation property of the SDAMI sieve on the true inputs (B3), the approximation error
$\alpha_n:=\inf_{f\in\mathcal F_n^{\mathrm{SDAMI}}(A^\star)} P\ell(f)$ satisfies $\alpha_n\to0$; therefore
\begin{equation}\label{eq:pop-risk-vanish-sdami}
P\ell(\widehat f_n)\ \xrightarrow{p}\ 0.
\end{equation}
To convert result (\ref{eq:pop-risk-vanish-sdami}) into prediction convergence, note the inequality
\[
\mathbf 1\!\left\{\big|\widehat f_n(\mathbf X)-f^\star(\mathbf X)\big|\ge\varepsilon\right\}
\;\le\; \frac{\ell(\widehat f_n)(\mathbf X)}{\varepsilon^2},
\qquad \varepsilon>0.
\]
Taking expectation over $\mathbf X$ and then over the training sample gives
\[
\mathbb P\!\left(\big|\widehat f_n(\mathbf X)-f^\star(\mathbf X)\big|\ge\varepsilon\right)
\;\le\; \frac{\mathbb E\big[P\ell(\widehat f_n)\big]}{\varepsilon^2}.
\]
The sieve’s norm constraints together with (B6) imply a square–integrable envelope on $\mathcal F_n^{\mathrm{SDAMI}}(A^\star)$, hence $\{P\ell(\widehat f_n)\}_n$ is uniformly integrable; combined with result (\ref{eq:pop-risk-vanish-sdami}) this yields $\mathbb E[P\ell(\widehat f_n)]\to0$. Consequently,
\[
\mathbb P\!\left(\big|\widehat f_n(\mathbf X)-f^\star(\mathbf X)\big|\ge\varepsilon\right)\ \longrightarrow\ 0
\quad\text{for every fixed }\varepsilon>0,
\]
i.e., $\widehat f_n(\mathbf X)\xrightarrow{p} f^\star(\mathbf X)$ at the design distribution $P_{\mathbf X}$. \qed

\section{Proof of Theorem~\ref{thm:sdami-finite-sample}}
\label{App: proofthm44}

This section proves the finite-sample prediction guarantee stated in
Theorem~\ref{thm:sdami-finite-sample}. The proof has two parts. Part
\textnormal{(i)} proves the finite-sample prediction bound under the true
active effect set \(A^\star\), which corresponds to the event that SDAMI
recovers the correct structural class. Part \textnormal{(ii)} proves the
corresponding prediction bound for the actually selected active effect set
\(\widehat A_n\), which is the data-dependent class used by Stage 3.

\paragraph{Part \textnormal{(i)}: Prediction bound under the true active effect set.}

Let
\[
\mathcal E_A=\{\widehat A_n=A^\star\}
\]
denote the event that SDAMI recovers the correct active effect set. By
definition,
\[
\mathbb P(\mathcal E_A)
=
1-\mathbb P(\widehat A_n\neq A^\star).
\]
Under Assumptions \textnormal{(A1)--(A7)}, the screening step and the
group-lasso decomposition step satisfy the finite-sample version of the
effect-level recovery argument. In particular, the sub-Gaussian noise
condition, restricted eigenvalue condition, irrepresentability condition,
and signal-strength condition imply that, on \(\mathcal E_A\), the active
main-effect and interaction blocks are retained while inactive blocks are
excluded. Therefore, on \(\mathcal E_A\), Stage 3 is fitted over the true
structural SDAMI class $\mathcal F_n^{\mathrm{SDAMI}}(A^\star).$

Next, let
\[
Z_n(A^\star)
=
\sup_{f\in\mathcal F_n^{\mathrm{SDAMI}}(A^\star)}
\left|
P\!\left[\{Y-f(\mathbf X)\}^2\right]
-
P_n\!\left[\{Y-f(\mathbf X)\}^2\right]
\right|.
\]
For \(t>0\), define \(R_n(A^\star,t)\) as a finite-sample complexity
radius satisfying
\[
\mathbb P\left\{
Z_n(A^\star)\le R_n(A^\star,t)
\right\}
\ge 1-e^{-t}.
\]
Under the capacity control and uniform envelope conditions in
\textnormal{(B5)} and \textnormal{(B6)}, such a radius is available from
standard empirical-process concentration bounds for the squared-loss class
induced by \(\mathcal F_n^{\mathrm{SDAMI}}(A^\star)\). Then, define
\[
\mathcal E_G(t)
=
 \left\{
Z_n(A^\star)\le R_n(A^\star,t)
\right\}
\]
Then, by the definition of \(R_n(A^\star,t)\),
\[
\mathbb P\{\mathcal E_G(t)\}\ge 1-e^{-t}.
\]
Since
\[
\mathcal E_A^c=\{\widehat A_n\neq A^\star\}
\quad\text{and}\quad
\mathbb P\{\mathcal E_G(t)^c\}\le e^{-t},
\]
the union bound gives
\[
\mathbb P\{\mathcal E_A\cap\mathcal E_G(t)\}
\ge
1-\mathbb P(\widehat A_n\neq A^\star)-e^{-t}.
\]

We now work on the event \(\mathcal E_A\cap\mathcal E_G(t)\). Since
\(\mathcal E_A\) holds, \(\widehat A_n=A^\star\), and the Stage 3 estimator
\(\widehat f_n\) belongs to
\(\mathcal F_n^{\mathrm{SDAMI}}(A^\star)\). By the empirical
near-optimality condition in \textnormal{(B4)}, for any
\(f\in\mathcal F_n^{\mathrm{SDAMI}}(A^\star)\),
\[
P_n(Y-\widehat f_n)^2
\le
P_n(Y-f)^2+\delta_n,
\]
where \(\delta_n\) is the optimization tolerance.

Using the generalization event \(\mathcal E_G(t)\), we have
\[
P(Y-\widehat f_n)^2
\le
P_n(Y-\widehat f_n)^2+R_n(A^\star,t).
\]
Combining this inequality with the empirical near-optimality inequality gives
\[
P(Y-\widehat f_n)^2
\le
P_n(Y-f)^2+\delta_n+R_n(A^\star,t).
\]
Applying \(\mathcal E_G(t)\) again to the fixed function \(f\), we obtain
\[
P_n(Y-f)^2
\le
P(Y-f)^2+R_n(A^\star,t).
\]
Therefore,
\[
P(Y-\widehat f_n)^2
\le
P(Y-f)^2+2R_n(A^\star,t)+\delta_n.
\]
Since the above inequality holds for every
\(f\in\mathcal F_n^{\mathrm{SDAMI}}(A^\star)\), taking the infimum over
this class yields
\[
P(Y-\widehat f_n)^2
\le
\inf_{f\in\mathcal F_n^{\mathrm{SDAMI}}(A^\star)}
P(Y-f)^2
+
2R_n(A^\star,t)
+
\delta_n.
\]

We now translate this risk bound into a prediction-error bound. Since
\(f^\star(\mathbf X)=\mathbb E(Y\mid \mathbf X)\), the squared-loss
excess-risk identity gives
\[
P(Y-f)^2-P(Y-f^\star)^2
=
P\!\left[
\{f(\mathbf X)-f^\star(\mathbf X)\}^2
\right].
\]
Applying this identity to \(f=\widehat f_n\) and to the comparator inside
the infimum gives
\[
P\!\left[
\{\widehat f_n(\mathbf X)-f^\star(\mathbf X)\}^2
\right]
\le
\inf_{f\in\mathcal F_n^{\mathrm{SDAMI}}(A^\star)}
P\!\left[
\{f(\mathbf X)-f^\star(\mathbf X)\}^2
\right]
+
2R_n(A^\star,t)
+
\delta_n.
\]

Finally, conditional on the training sample \(\mathcal D_n\), Markov's
inequality implies that, for every \(\varepsilon>0\),
\[
\mathbb P_{\mathbf X}
\left(
|\widehat f_n(\mathbf X)-f^\star(\mathbf X)|\ge \varepsilon
\,\middle|\,\mathcal D_n
\right)
\le
\frac{
E\!\left[
\left(\widehat f_n(\mathbf X)-f^\star(\mathbf X)\right)^2\mid\mathcal{D}_n
\right]
}{
\varepsilon^2
}
=
\frac{
P\!\left[
\{\widehat f_n(\mathbf X)-f^\star(\mathbf X)\}^2
\right]
}{
\varepsilon^2
}.
\]
Substituting the preceding inequality into the right-hand side yields
\[
\mathbb P_{\mathbf X}
\left(
|\widehat f_n(\mathbf X)-f^\star(\mathbf X)|\ge \varepsilon
\,\middle|\,\mathcal D_n
\right)
\le
\frac{
\inf_{f\in\mathcal F_n^{\mathrm{SDAMI}}(A^\star)}
P\!\left[
\{f(\mathbf X)-f^\star(\mathbf X)\}^2
\right]
+
2R_n(A^\star,t)
+
\delta_n
}{
\varepsilon^2
}.
\]
Since this bound holds on \(\mathcal E_A\cap\mathcal E_G(t)\), whose
probability is at least
\[
1-\mathbb P(\widehat A_n\neq A^\star)-e^{-t},
\]
Part \textnormal{(i)} follows.

\paragraph{Part \textnormal{(ii)}: Prediction bound under the selected active effect set.}

We next prove the finite-sample prediction bound for the actually selected
active effect set \(\widehat A_n\). This part does not require a uniform
generalization bound over all possible selected classes. Instead, it uses
the realized generalization error of the selected class produced by Stages
1--2.

For the selected SDAMI class
\[
\mathcal F_n^{\mathrm{SDAMI}}(\widehat A_n),
\]
define the realized selected-class generalization error
\[
R_n(\widehat A_n)
=
\sup_{f\in\mathcal F_n^{\mathrm{SDAMI}}(\widehat A_n)}
\left|
P(Y-f)^2-P_n(Y-f)^2
\right|.
\]
By construction, Stage 3 fits the estimator
\[
\widehat f_n\in\mathcal F_n^{\mathrm{SDAMI}}(\widehat A_n).
\]
By the empirical near-optimality condition in \textnormal{(B4)}, now applied
to the selected class, for any
\(f\in\mathcal F_n^{\mathrm{SDAMI}}(\widehat A_n)\),
\[
P_n(Y-\widehat f_n)^2
\le
P_n(Y-f)^2+\delta_n.
\]

Using the definition of \(R_n(\widehat A_n)\), we have
\[
P(Y-\widehat f_n)^2
\le
P_n(Y-\widehat f_n)^2+R_n(\widehat A_n).
\]
Combining this with the empirical near-optimality inequality gives
\[
P(Y-\widehat f_n)^2
\le
P_n(Y-f)^2+\delta_n+R_n(\widehat A_n).
\]
Again, by the definition of \(R_n(\widehat A_n)\),
\[
P_n(Y-f)^2
\le
P(Y-f)^2+R_n(\widehat A_n).
\]
Therefore,
\[
P(Y-\widehat f_n)^2
\le
P(Y-f)^2+2R_n(\widehat A_n)+\delta_n.
\]
Since this inequality holds for every
\(f\in\mathcal F_n^{\mathrm{SDAMI}}(\widehat A_n)\), taking the infimum over
the selected class yields
\[
P(Y-\widehat f_n)^2
\le
\inf_{f\in\mathcal F_n^{\mathrm{SDAMI}}(\widehat A_n)}
P(Y-f)^2
+
2R_n(\widehat A_n)
+
\delta_n.
\]

Using again the squared-loss excess-risk identity,
\[
P(Y-f)^2-P(Y-f^\star)^2
=
P\!\left[
\{f(\mathbf X)-f^\star(\mathbf X)\}^2
\right],
\]
we obtain
\[
P\!\left[
\{\widehat f_n(\mathbf X)-f^\star(\mathbf X)\}^2
\right]
\le
\inf_{f\in\mathcal F_n^{\mathrm{SDAMI}}(\widehat A_n)}
P\!\left[
\{f(\mathbf X)-f^\star(\mathbf X)\}^2
\right]
+
2R_n(\widehat A_n)
+
\delta_n.
\]

Finally, conditional on the training sample \(\mathcal D_n\), Markov's
inequality gives, for every \(\varepsilon>0\),
\[
\mathbb P_{\mathbf X}
\left(
|\widehat f_n(\mathbf X)-f^\star(\mathbf X)|\ge \varepsilon
\,\middle|\,\mathcal D_n
\right)
\le
\frac{
E\!\left[
\left(\widehat f_n(\mathbf X)-f^\star(\mathbf X)\right)^2\mid\mathcal{D}_n
\right]
}{
\varepsilon^2
}=
\frac{
P\!\left[
\{\widehat f_n(\mathbf X)-f^\star(\mathbf X)\}^2
\right]
}{
\varepsilon^2
}.
\]
Substituting the selected-class risk bound into the right-hand side yields
\[
\mathbb P_{\mathbf X}
\left(
|\widehat f_n(\mathbf X)-f^\star(\mathbf X)|\ge \varepsilon
\,\middle|\,\mathcal D_n
\right)
\le
\frac{
\inf_{f\in\mathcal F_n^{\mathrm{SDAMI}}(\widehat A_n)}
P\!\left[
\{f(\mathbf X)-f^\star(\mathbf X)\}^2
\right]
+
2R_n(\widehat A_n)
+
\delta_n
}{
\varepsilon^2
}.
\]
This proves Part \textnormal{(ii)}. Combining Parts \textnormal{(i)} and \textnormal{(ii)} completes the proof
of Theorem~\ref{thm:sdami-finite-sample}.

\section{Supplementary material for data generating process and hyperparameters selection}\label{appendix:hyperparameter}
In order to tune the hyperparameters, we performed a random stratified split of full training data into train set (80\%), validation set (10\%), and testing set (10\%) for all datasets. For datasets compiled from small-sized sparse settings (Chips, Diabetes, V1-cell), and medium-sized (Wine Quality, Bikeshare, and California Housing), we perform 5-fold cross validation for 5 different test splits. In addition, we summarize the details of cross validation on architecture selection, additional experiment results, and the visualization of either main effects or interactions effects from the numerical studies.
\subsection{Hyperparameter settings}\label{appendix-numerical-cv}
\paragraph{SDAMIs and DNNs}
Before building the neural network, the SDAMI's three-stage procedure requires careful tuning of regularization parameters. For the SpAM Screening, the $\lambda_1$ penalty is selected via Mallows $C_p$ where we set the basis dimension to 8. Subsequently, the $\lambda_2$ penalty is selected via 5-fold cross-validation with convergence tolerance is $1e-4$. However, we have to design appropriate vector for $\lambda_1$ and use $C_p$ value as selection criteria to determine the optimal $\lambda_1$. We tune the penalty term in the three-stage procedure for each task in the continuous bandwidths and summarize in Table~\ref{tab:lambda}.

\begin{table}[]
    \centering
    \caption{Continuous Bandwidths for different task in the three-stage procedure. $\lambda_1$ is selected via Mallow's $C_p$ and $\lambda_2$ is selected via 5-folds cross-validation.}
    \label{tab:lambda}
    \begin{tabular}{ccccc}
    \hline
     & Numerical Studies & Chip & Diabetes & V1 Cell  \\ \hline
     $\lambda_1$ & [0.01, 5) & [0.01, 1.5) & [1, 10) & [0.001, 0.03) \\ \hline
     $\lambda_2$ &logspace[-3, 1)&logspace[-3, 1)&logspace[-3, 1)&logspace[-3, 1)\\ \hline\midrule
    \end{tabular} 
    \begin{tabular}{cccc}
    \hline
     & Wine & Bikeshare & CA Housing \\ \hline
     $\lambda_1$         &   [0.1, 10)   & [0.01, 2.5)    &  [0.001, 4.5)      \\ \hline
     $\lambda_2$ &logspace[-1, 1)&  logspace[0.6, 1)&logspace[-1, 1)\\ \hline\midrule
    \end{tabular}
    
\end{table}

To determine the optimal neural network architecture for SDAMI and DNN baselines for numerical studies and small-sized dataset, we perform 5-fold cross-validation over three candidate configurations for each. Each configuration specifies the number and width of hidden layers in the subnetworks. We summarize the result of cross validation on configuration selection for SDAMI and DNN in Table \ref{tab: n300config}, and the hyperparameter specification in Table~\ref{tab:network}.

\begin{table}[t]
    \centering
    \caption{(RMSE) Performance of SDAMs and DNNs with respect to different configuration when $n=300$. The number (1), (2), and (3) correspond to the three architecture sizes of numerical studies specified in Table~\ref{tab:network}.}
    \label{tab: n300config}
    \begin{adjustbox}{max width=\textwidth}
    \begin{tabular}{lcccccccccccc}
    \toprule
    Method & \multicolumn{2}{c}{SDAMI(1)} & \multicolumn{2}{c}{SDAMI$^*$(2)} &\multicolumn{2}{c}{SDAMI(3)} & \multicolumn{2}{c}{DNN(1)} & \multicolumn{2}{c}{DNN$^*$(2)} &\multicolumn{2}{c}{DNN(3)}\\
    \cmidrule(lr){2-3} \cmidrule(lr){4-5} \cmidrule(lr){6-7}
    \cmidrule(lr){8-9} \cmidrule(lr){10-11} \cmidrule(lr){12-13}
    & MSE$\downarrow$ &STD$\downarrow$ & MSE$\downarrow$ &STD$\downarrow$& MSE$\downarrow$ &STD$\downarrow$& MSE$\downarrow$ &STD$\downarrow$& MSE$\downarrow$ &STD$\downarrow$& MSE$\downarrow$ &STD$\downarrow$  \\
    \midrule
    Case 1 & 2.64 & 3.23 & \textbf{0.43} & 0.65 & 0.48 & 0.71 & 14.11 & 0.71 & 14.10 & 0.73 & 13.95 & 0.68\\
    Case 2 & 0.94 & 1.04 & 0.38 & 0.62 & \textbf{0.29} & 0.56 & 5.31 & 0.40 & 5.26 & 0.31 & 5.23 & 0.30\\
    Case 3 & 1.20 & 1.21 & 0.46 & 0.63 & \textbf{0.29} & 0.45 & 5.76 & 0.39 & 5.70 & 0.32 & 5.62 & 0.26\\
    Case 4 & 0.94 & 1.03 & 0.34 & 0.55 & \textbf{0.32} & 0.58 & 7.07 & 0.48 & 6.98 & 0.38 & 6.95 & 0.36\\
    Case 5 & 0.72 & 0.90 & \textbf{0.35} & 0.58 & 0.37 & 0.65 & 5.80 & 0.38 & 5.78 & 0.35 & 5.74 & 0.36\\
    Case 6 & 0.33 & 0.23 & \textbf{0.25} & 0.21 & 0.25 & 0.21 & 1.03 & 0.17 & 0.99 & 0.20 & 0.37 & 0.19\\
    \hline
    \bottomrule
    \end{tabular}
    \end{adjustbox}
    
\end{table}

\begin{table*}[]
    \centering
    \caption{Model Specification for SDAMIs and DNNs}
    \label{tab:network}
    \begin{adjustbox}{max width=\textwidth}
    \begin{tabular}{c|c|c}
    \hline
    Hyperparameter& \begin{tabular}[c]{@{}c@{}}numerical studies/\\ small-sized dataset\end{tabular} & medium-sized dataset
    \\ \hline
    Architecture& {[}8, 6, 3{]}, {[}15, 12, 10{]}, {[}32,16,8{]}& {[}128, 64, 32, 16{]}, {[}128, 64, 32{]}, {[}64, 32, 16{]} \\ \hline
    Batchsize& 16, 32, 64& 1024, 2048\\ \hline
    Learning rate& 5e-2, 1e-2, 1e-3, 5e-3,& 5e-2, 1e-2, 1e-3, 5e-3,\\ \hline
    Activation & ReLu& ReLu\\ \hline
    Dropout& 0.0, 0.1& 0.0, 0.1\\
    \hline
    \bottomrule
    \end{tabular}
    \end{adjustbox}
    
\end{table*}

\paragraph{fSpAM}
We use fSpAM package \citep{ravikumar2009sparse} and set the basis dimension as $8$ with coordinate descent solver, and best $\lambda$ penalty among $\{0.01, 0.05, 0.1, 0.5\}$ for 5 times and return the best model.
\vspace{-0.3cm}
\paragraph{LASSO and LASSONET}
We use LASSO package \citep{tibshirani1996regression} with default setting and best $\lambda$ penalty among $\{0.001, 0.01, 0.1, 1.0\}$ via 5-fold cross validation. As for LASSONET \citep{lemhadri2021lassonet}, we consider the same architecture in Table~\ref{tab:network} and best $\lambda$ penalty among $\{0.001, 0.01, 0.1\}$ for model comparison.
\vspace{-0.3cm}
\paragraph{NAM}
We utilize NAM package \citep{agarwal2021neural} with number of embedded =32, number of hidden neuron =32, number of  layers=3, and the learning rate=0.0005.
\vspace{-0.3cm}
\paragraph{GAMI-NET}
We utilize the GAMI-NET PyTorch code \citep{yang2021gami}. We set the interact number =10, subnetwork size of main effect= (20,), subnetwork size of interaction=(20, 20), learning rates=$(0.001, 0.001, 0.0001)$, and loss threshold=0.01 and set early stop 100 rounds to ensure convergence.
\vspace{-0.3cm}
\paragraph{NODE-GAM and NODE-GA$^2$M} We utilize the default hyperparameters from NODE-GAM PyTorch code \citep{chang2021node}, and set the number of trees to a large number 500, arch = GAM, learning rate = 0.01, warm-up = 100, and max epoch = 20000 to ensure it converges.

\subsection{Optimal hyperparameters found in each dataset}\label{appendix-opt}
Here we report the best hyperparameters we find for 3 small-sized datasets and 3 medium-sized datasets in Table~\ref{tab: optimalnetwork}.

\begin{table*}[h]
    \caption{The optimal model specification for SDAMI architecture}
    \label{tab: optimalnetwork}
    \begin{adjustbox}{max width=\textwidth}
    \begin{tabular}{c|c|c|c|c|c|c|cc}
    \hline
    Hyperparameter& numerical studies & Chip & Diabetes & V1 Cell & Wine & BikeShare & CA Housing
    \\ 
    \hline
    Architecture& [15, 12, 10] & [8, 6, 3]& [32, 16, 8]& [15, 12, 10]& [128, 64, 32, 16]& [128, 64, 32]& [128, 64, 32]\\ \hline
    Batchsize& 32 & 64 & 64 & 32 & 2048 & 2048 & 2048 \\ \hline
    Learning rate& 5e-2 & 1e-3 & 5e-2 & 5e-2 & 5e-2 & 1e-2 & 1e-3 \\ \hline
    Activation & ReLu& ReLu & ReLu& ReLu & ReLu& ReLu & ReLu\\ \hline
    Dropout& 0.0 & 0.0 & 0.0 & 0.0 & 0.0 & 0.0 & 0.0 \\ \hline \midrule
    \end{tabular}
    \end{adjustbox}
    
\end{table*}

\section{Supplementary material for additional experiment results}\label{add-simulation}
\subsection{Complete comparison for numerical studies}
The performance comparison among different machine learning model is demonstrated in Table~\ref{tab: n300}~\ref{tab: n450}, and the results of SDAMI$-p$ for both the numerical studies and real data analysis in Table~\ref{tab: SDAMI-sim}~\ref{tab: SDAMI-real-sim}. Also, Table \ref{tab: tpr300} results for additional numerical experiments with different sample size and corresponding TPR/ FPR are demonstrated in the following block. In Table~\ref{tab:sdami_split_tpr}, we show that across 100 independent replications per scenario, SDAMI recovers all main effects perfectly in five of the six simulation settings ($\text{TPR}_{\text{main}} = 1.000$ with zero standard deviation for \texttt{only\_main}, \texttt{inter\_no\_overlap}, \texttt{inter\_mild\_overlap}, \texttt{inter\_strong\_overlap}, and \texttt{only\_inter}); the sole exception is \texttt{weak\_main} ($\text{TPR}_{\text{main}} = 0.748 \pm 0.025$), where the fourth main effect is intentionally attenuated by a factor of $0.01$ and falls below the Stage 2 group-lasso threshold. Pairwise interaction recovery follows the expected difficulty ordering, increasing from $\text{TPR}_{\text{inter}} = 0.830 \pm 0.376$ for \texttt{inter\_no\_overlap}, to $0.910 \pm 0.286$ for \texttt{inter\_mild\_overlap}, $0.980 \pm 0.140$ for \texttt{inter\_strong\_overlap}, and $0.955 \pm 0.160$ for \texttt{only\_inter}. False positive rates are negligible across all settings ($\text{FPR}_{\text{main}} \le 0.0003$, $\text{FPR}_{\text{inter}} = 0.000$ in every case), in sharp contrast to a LASSONET baseline whose main-effect FPR rises monotonically with sample size and reaches $0.49$ at $n = 450$ on \texttt{only\_main}.

\begin{table*}[t]
    \centering   
    \caption{(RMSE) The performance for 6 different case type when $n=300$; $\downarrow$ means the lowest the better while $\uparrow$ means the highest the better.}
    \label{tab: n300}
    \begin{adjustbox}{max width=1.0\textwidth}
    \begin{tabular}{lcccccccccc}
    \toprule
    & SDAMI  & DNN & fSpAM & LASSO & NAM & GAMI-NET & NODE-GA$^2$M & NODE-GAM \\
    \midrule
    Case 1 & \textbf{0.56}$_{\pm 1.04}$& 14.11$_{\pm 0.71}$ & 5.57$_{\pm 0.31}$ & 3.32$_{\pm 0.24}$& 10.10$_{\pm 0.92}$ & 2.62$_{\pm 0.63}$ & 2.66$_{\pm 0.29}$ & 2.74$_{\pm 0.50}$ \\
    Case 2 &\textbf{0.30}$_{\pm 0.57}$& 5.31$_{\pm 0.40}$ & 3.04$_{\pm 0.16}$ & 2.54$_{\pm 0.17}$ & 5.23$_{\pm 0.43}$ & 2.26$_{\pm 1.05}$ & 1.52$_{\pm 0.70}$ & 1.97$_{\pm 0.36}$  \\
    Case 3 &\textbf{0.27}$_{\pm 0.37}$& 5.31$_{\pm 0.40}$ & 3.04$_{\pm 0.16}$ & 2.54$_{\pm 0.17}$ & 4.01$_{\pm 0.43}$ & 1.27$_{\pm 0.52}$ & 0.71$_{\pm 0.25}$ & 2.26$_{\pm 0.45}$ \\
    Case 4 &\textbf{0.24}$_{\pm 0.41}$& 7.07$_{\pm 0.51}$ & 3.32$_{\pm 0.17}$ & 2.80$_{\pm 0.16}$ & 4.65$_{\pm 0.59}$ & 1.55$_{\pm 0.73}$ & 0.72$_{\pm 0.35}$ & 2.29$_{\pm 0.59}$  \\
    Case 5 &\textbf{0.41}$_{\pm 0.76}$& 5.80$_{\pm 0.38}$ & 3.45$_{\pm 0.16}$ & 2.98$_{\pm 0.20}$ & 3.84$_{\pm 0.53}$ & 1.05$_{\pm 0.61}$ & 0.57$_{\pm 0.19}$ & 2.14$_{\pm 0.44}$  \\
    Case 6 &\textbf{0.35}$_{\pm 0.21}$& 1.03$_{\pm 0.17}$ & 0.60$_{\pm 0.03}$ & 0.43$_{\pm 0.03}$ & 0.70$_{\pm 0.06}$ & 0.29$_{\pm 0.13}$ & 0.42$_{\pm 0.03}$ & 0.45$_{\pm 0.04}$ \\
    \hline
    \bottomrule
    \end{tabular}
    \end{adjustbox}
    
\end{table*}
\vspace{-0.5em}

\begin{table}[t]
    \centering    
    \caption{(RMSE) The performance for 6 different case type when $n=450$; $\downarrow$ means the lowest the better while $\uparrow$ means the highest the better.}
    \label{tab: n450}
    \adjustbox{max width=\textwidth}{
    \begin{tabular}{lcccccccccc}
    \toprule
    & SDAMI  & DNN & fSpAM & LASSO & NAM & GAMI-NET & NODE-GA$^2$M & NODE-GAM \\
    \midrule
    Case 1 &\textbf{0.21}$_{\pm 0.12}$ & 13.89$_{\pm 0.82}$ & 5.43$_{\pm 0.28}$ & 3.04$_{\pm 0.17}$& 5.60.$_{\pm 0.98}$ & 1.49$_{\pm 0.74}$ & 0.35$_{\pm 0.10}$ & 1.62$_{\pm 0.48}$ \\
    Case 2 &\textbf{0.14}$_{\pm 0.04}$ & 5.33$_{\pm 0.35}$ & 2.98$_{\pm 0.14}$ & 2.40$_{\pm 0.11}$ & 1.71$_{\pm 0.30}$ & 0.73$_{\pm 0.36}$ & 0.25$_{\pm 0.07}$ & 1.20$_{\pm 0.34}$  \\
    Case 3 &\textbf{0.17}$_{\pm 0.13}$ & 5.78$_{\pm 0.32}$ & 3.33$_{\pm 0.16}$ & 2.72$_{\pm 0.14}$ & 2.02$_{\pm 0.39}$ & 0.81$_{\pm 0.35}$ & 0.40$_{\pm 0.08}$ & 1.57$_{\pm0.35}$ \\
    Case 4 &\textbf{0.25}$_{\pm 0.53}$ & 7.14$_{\pm 0.50}$ & 3.24$_{\pm 0.15}$ & 2.61$_{\pm 0.13}$ & 2.54$_{\pm 0.41}$ & 0.93$_{\pm 0.35}$ & 0.39$_{\pm 0.06}$ & 1.49$_{\pm 0.35}$  \\
    Case 5 &\textbf{0.26}$_{\pm 0.18}$ & 5.82$_{\pm 0.39}$ & 3.41$_{\pm 0.15}$ & 2.76$_{\pm 0.13}$ & 2.09$_{\pm 0.48}$ & 0.59$_{\pm 0.35}$ & 0.38$_{\pm 0.08}$ & 1.45$_{\pm 0.35}$  \\
    Case 6 &\textbf{0.16}$_{\pm 0.12}$ & 1.06$_{\pm 0.13}$ & 0.59$_{\pm 0.03}$ & 0.39$_{\pm 0.02}$ & 0.52$_{\pm 0.04}$ & 0.16$_{\pm 0.05}$ & 0.37$_{\pm 0.03}$ & 0.37$_{\pm 0.03}$ \\
    \hline
    \bottomrule
    \end{tabular}}
\end{table}

\vspace{-0.5em}

\begin{table}[t]
    \centering
    \caption{Mean (standard deviation) of TPR and FPR over 100 simulations from SDAMI, LASSONET, SODA when $n=300$ where $(-)$ indicates value $<1e^{-5}$.}
    \label{tab: tpr300}
    \begin{adjustbox}{max width=\textwidth}
    \begin{tabular}{lcccccc}
    \toprule
    \multicolumn{1}{c}{Method} & \multicolumn{2}{c}{SDAMI}&\multicolumn{2}{c}{LASSONET} &\multicolumn{2}{c}{SODA} \\
    \cmidrule(lr){2-3} \cmidrule(lr){4-5} \cmidrule(lr){6-7} 
    & TPR$\uparrow$ & FPR$\downarrow$ & TPR$\uparrow$ & FPR$\downarrow$& TPR$\uparrow$ & FPR$\downarrow$ \\
    \midrule
    Case 1 & \textbf{1.000} (-) & $1.1\times10^{-5}$ (-) & 0.610 (0.124) & 0.013 (0.009) & 0.030 (0.096) & $5\times 10^{-4}$ ($2\times 10^{-4}$)\\
    Case 2 & \textbf{1.000} (-) & $1.1\times10^{-5}$ (-) & 0.455 (0.108) & 0.017 (0.008) & 0.020 (0.069) & $4\times 10^{-4}$ ($2\times 10^{-4}$)\\
    Case 3 & \textbf{0.750} (-) & $10^{-4}$ ($10^{-5}$)& 0.020 (0.098)& 0.045 (0.042) & 0.015 (0.060) & $6\times 10^{-4}$ ($4\times 10^{-4}$ \\
    Case 4 & \textbf{0.760} (0.049) & $10^{-4}$ ($10^{-5}$)& 0.010 (0.070) & 0.037 (0.018) & 0.025 (0.076) & $5\times 10^{-4}$ ($2\times 10^{-4}$)\\
    Case 5 & \textbf{0.753} (0.025)& $10^{-4}$ ($10^{-5}$)& 0.010 (0.070) & 0.039 (0.018) & 0.030 (0.082) & $5\times 10^{-4}$ ($2\times 10^{-4}$)\\
    Case 6 & \textbf{0.610} (0.044)& $10^{-4}$ ($10^{-5}$)& 0.020 (0.098) & 0.052 (0.066) & $-(-)$  & $5\times 10^{-4}$ ($2\times 10^{-4}$)\\
    \hline
    \bottomrule
    \end{tabular}
    \end{adjustbox}
    
\end{table}

\begin{table}[t]
\centering
\small
\caption{Variable-selection performance of SDAMI on six simulation scenarios,
reported separately for main effects and pairwise interactions.
Values are mean (standard deviation) over $N = 100$ independent replications.
NA indicates that the ground truth contains no components of the corresponding
type (no interactions in \texttt{only\_main}/\texttt{weak\_main}; no main effects in \texttt{only\_inter}). }
\label{tab:sdami_split_tpr}
\begin{adjustbox}{max width=1.0\textwidth}
\begin{tabular}{lccccc}
\toprule
Scenario & $\mathrm{TPR}_{\mathrm{main}}$ & $\mathrm{FPR}_{\mathrm{main}}$ & $\mathrm{TPR}_{\mathrm{inter}}$ & $\mathrm{FPR}_{\mathrm{inter}}$ & $F_1$ \\
\midrule
\texttt{only\_main}            & 1.000 ($-$) & $-$ & NA             & $-$ & 1.000 ($-$) \\
\texttt{weak\_main}            & 0.748 (0.025) & $-$ & NA& $-$ & 0.857 ($-$) \\
\texttt{inter\_no\_overlap}    & 1.000 ($-$) & $3\times10^{-4} $(0.001) & 0.830 (0.376) & $-$ & 0.963 (0.085) \\
\texttt{inter\_mild\_overlap}  & 1.000 ($-$) & $-$ & 0.910 (0.286) & $-$ & 0.987 (0.041) \\
\texttt{inter\_strong\_overlap}& 1.000 ($-$) & $-$ & 0.980 (0.140) & $-$ & 1.000 ($-$) \\
\texttt{only\_inter}           & NA              & $10^{-4} (10^{-3})$ & 0.955 (0.160) & $-$ & 0.967 (0.129) \\
\bottomrule
\end{tabular}
\end{adjustbox}

\end{table}

\begin{table*}[h]
    \centering
    \caption{(RMSE) The performance for 6 different case type for SDAMI$-p$; $\downarrow$ means the lowest the better while $\uparrow$ means the highest the better.}
    \label{tab: SDAMI-sim}
    \begin{adjustbox}{max width=1.0\textwidth}
    \begin{tabular}{cccccccc}
    \hline
    &         & \multicolumn{1}{c}{Case 1} & \multicolumn{1}{c}{Case 2} & \multicolumn{1}{c}{Case 3} & \multicolumn{1}{c}{Case 4} & \multicolumn{1}{c}{Case 5} & Case 6 \\ \hline
    \multirow{3}{*}{SDAMI$-p$} & n = 150   & 0.68 $_{\pm0.59}$ & 0.77 $_{\pm0.41}$ & 0.70 $_{\pm0.58}$ & 0.84 $_{\pm0.27}$ & 0.85 $_{\pm0.53}$
    & 0.27 $_{\pm0.25}$\\
    & n = 300 & 0.48 $_{\pm0.71}$ & 0.29 $_{\pm0.56}$ & 0.29 $_{\pm0.45}$ & 0.32 $_{\pm0.58}$ & 0.37 $_{\pm0.65}$
    & 0.25 $_{\pm0.21}$ \\ & n = 450 & 0.23 $_{\pm0.63}$ & 0.21 $_{\pm0.46}$ & 0.28 $_{\pm0.37}$ & 0.17 $_{\pm0.21}$ & 0.22 $_{\pm0.19}$
    & 0.14 $_{\pm0.18}$\\
    \hline
    \bottomrule
    \end{tabular}
    \end{adjustbox}
    
\end{table*}

\begin{table}[t]
    \centering
    \caption{(RMSE) The performance for 7 different real dataset for SDAMI$-p$; $\downarrow$ means the lowest the better while $\uparrow$ means the highest the better.}
    \label{tab: SDAMI-real-sim}
    \begin{tabular}{cccccccc}
    \hline
    & Chip & \multicolumn{1}{c}{Diabetes} & \multicolumn{1}{c}{V1 Cell} & \multicolumn{1}{c}{Wine} & \multicolumn{1}{c}{BikeShare} & \multicolumn{1}{c}{CA Housing} \\ \hline
    SDAMI$-p$ & 0.236 & 52.87 & 0.372 & 0.692 & 55.91 & 0.508 \\
    \hline
    \bottomrule
    \end{tabular}
    
\end{table}

\subsection{Uncertainty bands and baseline comparison}\label{case:visual}

In the section, we demonstrate the visualization of either main effects or interaction across each cases where the visualization result for Case 3 can be found in Figure~\ref{fig:case3visual}. In Case 1 to 5, the SDAMI can capture both linearity and nonlinearity underlying the true model. In the interaction-existed cases, we observes the SDAMI can still depict the response surface to approximate the underlying higher-order effects.

\begin{figure}
    \centering
    \includegraphics[width=1.0\linewidth]{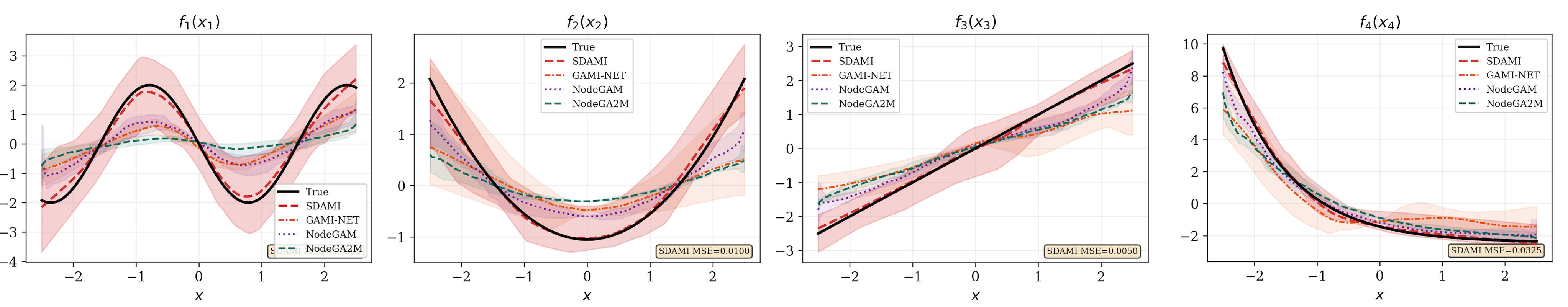}
    \includegraphics[width=1.0\linewidth]{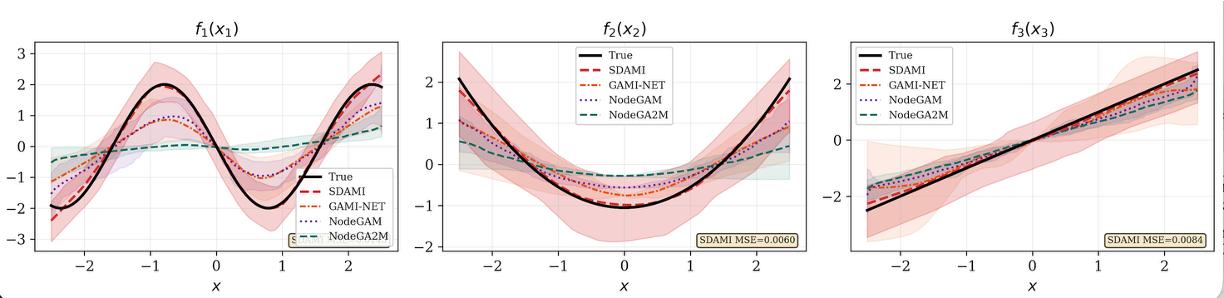}
    \caption{The estimated (red dashed lines) versus true additive component functions (solid black lines) for four main effects for (Upper panel) Case (1) and (Lower panel) Case (2).}
    \label{fig:Casevisualm}
\end{figure}

\begin{figure}
    \centering
    \includegraphics[width=1.0\linewidth]{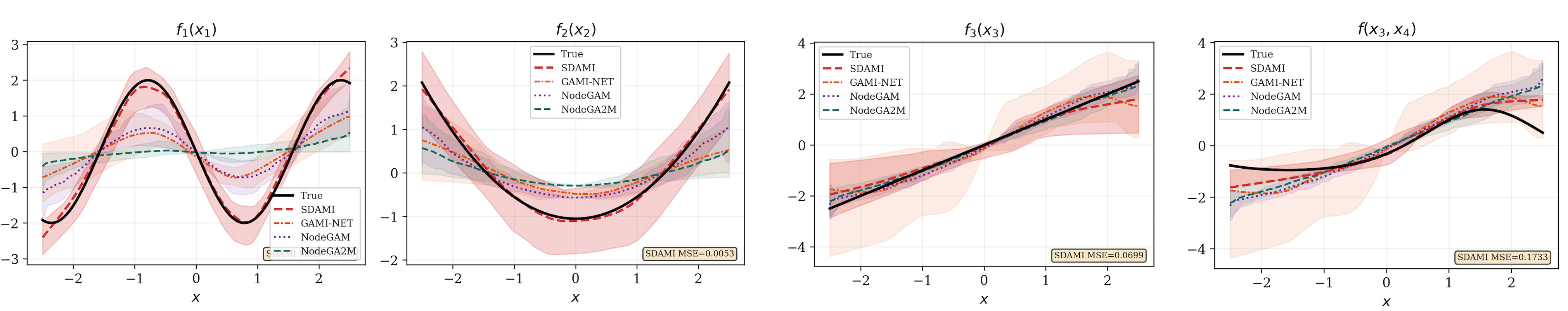}
    \includegraphics[width=1.0\linewidth]{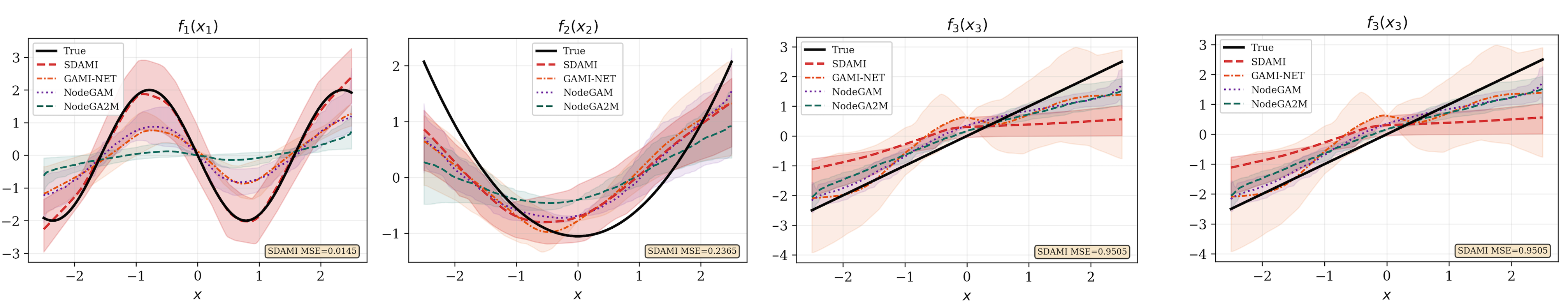}
    \includegraphics[width=1.0\linewidth]{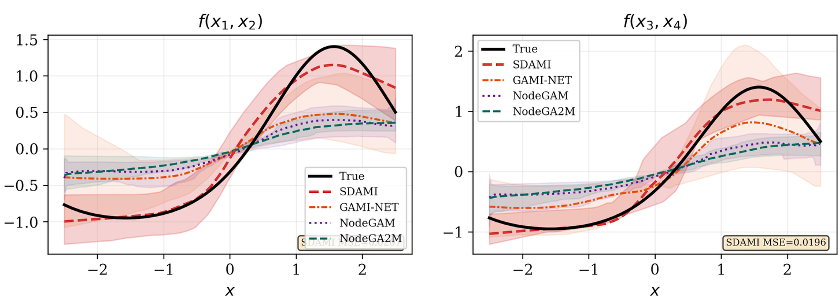}
    \caption{(Upper panel: Case (4); middle panel: Case (5)) The three figures on the left: Estimated (red dashed lines) versus true additive component functions (solid black lines) for three main effects; the two figures on the far right: the first shows the true response surface for interaction, and the second shows its estimated response surface. (Lower panel: Case (6)) The first and third shows the true response surface for interactions, and the second and fourth  shows corresponding estimated response surface.}
    \label{fig:Casevisuali}
\end{figure}

This appendix provides median recovered effect functions together with uncertainty bands across repeated runs, along with side-by-side comparisons against representative baselines. These visualizations complement the aggregate prediction metrics in the main text and make the variability of the recovered shapes explicit. The rest of case can be found in detail in the Figure \ref{fig:Casevisualm}, \ref{fig:Casevisuali}.

\subsection{Pareto-style comparisons}\label{app:pareto}

\begin{figure}[H]
    \centering
    \includegraphics[width=1.0\linewidth]{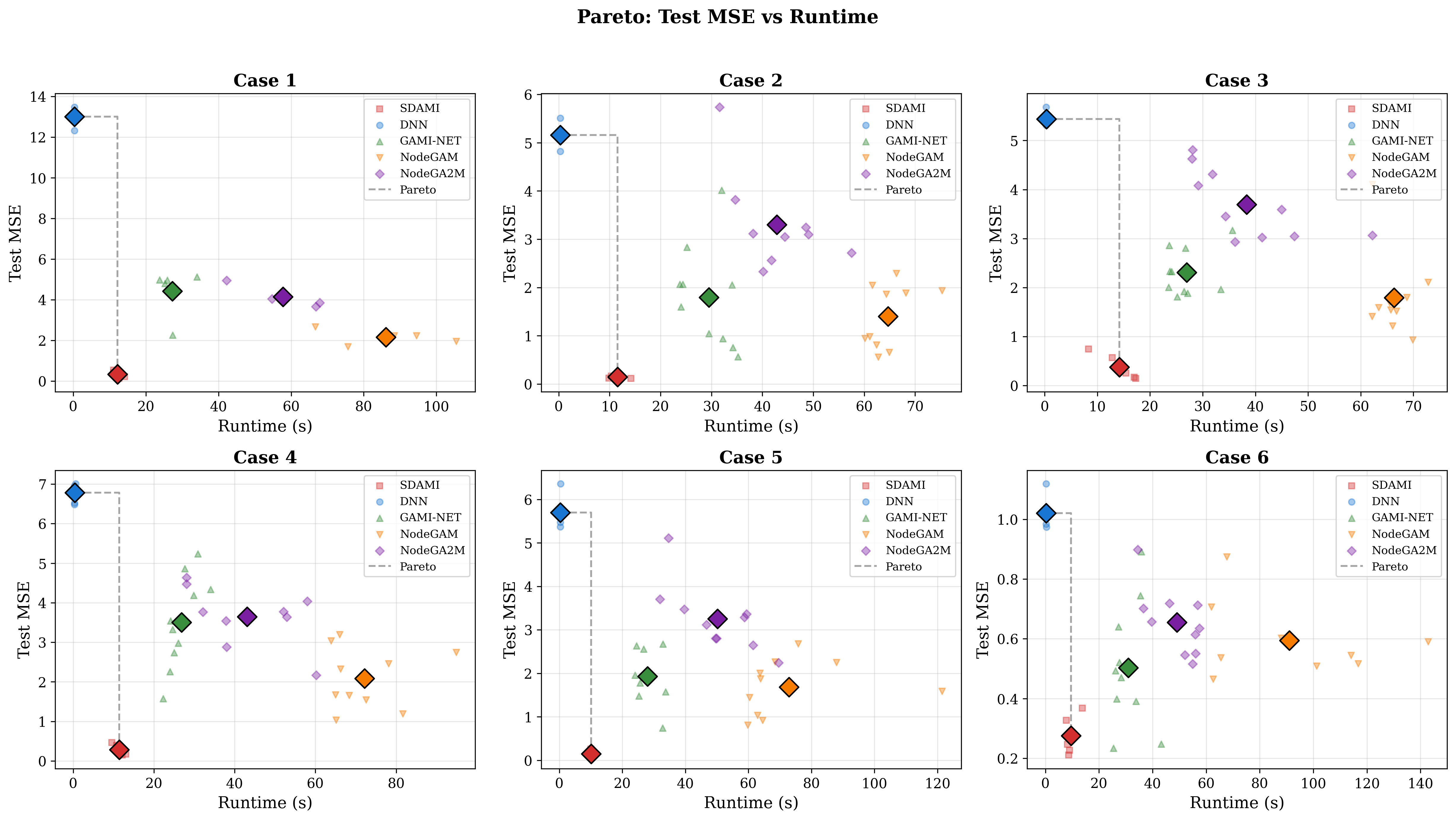}
    \caption{Pareto Comparison (MSE vs. Runtime)}
    \label{fig:pareto_mse}
\end{figure}

To assess the trade-off between predictive performance and computational cost, we report a Pareto-style comparison of mean squared error versus wall-clock runtime across the competing methods. Figure~\ref{fig:pareto_mse} is intended to show whether the predictive gains of SDAMI are achieved at a disproportionate computational cost. The results indicate that SDAMI lies close to the Pareto frontier, achieving strong prediction accuracy while maintaining a moderate runtime relative to alternative methods.

\begin{figure}[h!]
    \centering
    \includegraphics[width=\textwidth]{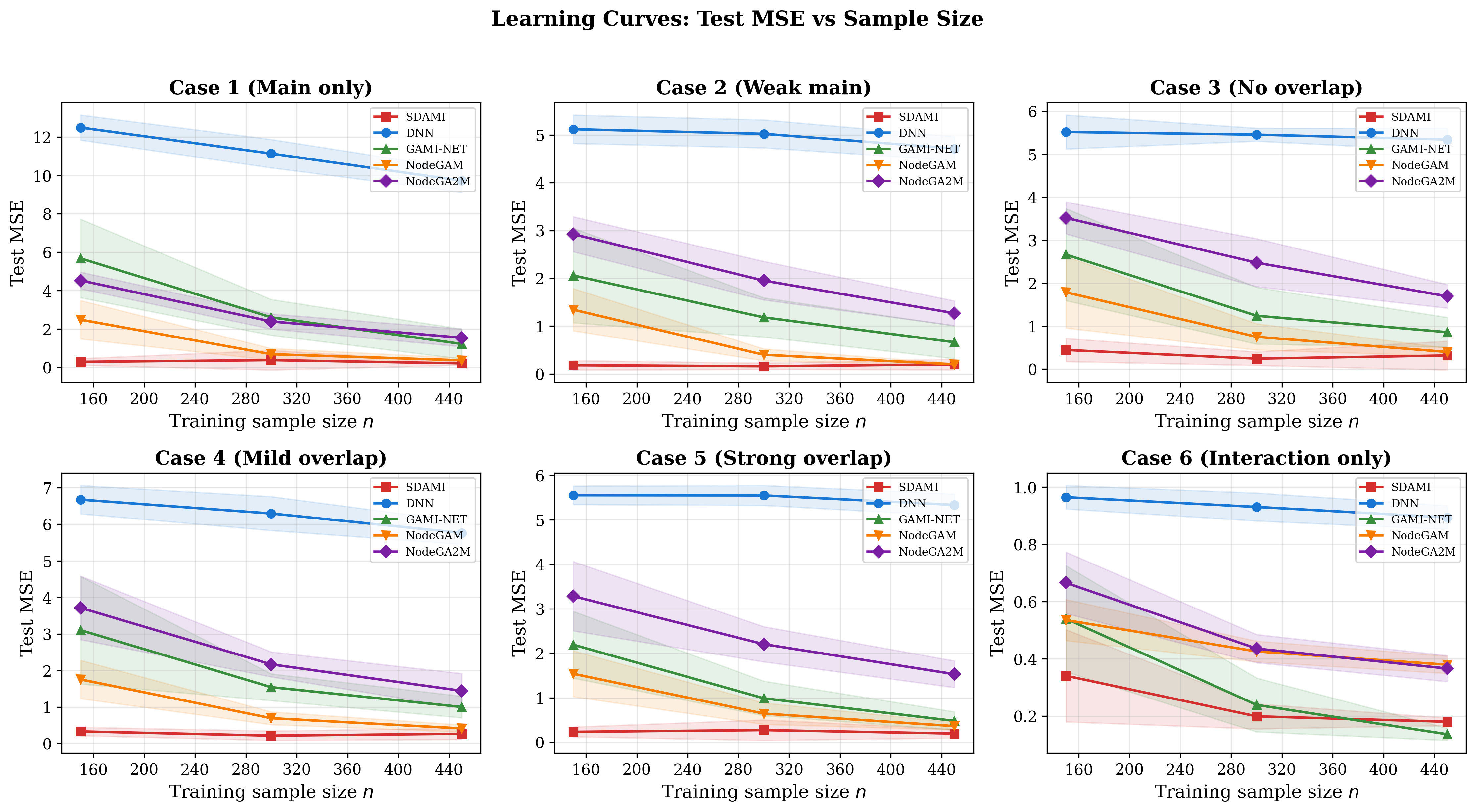}
    \caption{Learning curve for number of sample size versus MSE}
    \label{fig:learning_curve_sample_mse}
\end{figure}

\begin{figure}[t]
    \centering
    \includegraphics[width=\textwidth]{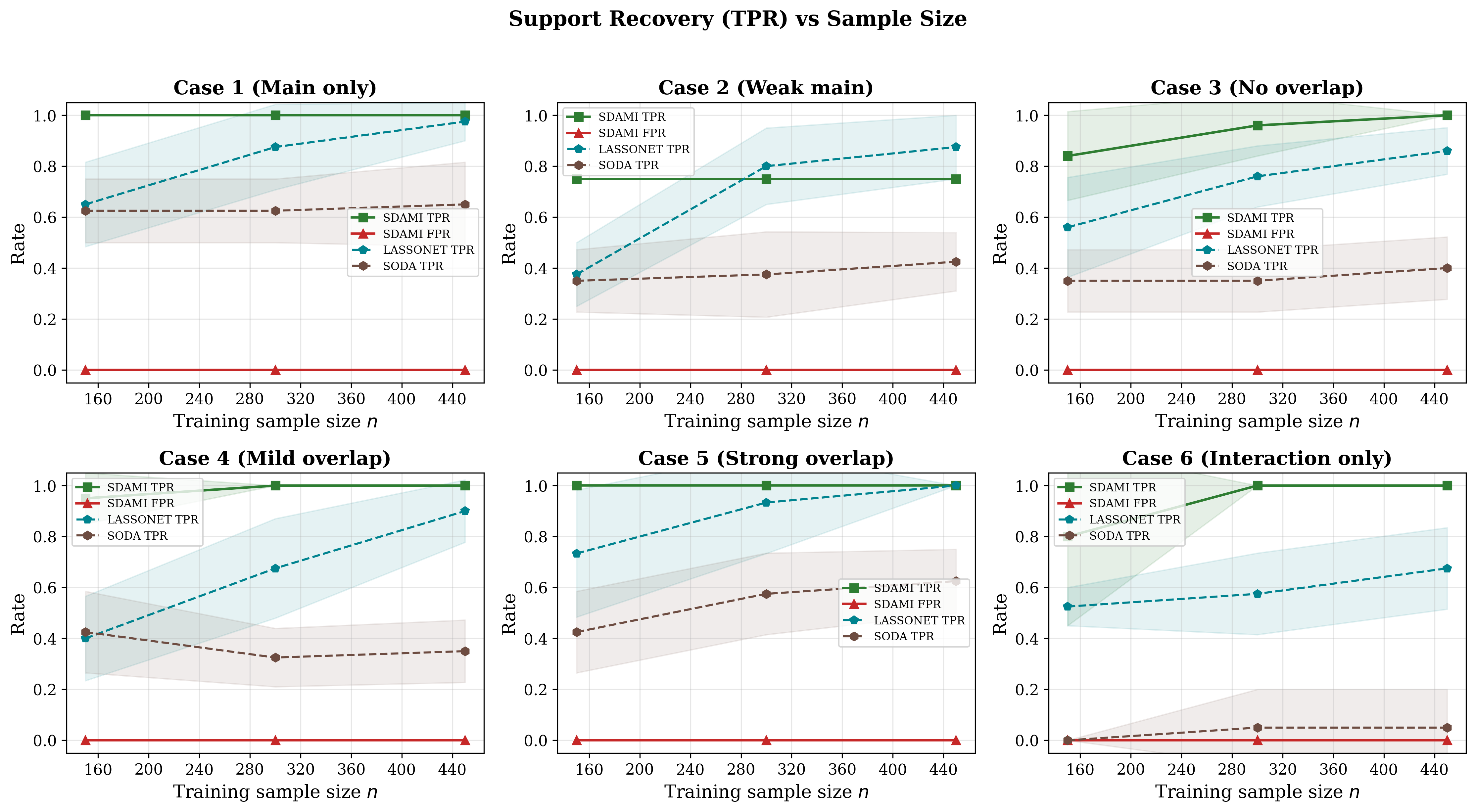}
    \caption{Learning curve for number of sample size versus TPR/ FPR}
    \label{fig:learning_curve_sample_tpr}
\end{figure}

In addition to the runtime-based Pareto comparison, we report learning curves showing prediction error and recovery rate as functions of sample size. These curves complement the fixed-sample comparisons in the main text by showing how rapidly SDAMI improves as the sample size increases in Figure~\ref{fig:learning_curve_sample_mse} and~\ref{fig:learning_curve_sample_tpr}. 

\subsection{Ablation studies of SDAMI pipeline}\label{app:ablation}
We conduct ablation studies to isolate the contribution of each stage in the SDAMI pipeline. In particular, we compare the full SDAMI model against variants that remove Stage 1 screening, remove Stage 2 decomposition, or replace the final neural subnetworks with simpler regression modules when applicable in Table~\ref{tab:ablation}. For each variant, we report prediction error, support-recovery metrics, and runtime. These experiments clarify the distinct role of each stage: Stage 1 primarily reduces the effective search space and improves scalability, Stage 2 improves structural precision by separating main and interaction supports, and Stage 3 provides flexible nonlinear approximation once the relevant structure has been identified.

\begin{table}[!htbp]
\centering
\caption{Ablation study results for SDAMI. Mean $\pm$ std over repeated runs.}
\label{tab:ablation}
\small
\begin{adjustbox}{max width=\textwidth}
\begin{tabular}{llccccc}
\toprule
Case & Ablation & MSE & F1 & TPR & FPR & Runtime (s) \\
\midrule
  Case 1 (Main only) & Full SDAMI & $0.31 \pm 0.17$ & $1.00 \pm 0.00$ & 1.00 & 0.00 & 17.9 \\
   & w/o Stage-1 screening & $1.72 \pm 0.84$ & $0.82 \pm 0.08$ & 0.70 & 0.00 & 1836.4 \\
   & w/o Stage-2 group-lasso & $0.30 \pm 0.12$ & $1.00 \pm 0.00$ & 1.00 & 0.00 & 11.8 \\
   & Main effects only & $0.25 \pm 0.15$ & $1.00 \pm 0.00$ & 1.00 & 0.00 & 18.0 \\
   & Ridge instead of neural & $0.27 \pm 0.16$ & $1.00 \pm 0.00$ & 1.00 & 0.00 & 15.3 \\
\cmidrule{1-7}
  Case 2 (Weak main) & Full SDAMI & $0.14 \pm 0.03$ & $0.86 \pm 0.00$ & 0.75 & 0.00 & 15.3 \\
   & w/o Stage-1 screening & $1.13 \pm 0.65$ & $0.66 \pm 0.10$ & 0.50 & 0.00 & 1716.8 \\
   & w/o Stage-2 group-lasso & $0.16 \pm 0.05$ & $0.86 \pm 0.00$ & 0.75 & 0.00 & 10.2 \\
   & Main effects only & $0.31 \pm 0.52$ & $0.86 \pm 0.00$ & 0.75 & 0.00 & 15.3 \\
   & Ridge instead of neural & $0.17 \pm 0.03$ & $0.86 \pm 0.00$ & 0.75 & 0.00 & 13.2 \\
\cmidrule{1-7}
  Case 3 (No overlap) & Full SDAMI & $0.22 \pm 0.13$ & $0.97 \pm 0.07$ & 0.96 & 0.00 & 30.2 \\
   & w/o Stage-1 screening & $1.43 \pm 0.66$ & $0.64 \pm 0.09$ & 0.48 & 0.00 & 1708.7 \\
   & w/o Stage-2 group-lasso & $0.26 \pm 0.05$ & $0.86 \pm 0.11$ & 1.00 & 0.01 & 12.6 \\
   & Main effects only & $0.65 \pm 0.08$ & $0.75 \pm 0.00$ & 0.60 & 0.00 & 29.5 \\
   & Ridge instead of neural & $0.64 \pm 0.06$ & $0.75 \pm 0.00$ & 0.60 & 0.00 & 27.7 \\
\cmidrule{1-7}
  Case 4 (Mild overlap) & Full SDAMI & $0.19 \pm 0.08$ & $0.97 \pm 0.06$ & 0.95 & 0.00 & 23.9 \\
   & w/o Stage-1 screening & $1.57 \pm 0.94$ & $0.63 \pm 0.13$ & 0.47 & 0.00 & 1728.4 \\
   & w/o Stage-2 group-lasso & $0.24 \pm 0.05$ & $0.84 \pm 0.12$ & 1.00 & 0.01 & 15.6 \\
   & Main effects only & $0.36 \pm 0.03$ & $0.86 \pm 0.00$ & 0.75 & 0.00 & 23.4 \\
   & Ridge instead of neural & $0.39 \pm 0.04$ & $0.86 \pm 0.00$ & 0.75 & 0.00 & 21.6 \\
\cmidrule{1-7}
  Case 5 (Strong overlap) & Full SDAMI & $0.19 \pm 0.09$ & $1.00 \pm 0.00$ & 1.00 & 0.00 & 20.3 \\
   & w/o Stage-1 screening & $0.42 \pm 0.63$ & $0.98 \pm 0.06$ & 0.97 & 0.00 & 1752.3 \\
   & w/o Stage-2 group-lasso & $0.23 \pm 0.07$ & $0.83 \pm 0.13$ & 1.00 & 0.01 & 11.2 \\
   & Main effects only & $0.24 \pm 0.12$ & $1.00 \pm 0.00$ & 1.00 & 0.00 & 19.3 \\
   & Ridge instead of neural & $0.24 \pm 0.03$ & $1.00 \pm 0.00$ & 1.00 & 0.00 & 17.3 \\
\cmidrule{1-7}
  Case 6 (Inter only) & Full SDAMI & $0.12 \pm 0.01$ & $1.00 \pm 0.00$ & 1.00 & 0.00 & 14.0 \\
   & w/o Stage-1 screening & $0.67 \pm 0.15$ & $0.46 \pm 0.22$ & 0.33 & 0.00 & 1192.0 \\
   & w/o Stage-2 group-lasso & $0.25 \pm 0.02$ & $0.96 \pm 0.10$ & 1.00 & 0.00 & 11.2 \\
   & Main effects only & $-$ & $0.00 \pm 0.00$ & 0.00 & 0.00 & 11.9 \\
   & Ridge instead of neural & $0.23 \pm 0.01$ & $1.00 \pm 0.00$ & 1.00 & 0.00 & 11.8 \\
\bottomrule
\end{tabular}
\end{adjustbox}
\end{table}

\subsection{Component--wise estimation accuracy}\label{app:component}
To complement the aggregate prediction metrics reported in the main text, we further examine component-wise estimation accuracy for the recovered main-effect and interaction functions. Specifically, for each simulated case, we compute the estimation error of each true component separately and compare SDAMI against the competing baselines. These results help distinguish overall predictive performance from recovery of the underlying functional structure, which is particularly important in an interpretable additive-plus-interaction framework. The corresponding table~\ref{tab:component_mse} shows that SDAMI more faithfully recovers the true effect components in most settings, especially when interaction structure is present.

\begin{table}[ht]
\centering
\caption{Component-wise MSE (SDAMI) / MSE (baselines) on true active components.}
\label{tab:component_mse}
\begin{tabular}{llrrrr}
\toprule
Scenario & Component & SDAMI & GAMI-NET & NodeGAM & NodeGA2M \\
\midrule
  Case 1 (Main only) & $f_1(x_1)$ & 0.226 & 0.933 & 0.737 & 1.605 \\
\addlinespace[2pt]
  & $f_2(x_2)$ & 0.080 & 0.585 & 0.539 & 0.558 \\
  & $f_3(x_3)$ & 0.034 & 0.552 & 0.219 & 0.384 \\
  & $f_4(x_4)$ & 0.060 & 2.533 & 1.676 & 1.669 \\
  Case 2 (Weak main) & $f_1(x_1)$ & 0.033 & 0.629 & 0.491 & 1.745 \\
\addlinespace[2pt]
  & $f_2(x_2)$ & 0.045 & 0.734 & 0.517 & 0.567 \\
  & $f_3(x_3)$ & 0.102 & 0.082 & 0.123 & 0.226 \\
  & $f_4(x_4)$ & — & 0.001 & 0.001 & 0.002 \\
  Case 3 (No overlap) & $f_1(x_1)$ & 0.082 & 0.764 & 0.558 & 1.765 \\
\addlinespace[2pt]
  & $f_2(x_2)$ & 0.013 & 0.688 & 0.501 & 0.599 \\
  & $f_3(x_3)$ & 0.109 & 0.135 & 0.145 & 0.293 \\
  & $f(x_4,x_5)$ & 0.035 & 0.417 & 0.429 & 0.476 \\
  Case 4 (Mild overlap) & $f_1(x_1)$ & 0.063 & 1.030 & 0.752 & 1.836 \\
\addlinespace[2pt]
  & $f_2(x_2)$ & 0.016 & 0.572 & 0.512 & 0.563 \\
  & $f_3(x_3)$ & 0.475 & 0.112 & 0.029 & 0.015 \\
  & $f(x_3,x_4)$ & 0.408 & 0.339 & 0.520 & 0.399 \\
  Case 5 (Strong overlap) & $f_1(x_1)$ & 0.057 & 0.716 & 0.601 & 1.644 \\
\addlinespace[2pt]
  & $f_2(x_2)$ & 0.399 & 0.888 & 0.719 & 0.610 \\
  & $f_3(x_3)$ & 1.121 & 0.618 & 0.306 & 0.256 \\
  & $f(x_2,x_3)$ & 0.442 & 1.117 & 1.035 & 0.644 \\
  Case 6 (Inter only) & $f(x_1,x_2)$ & 0.041 & 0.288 & 0.371 & 0.421 \\
\addlinespace[2pt]
  & $f(x_3,x_4)$ & 0.041 & 0.179 & 0.320 & 0.363 \\
\bottomrule
\end{tabular}
\end{table}
\subsection{Permutation test for component significance (B = 100, BH-adjusted)}\label{appendix:test}
We applied the permutation test described in Section~\ref{sec: ns}
to every component selected by SDAMI in each of the six simulation
scenarios, yielding $21$ tests in total ($14$ main effects and
$7$ pairwise interactions). With $B = 100$ permutations, the smallest $p$-value attainable is $1/101 \approx 0.0099$. \emph{All $21$ components
reject $H_0\!: f_j \equiv 0$ after Benjamini--Hochberg correction at
FDR~$0.05$} ($p_{\mathrm{BH}} \le 0.0198$ for every test); $20$ of the $21$ attain the floor $p_{\mathrm{raw}} = 0.0099$
($p_{\mathrm{BH}} = 0.0104$). The single component with a slightly
larger $p$-value is the main effect of $x_2$ in
\texttt{inter\_strong\_overlap} ($T_{\mathrm{obs}} = 0.0945$,
$p_{\mathrm{BH}} = 0.0198$), whose marginal contribution is partially absorbed into the strongly overlapping $(x_1, x_2)$ interaction; it nevertheless remains significant at the $0.05$ level.

The observed test statistics dominate the permutation null
distributions by one to three orders of magnitude across all scenarios. For example, in \texttt{only\_main} the four selected main effects yield $T_{\mathrm{obs}} \in \{2.10, 0.89, 2.13, 9.01\}$ against
permutation means of $\{0.010, 0.006, 0.027, 0.015\}$; in
\texttt{only\_inter} the two selected interactions yield
$T_{\mathrm{obs}} \in \{0.756, 0.770\}$ against null means of
$\{0.008, 0.005\}$. These results confirm that the components SDAMI selects are not artifacts of the post-selection neural fit: each surviving component carries genuine signal that cannot be explained by chance association after the feature columns are randomly permuted and the additive network re-estimated under the same architecture. The complete table is provided in Table~\ref{tab:perm_significance}.

\begin{table}[h]
\centering
\small
\caption{
Permutation test for the significance of every component selected by SDAMI on
the six simulation scenarios. The reported $p$-value is
$p_{\mathrm{raw}} = (1 + |\{b : T_j^{(b)} \ge T_j\}|)/(B+1)$, and $p_{\mathrm{BH}}$ is its
Benjamini--Hochberg adjusted value across all $21$ tests at FDR~$0.05$.
}
\label{tab:perm_significance}
\begin{adjustbox}{max width=1.0\textwidth}
\begin{tabular}{llrrrrrl}
\toprule
Scenario & Component & Type & $T_{obs}$ & $T_{perm}$ mean(std) & $p$ (raw) & $p_{BH}$ & Sig.($\alpha=.05$)\\
\midrule
\texttt{only\_main} & [0] & main & 2.1010 & 0.0104(0.0206) & 0.0099 & 0.0104 & $\checkmark$ \\
 & [1] & main & 0.8895 & 0.0058(0.0075) & 0.0099 & 0.0104 & $\checkmark$ \\
 & [2] & main & 2.1271 & 0.0269(0.0306) & 0.0099 & 0.0104 & $\checkmark$ \\
 & [3] & main & 9.0056 & 0.0147(0.0385) & 0.0099 & 0.0104 & $\checkmark$ \\
\texttt{weak\_main} & [0] & main & 2.1248 & 0.0073(0.0154) & 0.0099 & 0.0104 & $\checkmark$ \\
 & [1] & main & 0.8572 & 0.0032(0.0064) & 0.0099 & 0.0104 & $\checkmark$ \\
 & [2] & main & 2.0811 & 0.0163(0.0196) & 0.0099 & 0.0104 & $\checkmark$ \\
\texttt{inter\_no\_overlap} & [0] & main & 2.1180 & 0.0067(0.0153) & 0.0099 & 0.0104 & $\checkmark$ \\
 & [1] & main & 0.7909 & 0.0028(0.0067) & 0.0099 & 0.0104 & $\checkmark$ \\
 & [2] & main & 2.1243 & 0.0212(0.0244) & 0.0099 & 0.0104 & $\checkmark$ \\
 & (3,4) & interaction & 0.5228 & 0.0140(0.0280) & 0.0099 & 0.0104 & $\checkmark$ \\
\texttt{inter\_mild\_overlap} & [0] & main & 1.8749 & 0.0044(0.0102) & 0.0099 & 0.0104 & $\checkmark$ \\
 & [1] & main & 0.8290 & 0.0032(0.0055) & 0.0099 & 0.0104 & $\checkmark$ \\
 & [2] & main & 1.4379 & 0.0169(0.0343) & 0.0099 & 0.0104 & $\checkmark$ \\
 & (2,3) & interaction & 0.7707 & 0.0399(0.0585) & 0.0099 & 0.0104 & $\checkmark$ \\
\texttt{inter\_strong\_overlap} & [0] & main & 2.1414 & 0.0054(0.0117) & 0.0099 & 0.0104 & $\checkmark$ \\
 & [1] & main & 0.9497 & 0.0046(0.0100) & 0.0099 & 0.0104 & $\checkmark$ \\
 & [2] & main & 0.0945 & 0.0091(0.0172) & 0.0198 & 0.0198 & $\checkmark$ \\
 & (1,2) & interaction & 0.2094 & 0.0102(0.0174) & 0.0099 & 0.0104 & $\checkmark$ \\
\texttt{only\_inter} & (0,1) & interaction & 0.7555 & 0.0085(0.0168) & 0.0099 & 0.0104 & $\checkmark$ \\
 & (2,3) & interaction & 0.7703 & 0.0045(0.0072) & 0.0099 & 0.0104 & $\checkmark$ \\
\bottomrule
\end{tabular}
\end{adjustbox}
\end{table}

\section{Robustness to correlated and heteroscedastic designs}\label{app:robust}
\subsection{Experiments with correlated covariates}\label{app:corr}
To evaluate robustness beyond independent designs, we generate correlated covariates using two dependence structures: an AR(1) correlation model and a block-correlation model. In both settings, the covariate vector 
$X=(X_1,\dots,X_p)^\top$ is generated from a mean-zero multivariate Gaussian distribution with covariance matrix $\Sigma$, and the response is then generated from the same structural model used in the corresponding baseline simulation case. For the AR(1) design, we set $\Sigma_{jk}=\rho^{|j-k|}$, so that correlations decay with distance between feature indices. For the block design, the features are partitioned into blocks of equal size, and within each block the pairwise correlation is set to $\rho$, while correlations across different blocks are set to zero. We consider representative values of $\rho$ to assess the stability of SDAMI under moderate and stronger dependence. After generating the correlated covariates, we use the same main-effect and interaction functions as in the corresponding synthetic cases in the main text. This design isolates the effect of predictor dependence without changing the underlying regression structure. We report both prediction metrics and support-recovery metrics in order to assess whether SDAMI remains accurate and structurally reliable when the independence assumption is relaxed.

\begin{table}[h]
\centering
\caption{Test MSE under correlated predictors (mean $\pm$ std).}
\label{tab:corr_mse}
\small
\adjustbox{max width=\textwidth}{
\begin{tabular}{llcccccc}
\toprule
Scenario & Corr.\ Type & $\rho$ & SDAMI & DNN & GAMI-NET & NodeGAM & NodeGA2M \\
\midrule
  Case 1 (Main only) & AR(1) & 0.3 & \textbf{0.352{\scriptsize$\pm$0.242}} & 10.698{\scriptsize$\pm$0.943} & 1.901{\scriptsize$\pm$0.700} & 0.686{\scriptsize$\pm$0.158} & 2.711{\scriptsize$\pm$0.261} \\
   & AR(1) & 0.6 & \textbf{0.310{\scriptsize$\pm$0.241}} & 9.537{\scriptsize$\pm$0.747} & 1.338{\scriptsize$\pm$0.476} & 0.662{\scriptsize$\pm$0.112} & 2.612{\scriptsize$\pm$0.188} \\
   & Block & 0.3 & \textbf{0.335{\scriptsize$\pm$0.278}} & 10.560{\scriptsize$\pm$0.768} & 1.248{\scriptsize$\pm$0.445} & 0.771{\scriptsize$\pm$0.316} & 2.552{\scriptsize$\pm$0.487} \\
   & Block & 0.6 & \textbf{0.249{\scriptsize$\pm$0.105}} & 9.227{\scriptsize$\pm$0.630} & 1.937{\scriptsize$\pm$1.025} & 0.850{\scriptsize$\pm$0.348} & 2.287{\scriptsize$\pm$0.355} \\
\addlinespace[3pt]
  Case 3 (No overlap) & AR(1) & 0.3 & \textbf{0.218{\scriptsize$\pm$0.103}} & 5.469{\scriptsize$\pm$0.323} & 1.614{\scriptsize$\pm$0.767} & 0.685{\scriptsize$\pm$0.157} & 2.310{\scriptsize$\pm$0.269} \\
   & AR(1) & 0.6 & \textbf{0.217{\scriptsize$\pm$0.079}} & 5.253{\scriptsize$\pm$0.506} & 0.809{\scriptsize$\pm$0.325} & 0.743{\scriptsize$\pm$0.215} & 2.181{\scriptsize$\pm$0.239} \\
   & Block & 0.3 & \textbf{0.183{\scriptsize$\pm$0.061}} & 5.318{\scriptsize$\pm$0.403} & 1.264{\scriptsize$\pm$0.531} & 0.669{\scriptsize$\pm$0.172} & 2.414{\scriptsize$\pm$0.367} \\
   & Block & 0.6 & \textbf{0.279{\scriptsize$\pm$0.147}} & 4.474{\scriptsize$\pm$0.330} & 1.180{\scriptsize$\pm$0.744} & 0.747{\scriptsize$\pm$0.225} & 2.210{\scriptsize$\pm$0.225} \\
\addlinespace[3pt]
  Case 6 (Inter only) & AR(1) & 0.3 & \textbf{0.212{\scriptsize$\pm$0.027}} & 0.968{\scriptsize$\pm$0.070} & 0.218{\scriptsize$\pm$0.078} & 0.422{\scriptsize$\pm$0.027} & 0.475{\scriptsize$\pm$0.059} \\
   & AR(1) & 0.6 & 0.200{\scriptsize$\pm$0.031} & 0.934{\scriptsize$\pm$0.099} & \textbf{0.197{\scriptsize$\pm$0.059}} & 0.375{\scriptsize$\pm$0.043} & 0.502{\scriptsize$\pm$0.138} \\
   & Block & 0.3 & \textbf{0.223{\scriptsize$\pm$0.034}} & 0.958{\scriptsize$\pm$0.086} & 0.227{\scriptsize$\pm$0.082} & 0.424{\scriptsize$\pm$0.052} & 0.478{\scriptsize$\pm$0.052} \\
   & Block & 0.6 & 0.243{\scriptsize$\pm$0.159} & 0.916{\scriptsize$\pm$0.094} & \textbf{0.220{\scriptsize$\pm$0.089}} & 0.357{\scriptsize$\pm$0.037} & 0.402{\scriptsize$\pm$0.037} \\
\addlinespace[3pt]
\bottomrule
\end{tabular}}
\end{table}

\begin{table}[h]
\centering

\caption{Support recovery (TPR / FPR / F1) under correlated predictors.}
\label{tab:corr_support}
\small
\adjustbox{max width=\textwidth}{
\begin{tabular}{llcccccccccc}
\toprule
 & & & \multicolumn{3}{c}{SDAMI} & \multicolumn{3}{c}{LASSONET} & \multicolumn{3}{c}{SODA} \\
\cmidrule(lr){4-6} \cmidrule(lr){7-9} \cmidrule(lr){10-12}
Scenario & Corr.\ Type & $\rho$ & TPR & FPR & F1 & TPR & FPR & F1 & TPR & FPR & F1 \\
\midrule
  Case 1 (Main only) & AR(1) & 0.3 & 1.000 & 0.000 & 1.000 & 0.925 & 0.297 & 0.191 & 0.575 & 0.001 & 0.394 \\
   & AR(1) & 0.6 & 1.000 & 0.000 & 1.000 & 1.000 & 0.218 & 0.261 & 0.600 & 0.000 & 0.444 \\
   & Block & 0.3 & 1.000 & 0.000 & 1.000 & 0.925 & 0.284 & 0.204 & 0.550 & 0.000 & 0.397 \\
   & Block & 0.6 & 1.000 & 0.000 & 1.000 & 0.925 & 0.228 & 0.254 & 0.525 & 0.000 & 0.421 \\
\addlinespace[3pt]
  Case 3 (No overlap) & AR(1) & 0.3 & 1.000 & 0.000 & 1.000 & 0.800 & 0.288 & 0.303 & 0.375 & 0.000 & 0.278 \\
   & AR(1) & 0.6 & 1.000 & 0.000 & 1.000 & 0.780 & 0.226 & 0.368 & 0.300 & 0.000 & 0.233 \\
   & Block & 0.3 & 1.000 & 0.000 & 1.000 & 0.800 & 0.312 & 0.281 & 0.275 & 0.000 & 0.214 \\
   & Block & 0.6 & 0.920 & 0.000 & 0.950 & 0.660 & 0.243 & 0.303 & 0.250 & 0.000 & 0.196 \\
\addlinespace[3pt]
  Case 6 (Inter only) & AR(1) & 0.3 & 1.000 & 0.000 & 1.000 & 0.650 & 0.219 & 0.257 & 0.000 & 0.001 & 0.000 \\
   & AR(1) & 0.6 & 1.000 & 0.000 & 1.000 & 0.900 & 0.292 & 0.213 & 0.100 & 0.001 & 0.051 \\
   & Block & 0.3 & 1.000 & 0.000 & 1.000 & 0.750 & 0.282 & 0.242 & 0.000 & 0.001 & 0.000 \\
   & Block & 0.6 & 0.925 & 0.000 & 0.940 & 0.950 & 0.305 & 0.210 & 0.200 & 0.001 & 0.067 \\
\addlinespace[3pt]
\bottomrule
\end{tabular}
}
\end{table}

\subsection{Experiments under Heteroscedastic Noise}\label{app:heteroscedastic}
To examine robustness beyond homoscedastic Gaussian noise, we consider two heteroscedastic data-generating mechanisms. In both settings, the response is generated as $Y=\mu(X)+\epsilon$,
where $\mu(X)$ is the same structured regression function used in the corresponding synthetic case, and the noise term $\epsilon$ is conditionally Gaussian with mean zero but input-dependent variance.

In the first setting, denoted Hetero-$X$, we let $\epsilon\mid X\sim N(0,\sigma^2(1+X_1^2))$,
so that the noise variance increases with the magnitude of the first covariate. This setting introduces input-dependent heteroscedasticity while keeping the conditional mean structure unchanged. In the second setting, denoted Hetero-$\mu$, we let $\epsilon\mid X\sim N(0,\sigma^2(1+|\mu(X)|))$,
so that the variance grows with the magnitude of the conditional mean. This setting is more challenging because regions with larger signal amplitude are also noisier.

These two heteroscedastic designs allow us to assess whether SDAMI remains stable when the error variance is no longer constant across the input space. We report mean squared error together with support-recovery statistics, thereby evaluating both predictive robustness and structural robustness under variance misspecification.

\paragraph{Performance under correlated covariates.}
Table~\ref{tab:corr_support} reports the prediction and support-recovery results under correlated predictor designs. Compared with the baseline independent setting, all methods become more challenged when the covariates are correlated, as dependence among predictors makes both screening and attribution more difficult. Despite this, SDAMI remains competitive and typically achieves the strongest overall combination of low prediction error and high recovery accuracy across the considered correlation structures.

\paragraph{Performance under heteroscedastic noise.}
Table~\ref{tab:hetero_mse} summarizes the results under two heteroscedastic noise settings, one in which the noise variance depends on an input covariate and another in which it depends on the signal magnitude. As expected, heteroscedasticity increases the difficulty of the estimation problem by introducing nonconstant uncertainty across the input space. Nevertheless, SDAMI remains competitive in both settings and generally preserves its advantage in prediction accuracy relative to the baselines.

\begin{table}[H]
\centering
\caption{Test MSE under heteroscedastic noise (mean $\pm$ std, 10 repeats)}
\label{tab:hetero_mse}
\resizebox{\textwidth}{!}{%
\begin{tabular}{llccccc}
\toprule
Case & Noise & SDAMI & DNN & GAMI-NET & NodeGAM & NodeGA2M \\
\midrule
Only Main & Homo & 0.25$\pm$0.14 & 11.76$\pm$1.03 & 2.54$\pm$1.22 & 0.63$\pm$0.17 & 2.56$\pm$0.60 \\
 & Hetero-$X_1$ & 1.01$\pm$0.15 & 11.91$\pm$0.75 & 3.81$\pm$1.22 & 2.02$\pm$0.25 & 3.98$\pm$0.40 \\
 & Hetero-$\mu$ & 0.56$\pm$0.12 & 11.50$\pm$0.50 & 2.79$\pm$0.92 & 1.17$\pm$0.20 & 3.06$\pm$0.32 \\
\midrule
Weak Main & Homo & 0.12$\pm$0.01 & 4.97$\pm$0.35 & 1.02$\pm$0.65 & 0.43$\pm$0.24 & 1.74$\pm$0.31 \\
 & Hetero-$X_1$ & 0.88$\pm$0.05 & 5.68$\pm$0.21 & 1.89$\pm$0.54 & 1.87$\pm$0.33 & 2.92$\pm$0.37 \\
 & Hetero-$\mu$ & 0.33$\pm$0.03 & 5.18$\pm$0.25 & 1.49$\pm$0.49 & 0.81$\pm$0.09 & 2.19$\pm$0.33 \\
\midrule
Inter (no) & Homo & 0.18$\pm$0.14 & 5.43$\pm$0.39 & 1.20$\pm$0.52 & 0.60$\pm$0.19 & 2.37$\pm$0.26 \\
 & Hetero-$X_1$ & 1.25$\pm$0.41 & 6.11$\pm$0.22 & 2.23$\pm$0.73 & 2.02$\pm$0.18 & 3.41$\pm$0.48 \\
 & Hetero-$\mu$ & 0.53$\pm$0.19 & 5.62$\pm$0.31 & 1.69$\pm$0.49 & 1.07$\pm$0.27 & 2.60$\pm$0.44 \\
\midrule
Inter (mild) & Homo & 0.19$\pm$0.10 & 6.24$\pm$0.38 & 1.44$\pm$0.49 & 0.58$\pm$0.16 & 2.29$\pm$0.25 \\
 & Hetero-$X_1$ & 1.00$\pm$0.10 & 7.04$\pm$0.28 & 2.50$\pm$0.65 & 1.94$\pm$0.24 & 3.42$\pm$0.41 \\
 & Hetero-$\mu$ & 0.45$\pm$0.11 & 6.48$\pm$0.27 & 2.03$\pm$0.54 & 0.96$\pm$0.08 & 2.91$\pm$0.38 \\
\midrule
Inter (strong) & Homo & 0.14$\pm$0.03 & 5.47$\pm$0.28 & 1.22$\pm$0.41 & 0.66$\pm$0.18 & 2.22$\pm$0.74 \\
 & Hetero-$X_1$ & 0.93$\pm$0.06 & 6.29$\pm$0.35 & 1.95$\pm$0.64 & 1.96$\pm$0.38 & 3.56$\pm$0.78 \\
 & Hetero-$\mu$ & 0.38$\pm$0.04 & 5.63$\pm$0.39 & 1.38$\pm$0.46 & 1.02$\pm$0.24 & 2.35$\pm$0.34 \\
\midrule
Only Inter & Homo & 0.11$\pm$0.01 & 0.96$\pm$0.07 & 0.24$\pm$0.15 & 0.41$\pm$0.04 & 0.47$\pm$0.09 \\
 & Hetero-$X_1$ & 1.28$\pm$0.24 & 1.64$\pm$0.11 & 1.18$\pm$0.15 & 1.53$\pm$0.14 & 1.38$\pm$0.14 \\
 & Hetero-$\mu$ & 0.25$\pm$0.07 & 1.03$\pm$0.08 & 0.42$\pm$0.09 & 0.53$\pm$0.04 & 0.57$\pm$0.08 \\
\bottomrule
\end{tabular}}
\end{table}

\begin{table}[H]
\centering
\caption{Support recovery under heteroscedastic noise (SDAMI vs LassoNet vs SODA, $n=300$)}
\label{tab:hetero_support}
\resizebox{\textwidth}{!}{%
\begin{tabular}{llccccccccc}
\toprule
 & & \multicolumn{3}{c}{SDAMI} & \multicolumn{3}{c}{LassoNet} & \multicolumn{3}{c}{SODA} \\
\cmidrule(lr){3-5} \cmidrule(lr){6-8} \cmidrule(lr){9-11}
Case & Noise & TPR & FPR & F1 & TPR & FPR & F1 & TPR & FPR & F1 \\
\midrule
Only Main & Homo & 1.000$\pm$0.000 & 0.000$\pm$0.000 & 1.000$\pm$0.000 & 0.875$\pm$0.168 & 0.387$\pm$0.195 & 0.129$\pm$0.047 & 0.625$\pm$0.125 & 0.001$\pm$0.000 & 0.395$\pm$0.065 \\
 & Hetero-$X_1$ & 1.000$\pm$0.000 & 0.000$\pm$0.000 & 1.000$\pm$0.000 & 0.675$\pm$0.160 & 0.122$\pm$0.115 & 0.285$\pm$0.120 & 0.600$\pm$0.122 & 0.000$\pm$0.000 & 0.404$\pm$0.085 \\
 & Hetero-$\mu$ & 1.000$\pm$0.000 & 0.000$\pm$0.000 & 1.000$\pm$0.000 & 0.850$\pm$0.122 & 0.242$\pm$0.140 & 0.188$\pm$0.065 & 0.625$\pm$0.125 & 0.000$\pm$0.000 & 0.437$\pm$0.096 \\
\midrule
Weak Main & Homo & 0.750$\pm$0.000 & 0.000$\pm$0.000 & 0.857$\pm$0.000 & 0.800$\pm$0.150 & 0.218$\pm$0.126 & 0.203$\pm$0.113 & 0.375$\pm$0.168 & 0.000$\pm$0.000 & 0.286$\pm$0.119 \\
 & Hetero-$X_1$ & 0.750$\pm$0.000 & 0.000$\pm$0.000 & 0.857$\pm$0.000 & 0.650$\pm$0.200 & 0.149$\pm$0.147 & 0.283$\pm$0.164 & 0.375$\pm$0.125 & 0.000$\pm$0.000 & 0.279$\pm$0.070 \\
 & Hetero-$\mu$ & 0.750$\pm$0.000 & 0.000$\pm$0.000 & 0.857$\pm$0.000 & 0.675$\pm$0.225 & 0.138$\pm$0.137 & 0.314$\pm$0.170 & 0.375$\pm$0.168 & 0.000$\pm$0.000 & 0.288$\pm$0.132 \\
\midrule
Inter (no) & Homo & 0.960$\pm$0.120 & 0.000$\pm$0.000 & 0.975$\pm$0.075 & 0.760$\pm$0.120 & 0.235$\pm$0.143 & 0.271$\pm$0.212 & 0.350$\pm$0.122 & 0.001$\pm$0.000 & 0.242$\pm$0.079 \\
 & Hetero-$X_1$ & 0.780$\pm$0.209 & 0.000$\pm$0.000 & 0.860$\pm$0.141 & 0.660$\pm$0.201 & 0.179$\pm$0.176 & 0.264$\pm$0.099 & 0.350$\pm$0.166 & 0.001$\pm$0.000 & 0.234$\pm$0.074 \\
 & Hetero-$\mu$ & 0.860$\pm$0.180 & 0.000$\pm$0.000 & 0.914$\pm$0.112 & 0.660$\pm$0.201 & 0.195$\pm$0.169 & 0.291$\pm$0.194 & 0.350$\pm$0.122 & 0.001$\pm$0.000 & 0.232$\pm$0.074 \\
\midrule
Inter (mild) & Homo & 0.950$\pm$0.100 & 0.000$\pm$0.000 & 0.971$\pm$0.057 & 0.675$\pm$0.195 & 0.227$\pm$0.129 & 0.181$\pm$0.099 & 0.325$\pm$0.115 & 0.000$\pm$0.000 & 0.274$\pm$0.093 \\
 & Hetero-$X_1$ & 0.925$\pm$0.115 & 0.000$\pm$0.000 & 0.957$\pm$0.065 & 0.675$\pm$0.225 & 0.175$\pm$0.166 & 0.278$\pm$0.173 & 0.300$\pm$0.100 & 0.000$\pm$0.000 & 0.256$\pm$0.080 \\
 & Hetero-$\mu$ & 0.925$\pm$0.115 & 0.000$\pm$0.000 & 0.957$\pm$0.065 & 0.650$\pm$0.229 & 0.132$\pm$0.156 & 0.322$\pm$0.168 & 0.300$\pm$0.100 & 0.000$\pm$0.000 & 0.245$\pm$0.079 \\
\midrule
Inter (strong) & Homo & 1.000$\pm$0.000 & 0.000$\pm$0.000 & 1.000$\pm$0.000 & 0.933$\pm$0.200 & 0.218$\pm$0.129 & 0.205$\pm$0.120 & 0.575$\pm$0.160 & 0.000$\pm$0.000 & 0.392$\pm$0.095 \\
 & Hetero-$X_1$ & 1.000$\pm$0.000 & 0.000$\pm$0.000 & 1.000$\pm$0.000 & 0.900$\pm$0.213 & 0.195$\pm$0.149 & 0.284$\pm$0.222 & 0.475$\pm$0.135 & 0.000$\pm$0.000 & 0.345$\pm$0.105 \\
 & Hetero-$\mu$ & 1.000$\pm$0.000 & 0.000$\pm$0.000 & 1.000$\pm$0.000 & 0.967$\pm$0.100 & 0.153$\pm$0.115 & 0.309$\pm$0.205 & 0.525$\pm$0.135 & 0.000$\pm$0.000 & 0.393$\pm$0.066 \\
\midrule
Only Inter & Homo & 1.000$\pm$0.000 & 0.000$\pm$0.000 & 1.000$\pm$0.000 & 0.575$\pm$0.160 & 0.210$\pm$0.193 & 0.210$\pm$0.151 & 0.050$\pm$0.150 & 0.001$\pm$0.000 & 0.025$\pm$0.075 \\
 & Hetero-$X_1$ & 0.450$\pm$0.312 & 0.000$\pm$0.000 & 0.551$\pm$0.329 & 0.600$\pm$0.122 & 0.090$\pm$0.120 & 0.381$\pm$0.173 & 0.000$\pm$0.000 & 0.001$\pm$0.000 & 0.000$\pm$0.000 \\
 & Hetero-$\mu$ & 0.875$\pm$0.301 & 0.000$\pm$0.000 & 0.886$\pm$0.298 & 0.675$\pm$0.160 & 0.192$\pm$0.252 & 0.327$\pm$0.234 & 0.000$\pm$0.000 & 0.001$\pm$0.000 & 0.000$\pm$0.000 \\
\bottomrule
\end{tabular}}
\end{table}

\subsection{Binary-response extension}\label{app:binomial}
We generate the binomial data by following Table~\ref{tab:binom_general},~\ref{tab:binom_components}, and~\ref{tab:binom_cases}. The SDAMI pipeline is adapted to the binary setting as follows. In Stage 1, we replace the Gaussian additive screening model by a quasi-likelihood additive model under the logit link in order to screen variables with either main effects or effect footprints. In Stage 2, we apply grouped selection to the screened set using a logistic or quasi-likelihood-based grouped formulation, depending on the implementation. In Stage 3, the final structured neural model is trained using binary cross-entropy loss rather than squared loss. This adaptation preserves the same three-stage logic of screening, decomposition, and structured nonlinear estimation.

Table~\ref{appendix:binom_response}report the results of the binary-response extension of SDAMI. Across the considered simulation scenarios, SDAMI achieves strong classification performance, as measured by both accuracy and negative log-likelihood, while also maintaining high support-recovery quality. This indicates that the structural advantages of the screening--decomposition--estimation pipeline are not restricted to the Gaussian regression setting.
\begin{table}[H]
\centering
\caption{Binomial simulation design: general settings.}
\label{tab:binom_general}
\begin{tabular}{ll}
\toprule
\textbf{Parameter} & \textbf{Value} \\
\midrule
Response distribution & $Y_i \sim \mathrm{Bernoulli}\bigl(\sigma(\eta_i)\bigr)$ \\
Link function $\sigma(\cdot)$ & Logistic: $\sigma(z) = 1/(1 + e^{-z})$ \\[4pt]
Linear predictor & $\displaystyle \eta_i = \frac{3}{\hat{\sigma}_\mu} \sum_{k} \frac{g_k(\mathbf{x}_i)}{\hat{\sigma}_k}$ \\[10pt]
Signal scaling $\hat{\sigma}_\mu$ & Std.\ dev.\ of total signal $\sum_k \tilde{g}_k$ over training samples \\
Component scaling $\hat{\sigma}_k$ & Std.\ dev.\ of component $g_k$ over training samples \\
Scaling constant & 3 (ensures non-trivial class separation) \\
\midrule
Sample size $n$ & 500 \\
Number of predictors $k$ & 150 \\
Predictor distribution $X_j$ & $\mathrm{Uniform}(-2.5,\; 2.5)$, i.i.d. \\
Number of replications & 10 \\
Oracle NLL lower bound & 0.33--0.35 (computed from true $\eta_i$) \\
\bottomrule
\end{tabular}
\end{table}

\begin{table}[H]
\centering
\caption{Component functions $g_k(\cdot)$ for the binomial simulation; all functions are centered before computing $\eta_i$.}
\label{tab:binom_components}
\begin{tabular}{lll}
\toprule
\textbf{Label} & \textbf{Type} & \textbf{Functional form} \\
\midrule
$g_1(x) = f_1(x)$ & Main effect & $-2\sin(2x)$ \\
$g_2(x) = f_2(x)$ & Main effect & $x^2/2 + 1$ \\
$g_3(x) = f_3(x)$ & Main effect & $x - 1/2$ \\
$g_4(x) = f_4(x)$ & Main effect & $e^{-x} + e^{-1} - 1$ \\
$g_{\mathcal{I}}(x_a, x_b) = f_{\mathcal{I}}(x_a,x_b)$ & Interaction & $\exp\bigl(\sin(x_a) + \cos(x_b) - 1\bigr)$ \\
\bottomrule
\end{tabular}
\end{table}

\begin{table}[H]
\centering
\caption{Binomial simulation cases. Each case uses the same component functions as the Gaussian counterpart, passed through the logistic link to generate binary responses.}
\label{tab:binom_cases}
\small
\begin{tabular}{@{}llccp{5.5cm}@{}}
\toprule
\textbf{Case} & \textbf{Label} & $|\mathcal{M}|$ & $|\mathcal{I}|$ & \textbf{Components entering $\eta_i$} \\
\midrule
1 & Only Main
  & 4 & 0
  & $g_1(x_1) + g_2(x_2) + g_3(x_3) + g_4(x_4)$ \\[4pt]

3 & No Overlap
  & 3 & 2
  & $g_1(x_1) + g_2(x_2) + g_3(x_3) + g_{\mathcal{I}}(x_4, x_5)$ \\[4pt]

5 & Strong Overlap
  & 3 & 2
  & $g_1(x_1) + g_2(x_2) + g_3(x_3) + g_{\mathcal{I}}(x_2, x_3)$ \\[4pt]

6 & Inter Only
  & 0 & 4
  & $g_{\mathcal{I}}(x_1, x_2) + g_{\mathcal{I}}(x_3, x_4)$ \\
\bottomrule
\end{tabular}
\end{table}

\begin{table}[htbp]
  \centering
  \small
  \caption{Binomial GLM performance: SDAMI vs. DNN (mean $\pm$ std)}\label{appendix:binom_response}
  \begin{adjustbox}{max width=\textwidth}
  \begin{tabular}{lcccc|ccc}
    \toprule
    \multirow{2}{*}{Case} & \multicolumn{2}{c}{Loglikelihood (Neg) $\downarrow$} & \multicolumn{2}{c}{Accuracy $\uparrow$} & \multicolumn{2}{|c}{Support Recovery (SDAMI)} \\
    \cmidrule(r){2-3} \cmidrule(lr){4-5} \cmidrule(l){6-7}
    & SDAMI & DNN & SDAMI & DNN & TPR $\uparrow$ & FPR $\downarrow$ \\
    \midrule
    Only Main          & $\mathbf{0.39 \pm 0.03}$ & $0.68 \pm 0.01$ & $\mathbf{0.82 \pm 0.01}$ & $0.56 \pm 0.02$ & $1.00 \pm 0.00$ & $0.00 \pm 0.00$ \\
    Inter (no overlap) & $\mathbf{0.44 \pm 0.07}$ & $0.69 \pm 0.01$ & $\mathbf{0.78 \pm 0.05}$ & $0.57 \pm 0.02$ & $0.90 \pm 0.12$ & $0.00 \pm 0.00$ \\
    Inter (strong)     & $\mathbf{0.39 \pm 0.02}$ & $0.69 \pm 0.01$ & $\mathbf{0.82 \pm 0.01}$ & $0.55 \pm 0.03$ & $0.90 \pm 0.12$ & $0.00 \pm 0.00$ \\
    Only Inter         & $\mathbf{0.42 \pm 0.01}$ & $0.67 \pm 0.00$ & $\mathbf{0.81 \pm 0.01}$ & $0.58 \pm 0.01$ & $1.00 \pm 0.00$ & $0.00 \pm 0.00$ \\
    \bottomrule
  \end{tabular}
  \end{adjustbox}
  \vspace{1pt} \\
  \raggedright \footnotesize \textit{* Bold values indicate significantly better performance within the metric.}
\end{table}

\section{Computational complexity analysis}\label{appendix:computation}
The proposed three-stage procedure achieves computational efficiency by progressively reducing the search space. The SpAM Screening fits $k$ univariate smooth functions via coordinate descent, with complexity $\mathcal{O}(p\cdot n\log(n))$, where $\log(n)$ reflects smoothing spline computations. Subsequently, the group Lasso decomposition expands each screened variable into an orthogonal basis of dimension $b$ and applies group Lasso with $e$ iterations, yielding complexity $\mathcal{O}(e p b n)$. Lastly, the neural network fitting with each subnetworks with depth $L$ and width $W$ for approximate $E$ epochs. The total complexity is $\mathcal{O}(p\cdot n\log(n)+epnb+EnLW^2)$.

In high-dimensional settings where the full pairwise interaction space is $\binom{k}{2}$, SDAMI's three-stage decomposition yields substantial computational savings: by focusing only on screened variables. The computational cost is summarized in Table \ref{tab:runtime_comparison}.

\begin{table}[t]
\centering
\caption{Computational cost comparison across methods and datasets. 
Times reported in seconds, averaged over 10 runs for simulation and real dataset on GPU: Tesla V100-SXM2-32GB.}\label{tab:runtime_comparison}
\begin{adjustbox}{max width=\textwidth}
\begin{tabular}{lcccc}
\toprule
\textbf{Method} & \multicolumn4{c}{\textbf{Runtime (seconds)}} \\
\cmidrule(r){2-5}
 & \textbf{Simulation} & \textbf{Wine} & \textbf{BikeShare} & \textbf{CA Housing} \\
 & $(n=150,p=150)$ & $(n=4892, p=12)$ & $(n=17379, p=12)$ & $(n=20640, p=8)$\\
\midrule
SDAMI & 6.04 & 16.40 & 77.54 & 78.62 \\
DNN & 0.42 & 58.84 & 58.6 & 7.02  \\
fSpAM & 0.001 & 0.004 & 0.008 & 0.005  \\
LASSO & 0.01 & 0.01 & 0.02 & 0.01  \\
LASSONET & 43.72  & 168.04 & 363.88 & 391.01 \\
NAM & 7.04 & 19.78 & 65.31 & 61.24  \\
GAMI-Net & 38.22 & 23.96 & 55.67 & 40.62 \\
NODE-GAM & 130.55  & 458.33 & 1199.69 &  1871.75 \\
NODE-GA$^2$M & 185.75 & 448.15 & 1858.62 & 1864.26 \\
    \hline
    \bottomrule
\end{tabular}
\end{adjustbox}
\end{table} 

\section{Real data analysis}\label{appendix:real-data}
This section illustrates the additional experiment on two real datasets with redundant features including the parameter settings and corresponding explanation on the visualization.  Besides, we also use data without redundant features such as wine quality, bike share, and California housing. The description is summarized in Table~\ref{tab:datadescription}. 

\begin{table*}[h]
\caption{Table of real datasets with sources}
\label{tab:datadescription}
\begin{adjustbox}{max width=1.0\textwidth}
\begin{tabular}{ccccc}
Dataset & Source & Samples & Features & \multicolumn{1}{c}{Description} \\ \hline
Chip & \citep{hsu2020extraction} & 100 & 9 baseline + 21 noise & MOSFET device lifetime \\
Diabetes & scikit-learn & 200 & 10 baseline + 40 noise & Serum measurements \\
V1 fMRI & \cite{kay2008identifying} & 300 & 1800 Gabor & Primary visual cortex responses \\ \hline
Wine Quality & UCI ML Repository & 4898 & 11 & Wine quality \\
BikeShare & UCI ML Repository & 17379 & 12 & Capital Bikeshare hourly rental counts\\
California Housing & scikit-learn & 20640 & 8 & Median house values  \\ \hline
\end{tabular}
\end{adjustbox}

\end{table*}

\subsection{Surrogate modeling of product lifetime}\label{appendix:product}
This subsection showcases the application of SDAMI in evaluating prediction performance, positioning it as an effective surrogate technique— a key approach in the field of computer experiments \citep{santner2019design, wu2011experiments}. Surrogate modeling serves as a statistical approximation of computationally intensive simulations, facilitating the efficient study of complex system dynamics. 

We illustrate this with the analysis of electronic device lifetimes, which can fail due to mechanisms such as front-end-of-line time-dependent dielectric breakdown (FEOL TDDB) \citep{yang2017front}. This failure occurs when traps accumulate in the gate oxide layer from electrical and thermal stress during operation, eventually creating conductive paths leading to device malfunction. The lifetime distribution for these components is captured by the following function, as characterized in prior work \citep{hsu2020extraction}:
\begin{equation}\label{eq: lifetime}
    S(t)  = \exp\left(-\left(\frac{t}{\text{A}_{\text{FEOL}} (\text{WL})^{-\frac{1}{\beta}} e^{-\frac{1}{\beta}} V^{a+bT} \exp\left(\frac{cT+d}{T^2}\right) s^{-1}}\right)^\beta\right),
\end{equation}
where the inputs include process-dependent constants $\text{A}_{\text{FEOL}}, a, b, c, d$, voltage $V$ and temperature $T$,  width $W$ and length $L$  of the device, the probability of stress $s$, and shape parameter $\beta$ describing failure progression over time.  

Although simulating such experiments is straightforward, accurately extracting main and higher-order effects under data sparsity requires sophisticated and interpretable modeling. To that end, we employ the  {\it MaxPro design} \citep{joseph2015maximum} to generate space-filling experiments spanning all input factors, with details in Table \ref{tab:part}. The dataset includes 100 observations with 9 covariates, augmented by 21 irrelevant noise features randomly sampled uniformly within $[0,1)$ to test model sparsistency and interaction detection. The log-transformation of the true model is given by
\begin{align}\label{eqn: logchip}
    \log(\eta)=\log(\text{A}_{FEOL})-\frac{1}{\beta}\log(\text{WL})-\frac{1}{\beta}+({a+bT})\log(V)+\bigg(\frac{cT+d}{T^2}\bigg)-\log(s),
\end{align}
where $s$ is constant and $\eta$ corresponds to a $63\%$ failure quantile from the generalized Wei-bull model (\ref{eq: lifetime}). This representation admits an additive decomposition involving univariate and bi-variate functions \ref{eqn: logchip},
\begin{align*}
    y = \alpha+\underset{i}{\sum}f_i(x_i)+\underset{i\neq j}{\sum}f_{ij}(x_i, x_j)+\cdots+\epsilon.
\end{align*}
allowing comprehensive identification of relevant main and interaction effects. Table \ref{tab: real} presents the comparative performance of various techniques, including SDAMI, NAM, GAMI-Net, NODE-GAM, NODE-GA$^2$M, DNN, LASSO, LASSONET, and fSpAM, demonstrating SDAMI's prominence in recovering complex dependency structures in sparse, high-dimensional settings.

\begin{figure}
    \centering
    \includegraphics[width=0.8\linewidth]{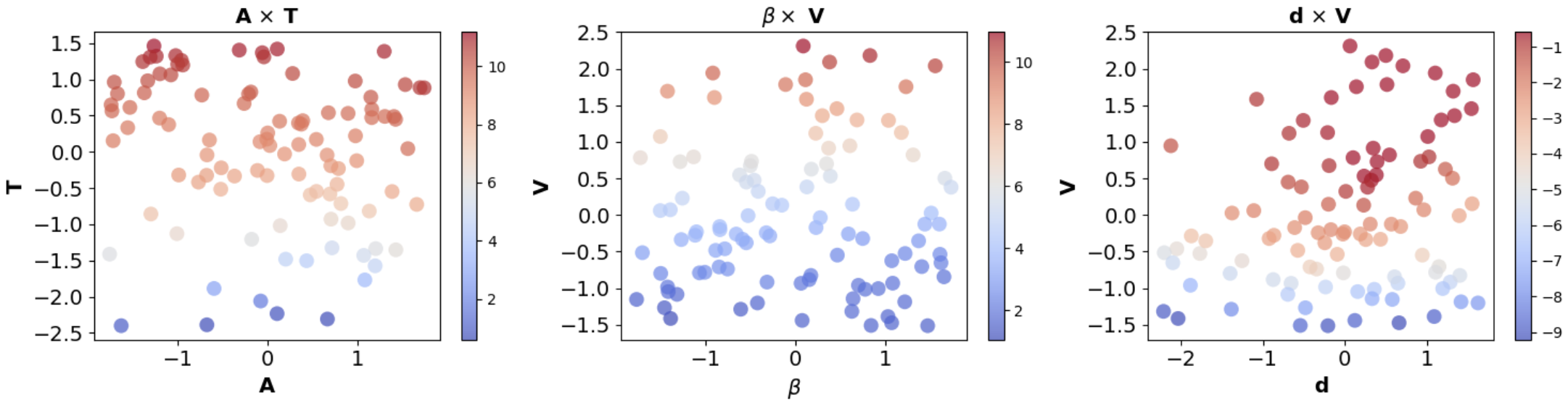}
    \caption{The shape plots of 3 interactions of SDAMI trained on chip dataset.
    }
    \label{fig:chipvisual}
\end{figure}

Given the visualization of effects from Figure \ref{fig:chipvisual}, we can observe that the contribution of main effect is relatively weak. Besides, the interactions have an obvious effect on the response. To be more specific, when $(A, T)$, $(d,V)$ and $(\beta, V)$ goes up, the response will increase. This phenomenon is predictable because in Equation \ref{eqn: logchip}, the higher-order effects are dominant over main effects but the main effects still exist due to its marginal effect on the response.

\begin{table}[h!]
\centering
\caption{Parameter table for generating space-filling experiment on MOSFET device}
\label{tab:part}
\begin{tabular}{|ccc|}
\hline
Parameter & Lower & Upper \\ \hline
a         & $-81.9$  & $-74.1$ \\ \hline
b         &  $7.69\times10^{-2}$ &  $8.51\times10^{-2}$ \\ \hline
c         &  $8.37\times10^{3}$& $9.25\times10^{3}$ \\ \hline
d         & $-8.14\times10^{5}$ & $-7.33\times10^{5}$ \\ \hline
$\beta$      & $1.476$  &  $1.804$     \\ \hline
V         &   $1.2$    &   $1.3$    \\ \hline
T         &   $120$    &   $180$    \\ \hline
WL        &   $4\times10^{-4}$ & $6\times10^{-4}$\\ \hline
A$_{\text{FEOL}}$  &  $4.75\times10^{-7}$  &  $5.25\times10^{-7}$\\ \hline
s         &   $1$    &   $1$    \\ \hline
\end{tabular}

\end{table}

\subsection{Diabetes response prediction}\label{appendix:diabetes}
For this analysis, we utilize the well-known diabetes dataset from the scikit-learn library, which contains 442 observations and ten baseline covariates. These features capture key demographic and physiological measurements, such age (in years), sex (0: female, 1: male), body mass index (BMI), mean arterial blood pressure, and six standardized blood serum variables known to be relevant for diabetes progression. The target variable is a quantitative measure of disease progression observed one year after baseline, making the dataset suitable for regression modeling and biomarker analysis.

To thoroughly evaluate sparse additive modeling methods under high-dimensional constraints, we purposefully restrict the sample size to $n=200$ and augment the original dataset with $40$ synthetic covariates, each drawn independently from a uniform distribution on the interval $[0, 1)$ distribution. These additional features are explicitly designed to act as non-informative noise, challenging each model's ability to discern relevant predictors. Thus, the expanded dataset includes $50$ covariates in total, with the genuine signal confined to the original ten baseline measurements. Standard preprocessing, including normalization and scaling of all features, is performed to ensure comparability and numerical robustness in downstream modeling. This controlled, high-dimensional experimental setup provides a rigorous testbed for assessing the sensitivity and variable selection performance of SpAM, and other advanced machine learning algorithms in biomedical contexts.

\begin{figure}
    \centering
    \includegraphics[width=1.0\linewidth]{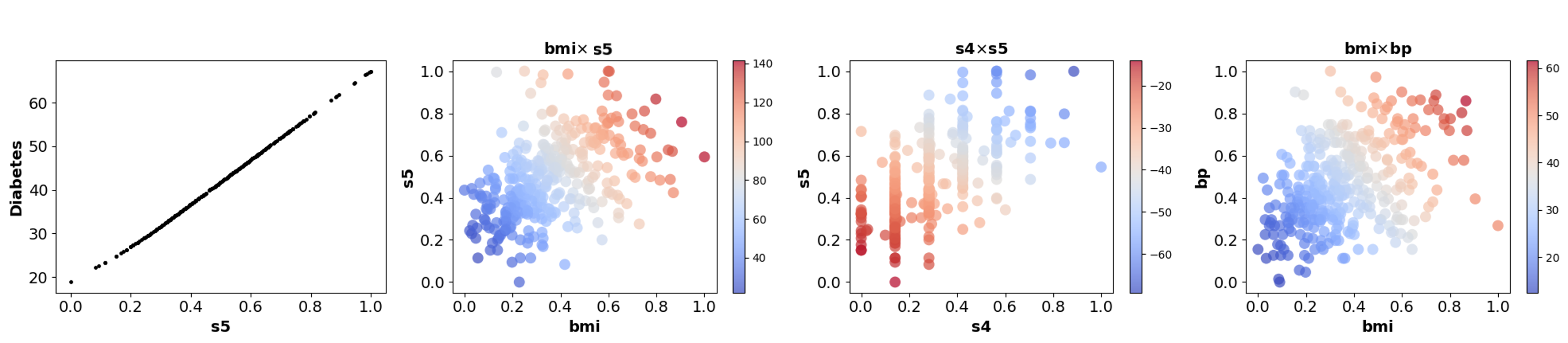}
    \caption{(Diabetes dataset) The first figures on the left: Predicted marginal response of target with respect to main effect feature; the three figures on the right: the shape plots of 3 Interactions of SDAMI trained on Diabetes Dataset.}
    \label{fig:casediabete}
\end{figure}

Visualization of the estimated effects in Figure \ref{fig:casediabete} reveals several interpretable patterns. One main effect and three interaction terms were selected by SDAMI. First, the log of Serum Triglycerides Level ($s5$) has a positive association with diabetes disease progression. The research shows that among patients with type 2 diabetes, those with elevated $s5$ had significantly worse glycemic control even when treated with insulin \citep{zheng2019association}. Second, higher disease progression when (high BMI and elevated $s5$) and (high BMI and elevated blood pressure) are present simultaneously. These combinations characterize the beginning stages of Cardiovascular-Kidney-Metabolic syndrome, which dramatically accelerates diabetes complications. The observed relationships align well with clinical expectations and domain knowledge. 

Table \ref{tab: real} summarizes model performance for diabetes response prediction. SDAMI with interaction modeling consistently outperforms alternative machine learning methods, offering superior predictive accuracy alongside enhanced interpretability thanks to its explicit feature selection and effect visualization capabilities. 

\begin{figure}
    \centering
    \includegraphics[width=1.0\linewidth]{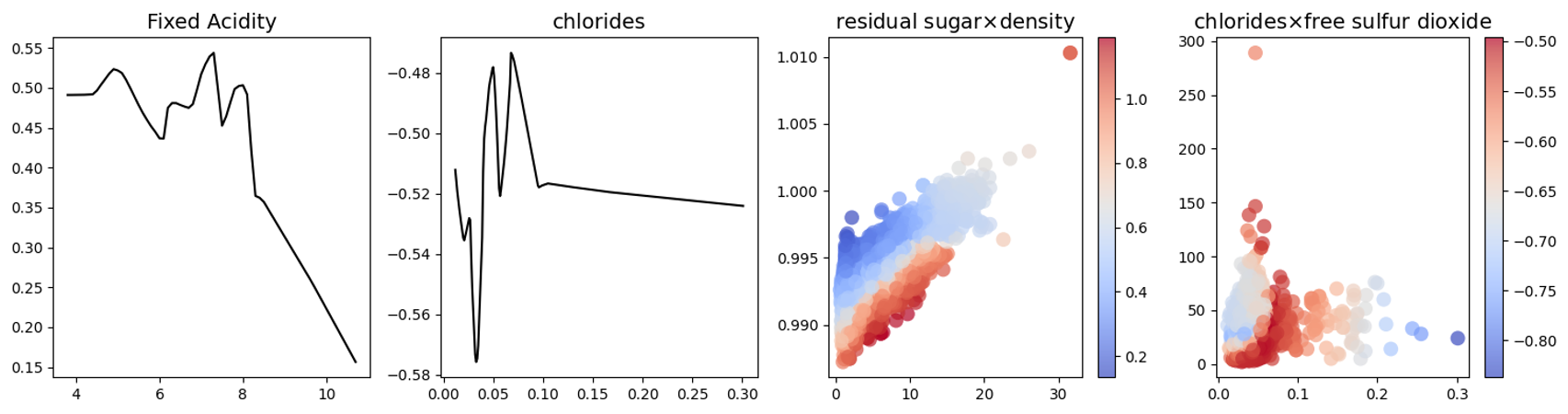}
    \caption{The shape plots of selected main effects and interaction of SDAMI trained on wine dataset.}
    \label{fig:case-wine}
\end{figure}

\begin{figure}
    \centering
    \includegraphics[width=1.0\linewidth]{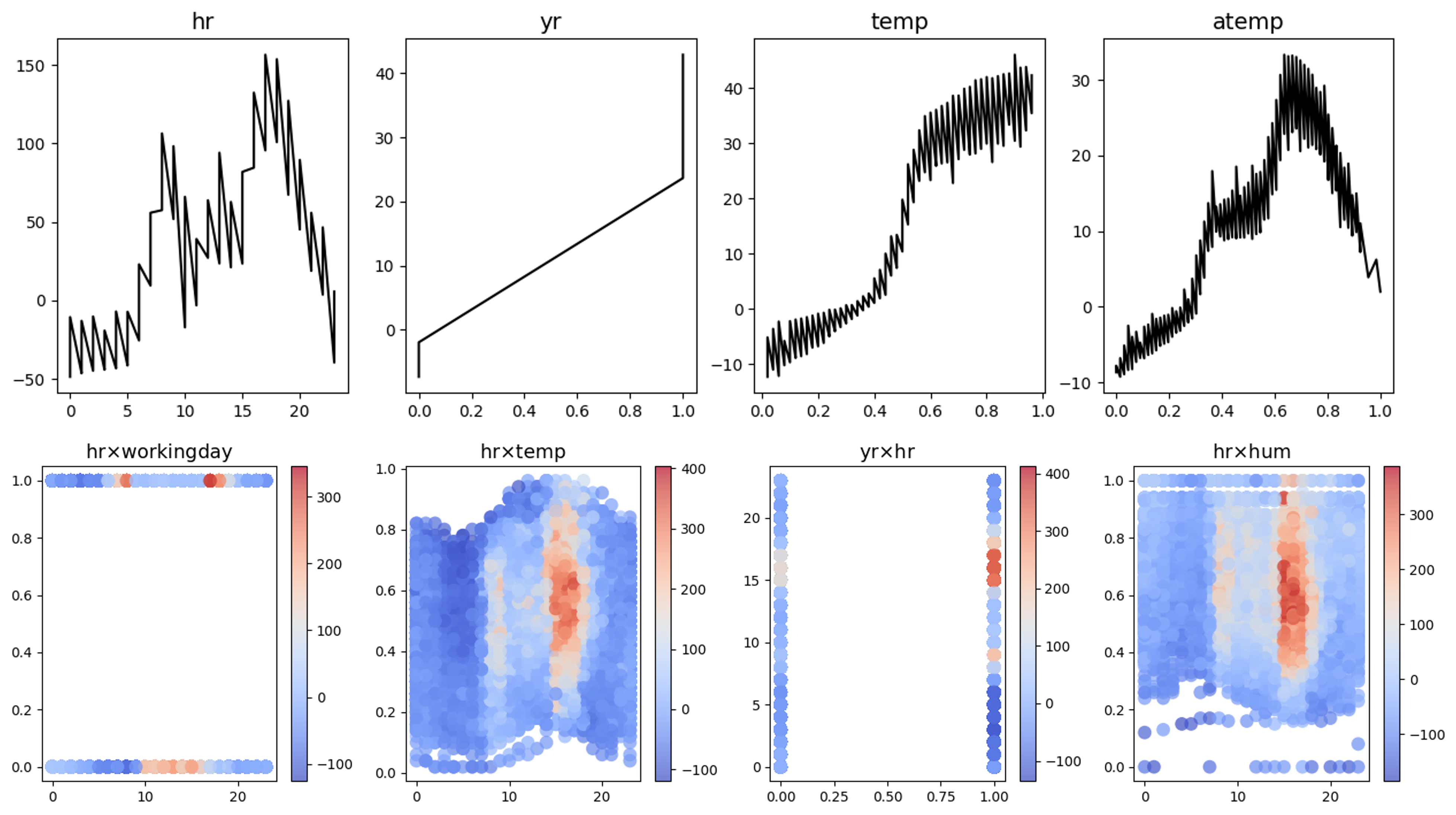}
    \caption{The shape plots of selected main effects and interaction of SDAMI trained on Bikeshares Dataset.}
    \label{fig:case-bike}
\end{figure}

\begin{figure}
    \centering
    \includegraphics[width=1.0\linewidth]{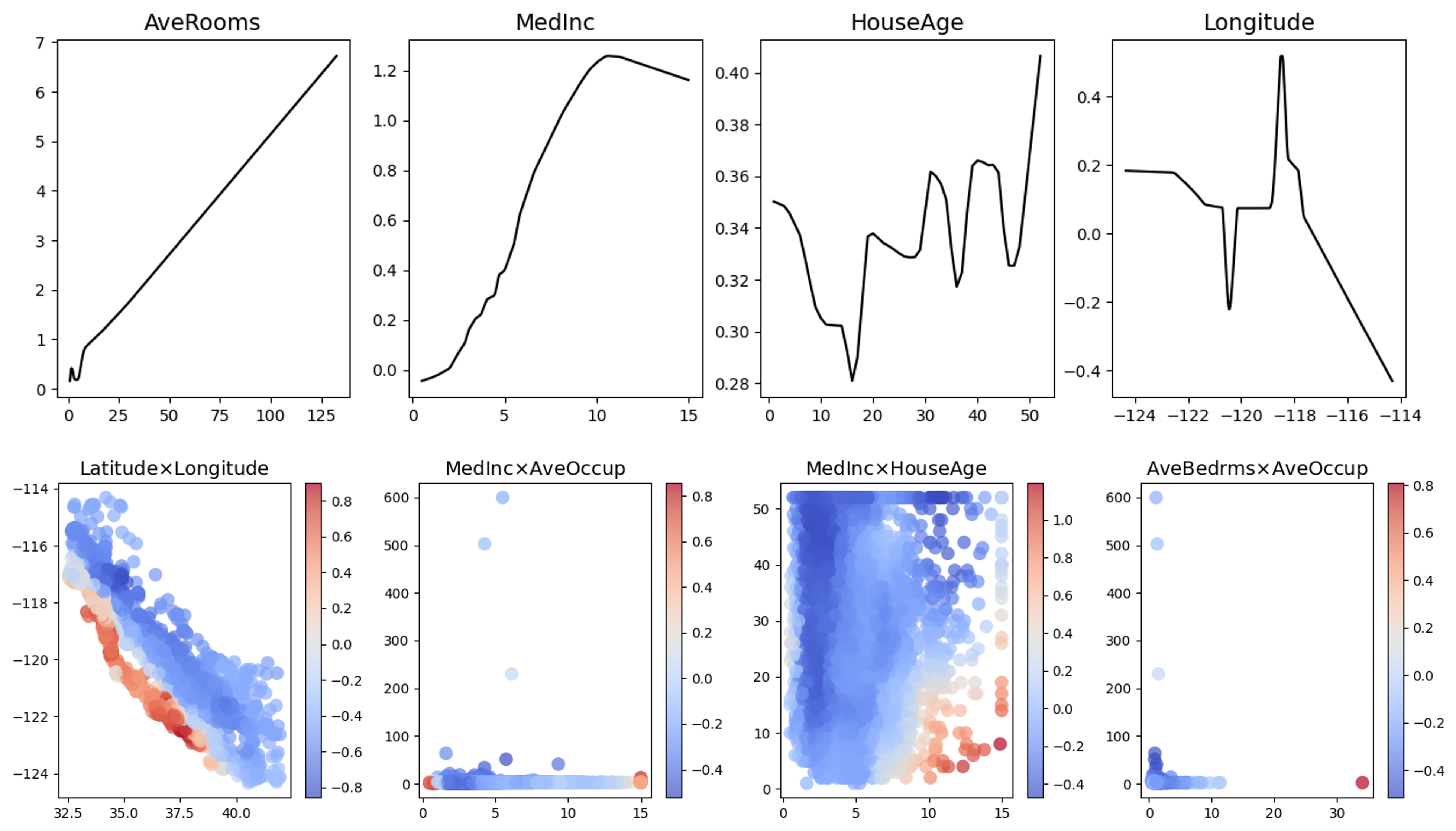}
    \caption{The shape plots of selected main effects and interaction of SDAMI trained on California housing dataset.}
    \label{fig:case-ca}
\end{figure}

We select the relative important main effects and interaction term for Wine, Bikeshares, and California housing where the shape plot demonstrate the relationship between features and the targeted response. The shape plots are provided in Figure~\ref{fig:case-wine}, \ref{fig:case-bike}, and \ref{fig:case-ca}. 

\subsection{Human primary visual cortex dataset}\label{appendix-v1}

The V1 fMRI dataset \citep{kay2008identifying} records voxel responses from human primary visual cortex at 2\,mm\(\times\)2\,mm\(\times\)2.5\,mm resolution on a 4-T scanner while subjects viewed grayscale natural images through a circular aperture. Stimuli are flashed three times per second with interleaved blanks, and signals are preprocessed to reduce noise and nonstationarity. Each voxel reflects pooled, rectified activity organized by a receptive-field hierarchy over space, frequency, and phase.

\begin{figure}
    \centering
    \includegraphics[width=1.0\linewidth]{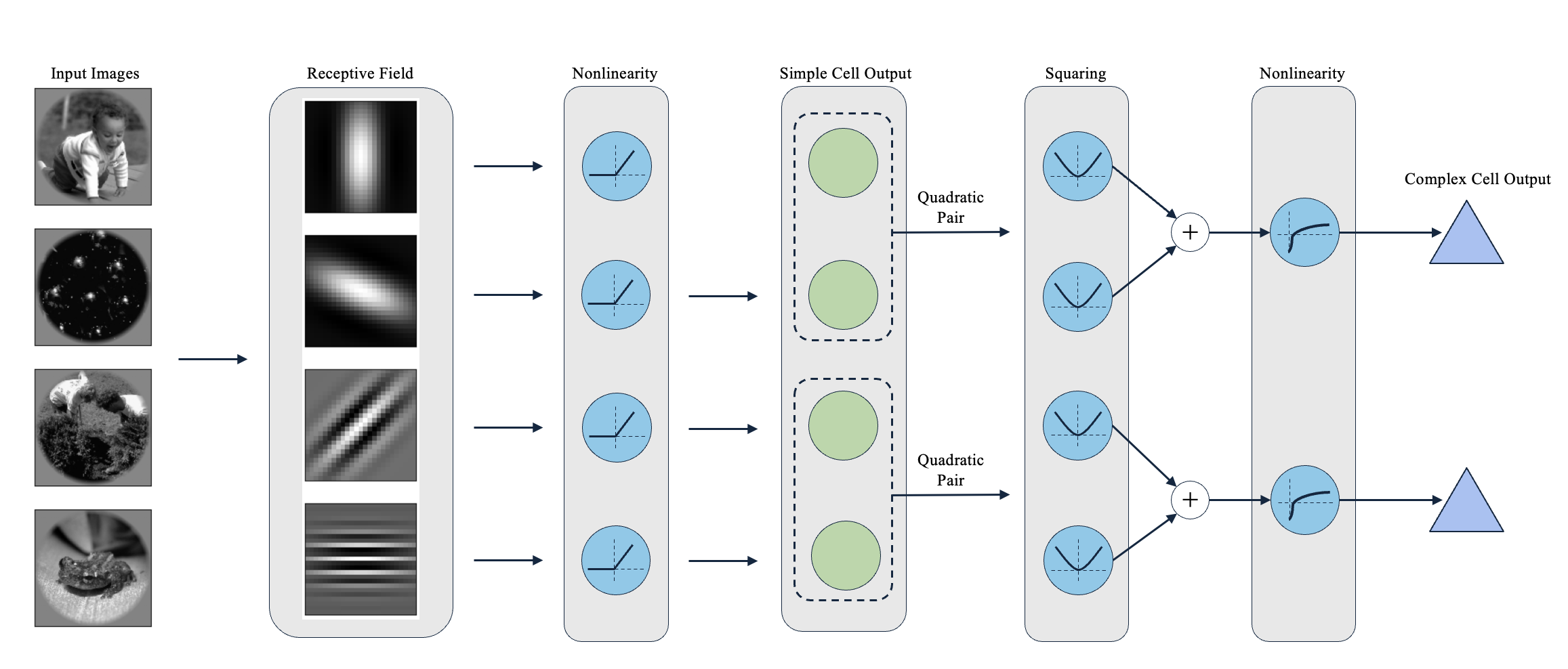}
    \caption{The formation of complex cells arises from nonlinear activation of quadratic pairs of simple cells generated by Gabor-wavelet filters applied to the input.}
    \label{fig:img2cp}
\end{figure}

\begin{figure}
    \centering
    \includegraphics[width=0.6\linewidth]{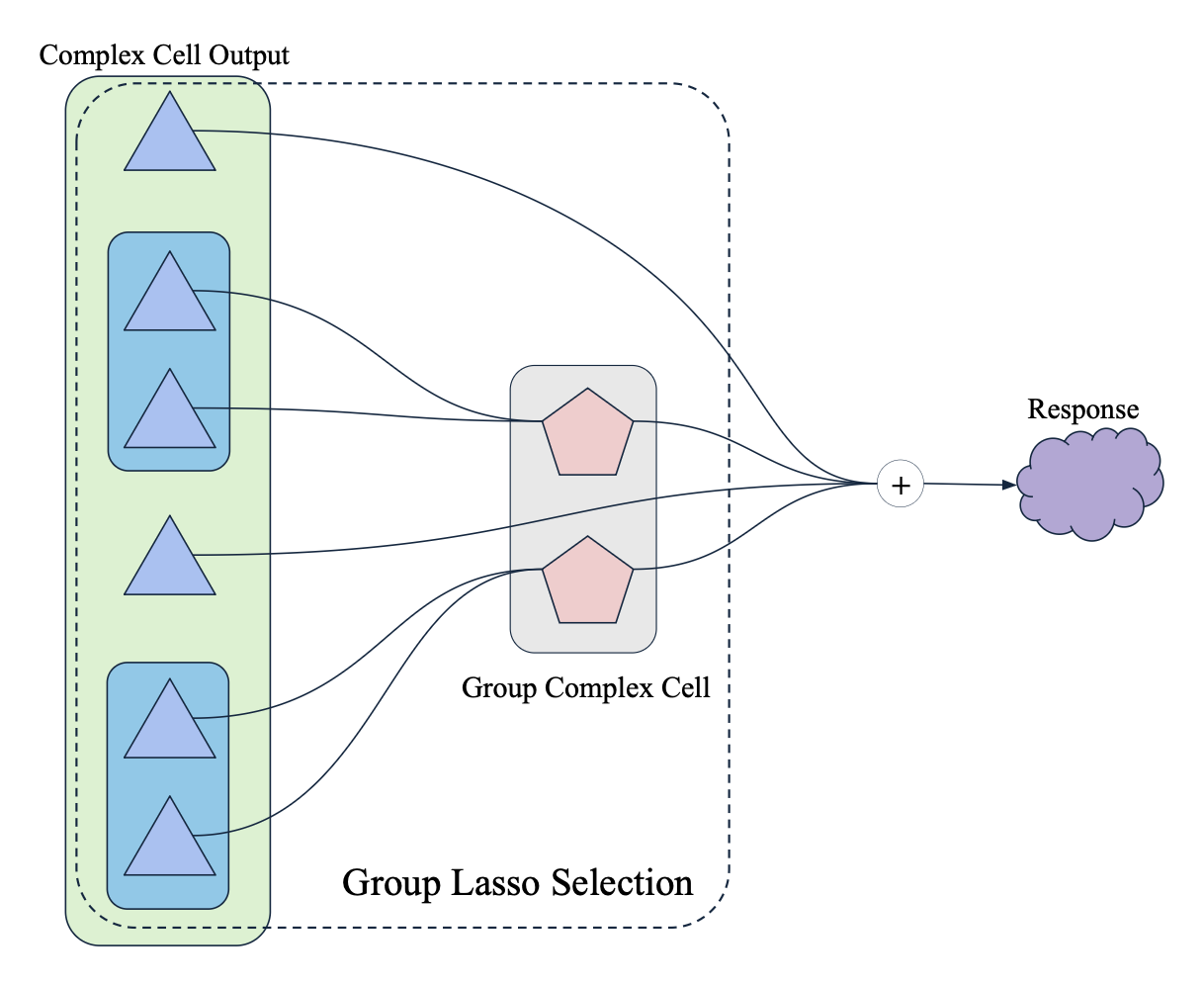}
        \caption{The formation of response arises from complex cells and group complex cells selected by group lasso applied to the input.}
        \label{fig:cp2rs}
\end{figure}

In Figure~\ref{fig:img2cp}, these images are first passed through localized, orientation- and phase-sensitive Gabor filters to mimic simple-cell receptive fields; outputs then undergo nonlinear transforms to produce single-cell responses. Complex cells are formed by pooling quadrature-phase pairs (square–sum–nonlinearity), yielding phase-invariant responses. Subsequently, in Figure~\ref{fig:cp2rs}, these complex cells will be fed into Group Lasso, and be identified as single input (main effect) or composite input (interaction)

\end{document}